\documentclass{article}

\usepackage[preprint]{neurips_2026}

\usepackage[utf8]{inputenc}
\usepackage[T1]{fontenc}
\usepackage{hyperref}
\usepackage{url}
\usepackage{booktabs}
\usepackage{amsfonts}
\usepackage{nicefrac}
\usepackage{amssymb}
\usepackage{microtype}
\usepackage{xcolor}

\usepackage{empheq}
\usepackage{threeparttable}

\usepackage{amsmath,amsfonts,bm}

\def\eqref#1{equation~\ref{#1}}

\def\1{\bm{1}}

\DeclareMathAlphabet{\mathsfit}{\encodingdefault}{\sfdefault}{m}{sl}
\SetMathAlphabet{\mathsfit}{bold}{\encodingdefault}{\sfdefault}{bx}{n}

\newcommand{\Var}{\mathrm{Var}}

\newcommand{\Cov}{\mathrm{Cov}}

\DeclareMathOperator{\Tr}{Tr}

\newcommand{\red}[1]{{\color{red}#1}}
\usepackage{amsthm}
\usepackage{cancel}
\usepackage{ulem}

\usepackage[most]{tcolorbox}

\tcbset{
  mytheoremstyle/.style={
    enhanced,
    breakable,
    colback=gray!5,
    colframe=gray!60,
    boxrule=0.6pt,
    arc=3pt,
    boxsep=3pt,
    top=5pt,
    bottom=5pt,
    left=6pt,
    right=6pt,
    before skip=8pt plus 2pt minus 2pt,
    after skip=8pt plus 2pt minus 2pt,
    segmentation style={solid, black!45, line width=1pt},
  }
}

\newtheoremstyle{mystyle}
  {15pt}
  {15pt}
  {\itshape}
  {2em}
  {\bfseries}
  {.}
  {.5em}
  {}
\newtheoremstyle{modern}%
  {12pt}%
  {12pt}%
  {\itshape}%
  {}%
  {\bfseries}%
  {.}%
  {.5em}%
  {\thmname{#1}\thmnumber{ #2}\thmnote{ (\normalfont\itshape\bfseries#3)}}%
\theoremstyle{modern}

\newtheorem{hypothesis}{Hypothesis}

\newtheorem{theorem}{Theorem}[section]
\newtheorem{lemma}[theorem]{Lemma}
\newtheorem{corollary}[theorem]{Corollary}
\newtheorem{proposition}[theorem]{Proposition}

\newtheorem{remark}[theorem]{Remark}
\newtheorem{definition}[theorem]{Definition}

\newtheorem{condition}[theorem]{Regularity condition}

\tcolorboxenvironment{hypothesis}{
    mytheoremstyle,
    colback=gray!5!white,
    colframe=gray!70!white
}
\tcolorboxenvironment{theorem}{
    mytheoremstyle,
    colback=blue!5!white,
    colframe=blue!70!white
}
\tcolorboxenvironment{lemma}{
    mytheoremstyle,
    colback=blue!5!white,
    colframe=blue!70!white
}
\tcolorboxenvironment{corollary}{
    mytheoremstyle,
    colback=gray!5!white,
    colframe=gray!70!white
}
\tcolorboxenvironment{proposition}{
    mytheoremstyle,
    colback=blue!5!white,
    colframe=blue!70!white
}
\tcolorboxenvironment{heuristic}{
    mytheoremstyle,
    colback=blue!5!white,
    colframe=blue!70!white
}
\tcolorboxenvironment{remark}{
    mytheoremstyle,
    colback=gray!5!white,
    colframe=gray!70!white
}
\tcolorboxenvironment{definition}{
    mytheoremstyle,
    colback=blue!5!white,
    colframe=blue!70!white
}
\tcolorboxenvironment{condition}{
    mytheoremstyle,
    colback=gray!5!white,
    colframe=gray!70!white
}
\tcolorboxenvironment{conjecture}{
    mytheoremstyle,
    colback=blue!5!white,
    colframe=blue!70!white
}

\newcommand{\sumbetak}{\sum_k\beta_k}
\newcommand{\sumbetaksmall}{\textstyle\sum_k\beta_k}
\newcommand{\sumbetashort}{{B}}
\newcommand{\sumbetakshort}{{B}}
\newcommand{\pPoIE}{p^{\textrm{PoIE}}}
\newcommand{\pk}{p^{\smash{(k)}}}
\newcommand{\pkt}{p^{\smash{(k)}}_t}
\newcommand{\pjt}{p^{\smash{(j)}}_t}
\newcommand{\pktdata}{p^{\smash{(k)}}_{t=\timeone}}
\newcommand{\knob}{{\lambda}}

\newcommand{\Langevin}{\textsc{Langevin}}

\newcommand{\NSR}{\textrm{NSR}}
\newcommand{\SNR}{\textrm{SNR}}
\newcommand{\PoE}{\textrm{PoE}}

\newcommand{\noiseparam}{{\xi}}
\newcommand{\driftparam}{{\alpha}}
\newcommand{\proposaldrift}{\mathbf{f}_t}

\newcommand{\tinit}{{t_{\textrm{init}}}}
\newcommand{\timezero}{{0}}
\newcommand{\timeone}{{1}}

\newcommand{\logweighingfunction}{{log-weight rate}}
\newcommand{\lineardiffusionmodel}{{linear diffusion model}}
\newcommand{\Lineardiffusionmodel}{{Linear diffusion model}}
\newcommand{\nonlineardiffusionmodel}{{nonlinear diffusion model}}
\newcommand{\Nonlineardiffusionmodel}{{Nonlinear diffusion model}}

\newcommand{\sqnorm}[1]{\left\lVert #1 \right\rVert^2_2}
\newcommand{\inlinesqnorm}[1]{\lVert #1 \rVert^2_2}

\newcommand{\LineBetweenStatementAndProof}{\vspace{-4pt}\tcbline\vspace{-7pt}}

\usepackage{hyperref}
\usepackage{url}
\usepackage{comment}
\usepackage{graphicx}
\usepackage{caption}
\usepackage{mathtools}

\title{Selecting Stable Sampling Dynamics for Compositional Diffusion: A Feynman--Kac  Weight Reduction}

\author{Sunjin Kim \\
        University of Arizona \\
        \texttt{sunjinkim@arizona.edu} \\
	\And
        Sungsoo Ahn \\
        KAIST \\
        \texttt{sungsoo.ahn@kaist.ac.kr}\\ 
        \And
        Michael Chertkov \\
        University of Arizona \\
        \texttt{chertkov@arizona.edu}
}

\begin{document}

\maketitle

\begin{abstract}
Given $K$ separately trained diffusion models, how should one sample from a $\beta_k$-weighted product of their terminal laws without jointly retraining the experts? Existing Feynman--Kac constructions provide many weighted stochastic differential equations (SDEs) with the correct target marginal. The consequential question is how to select useful proposal dynamics among them. We study an explicit family of weighted diffusions that realizes the same product-of-intermediate-experts trajectory. A scalar parameter $\knob$ controls how strongly the proposal drift follows the composed expert score, while an additional Langevin component can be varied independently. Writing $B=\sum_{k=1}^{K}\beta_k>0$, we identify the distinguished choice $\knob=1/B$. 
At this value, the aggregate squared-score term in the instantaneous Feynman--Kac log-weight rate cancels exactly. For positive expert exponents, the remaining score-disagreement contribution is non-positive. We call this structural cancellation \emph{Feynman--Kac reduction}. Under bounded-divergence assumptions, it yields an upper bound on the instantaneous unnormalized log-weight rate for all $\lambda\leq 1/B$. 
For linear diffusion protocols with compactly supported expert data, $\knob=1/B$ yields two-sided bounds at every nonterminal time, for exponents of either sign. A corresponding result holds for a single expert under the same divergence assumption. In these two-sided settings, the selected value is unique within the specified family in a universal worst-case sense. These are guarantees on the Feynman--Kac weight dynamics, not claims of minimum variance or globally optimal computational cost. Experiments on Gaussian mixtures and molecular sampling compare the analytically predicted stability with actual sampling quality, including instances in which parameter choices different from $\knob=1/B$ perform better empirically. The result is a training-free, distribution-agnostic principle with explicit limits on what this analytic weight-stability criterion implies for finite-budget sampling.
\end{abstract}

\begin{figure}[htbp]

    \centering
    \includegraphics[width=1.0\linewidth]{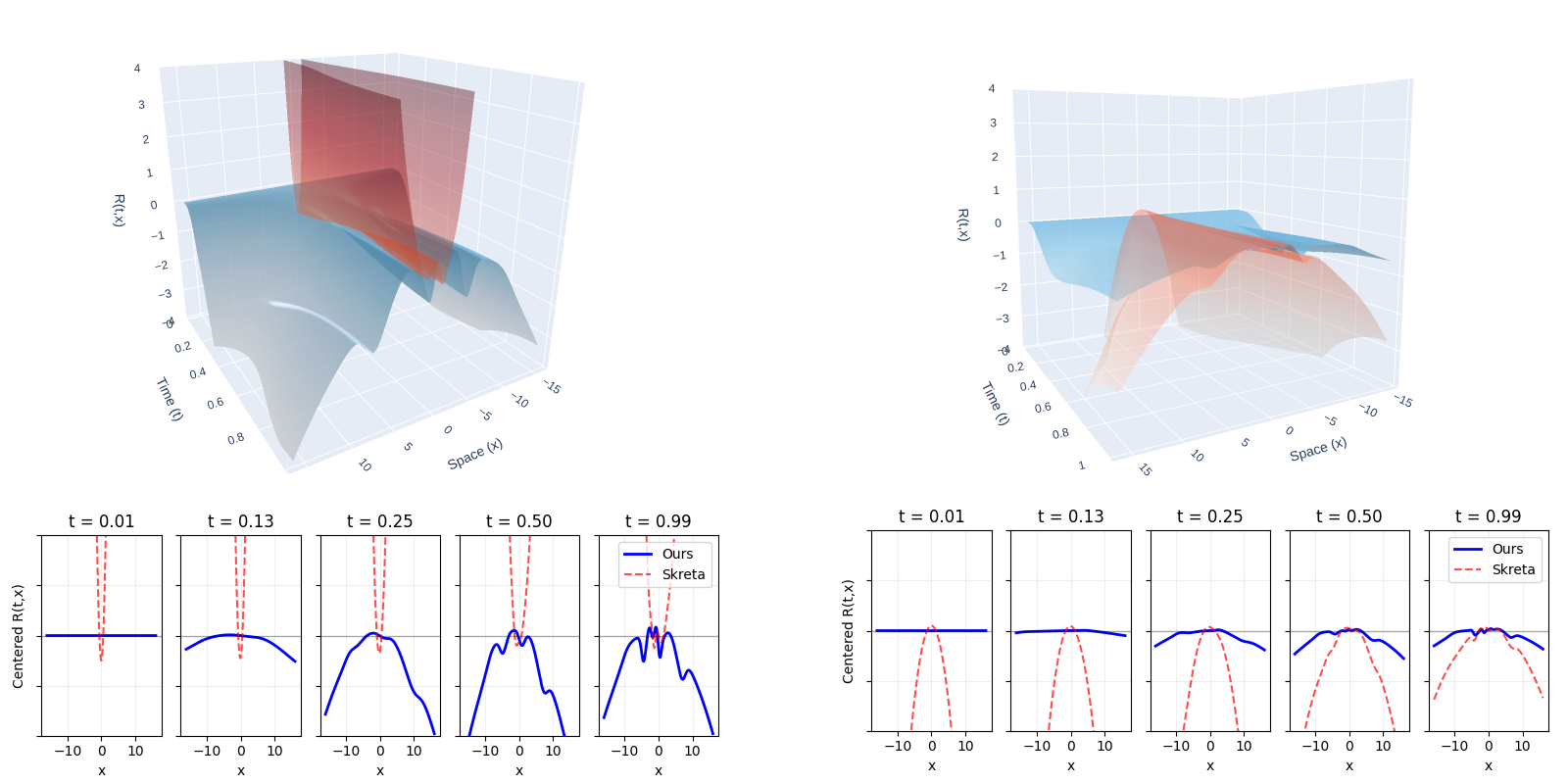} 
    \captionsetup{width=0.9\linewidth}
    \caption{
    A comparison of the \logweighingfunction{} from \cite{skreta_feynman-kac_2025} \(\knob=1\) (red) and that explored in this paper \(\knob=1/(\sumbetak)=1/B\) (blue). Two noteworthy aspects: 1. \(\knob=1/B\) is the \textit{largest} \(\knob\) which guarantees an upper-bounded \logweighingfunction{} (Prop.~\ref{prop:positive-exponents-implies-above-boundedness-MAIN}). Meanwhile, \(\knob=1\) may result in an upper-unbounded \logweighingfunction{} (when \(1>1/B\)) (left plot), or (when \(1<1/B\)) produces an upper-bounded \logweighingfunction{} with a larger curvature (right plot). This follows (in the \(\sumbetak=1\) setting) from Hölder's inequality. Second \(\knob=1/B\) is the only value for which the \logweighingfunction{} does not converge to a quadratic with \textit{infinite} curvature as \(t\downarrow0\) (Appendix.~\ref{sec:the-t-downarrow-zero-limit}).
    }
    \label{fig:paraboloid-bounded-above}
\end{figure}

\section{Introduction}\label{sec:Introduction}

\paragraph{The problem setting.} Consider $K$ pre-trained diffusion-model \cite{sohl-dickstein_deep_2015} experts on $\mathbb{R}^d$, each mapping a common easily sampled initial distribution $p_0$ (typically Gaussian) to \(K\) distinct terminal distributions $p_{\mathrm{data}}^{(k)}$. We consider the class
\begin{equation}\label{eq:nonlinear-diffusion-model-INTRO}
    d\mathbf{x}_t^{(k)}=\left[-\mathbf{F}_t\bigl(\mathbf{x}_t^{(k)}\bigr)
        \!+\!\noiseparam_t^2
        \nabla_{\mathbf{x}}\log p_t^{(k)}\bigl(\mathbf{x}_t^{(k)}\bigr)
    \right]dt
    \!+\!\noiseparam_t\,d\mathbf{W}_t^{(k)},
    \quad
    k=1,\ldots,K,\quad t\in[0,1],
\end{equation}
where $p_t^{(k)}$ denotes the marginal density of the $k$th expert, $p_0^{(k)}=p_0$, and $p_1^{(k)}=p_{\mathrm{data}}^{(k)}$. The base drift $\mathbf{F}_t$ and noise coefficient $\noiseparam_t$ are shared among the experts, whereas their score fields $\nabla_{\mathbf{x}}\log p_t^{(k)}$ differ. Equivalently, the experts share a common noising process but reverse toward different terminal laws.

Our terminal objective is the product-of-experts (PoE) distribution
\begin{equation}\label{eq:PoE-INTRO}
    p^{\PoE}(\mathbf{x})
    =
    \frac{1}{Z}
    \prod_{k=1}^{K}
    \bigl(p_{\mathrm{data}}^{(k)}(\mathbf{x})\bigr)^{\beta_k}.
\end{equation}

\paragraph{Overview of existing methods.}
A useful restriction on our setting leads to a large family of tractable \cite{he2026rne} solutions: One prescribes the corresponding \emph{product of intermediate experts} (PoIE) trajectory \footnote{Or time-varying exponent \(\beta_{k,t}\) generalization thereof \cite{lee2025collapse}.} connecting the common initial law to this terminal target,
\begin{equation}\label{eq:PoIE-INTRO}
    \pPoIE_t(\mathbf{x})
    =
    \frac{1}{Z_t}
    \prod_{k=1}^{K}
    \bigl(p_t^{(k)}(\mathbf{x})\bigr)^{\beta_k},
    \qquad t\in[0,1],
\end{equation}
so that $\pPoIE_{t=1}=p^{\PoE}$.
The intuitive reason for this giving rise to tractability is its instantaneous score is directly available,
\[
    \nabla_{\mathbf{x}}\log \pPoIE_t(\mathbf{x})
    =
    \sum_{k=1}^{K}
    \beta_k\nabla_{\mathbf{x}}\log p_t^{(k)}(\mathbf{x}).
\]

However, simply inserting this composed score into the diffusion model drift
does \emph{not}, in general, transport the prescribed PoIE marginals
\cite{du2023reduce}. Exact realization can instead be achieved with a weighted particle system \cite{skreta_feynman-kac_2025},
\begin{equation}\label{eq:WSODE-Intro}
    \begin{cases}
        d\mathbf{x}_t=\mathbf{f}_t(\mathbf{x}_t)\,dt+g_t\,d\mathbf{W}_t,\\[1mm]
        d\log w_t=h_t(\mathbf{x}_t)\,dt,
    \end{cases}
\end{equation}
whose weighted particle density follows
$\pPoIE_t$ and therefore terminates at the desired PoE distribution. 

Existing result (\cite{he2026rne}, Appendix.~\ref{sec:TRACTABLE-which-WSDEs-sampling-from-PoIE-of-nonlinear-diffusion-models-ARE-TRACTABLE}) shows that almost \emph{any} choice of $(\mathbf{f}_t,g_t)$ admits a corresponding \emph{computationally tractable} Feynman--Kac corrector $h^{\mathbf{f}_t,g_t}_t$ which results in the PoIE trajectory. 
The problem addressed here is therefore one of \emph{selection}, 
i.e. optimization over the space \((\mathbf{f}_t,g_t)\). The (non-unique) optimum solution  \((\mathbf{f}^*_t,g^*_t)\) is  that which results in a flat (constant with respect to \(\mathbf{x}_t\)) \logweighingfunction{} \(h^{\mathbf{f}^*,g^*}_t(\mathbf{x}_t)=c_t\) \cite[Thm.~8.3.1]{ambrosio2005gradient}---but it is in general intractable (e.g., \cite[Sec.~3.2]{ren2026driftlite}). Previous works exploring this space rely on empirical trial-and-error optimization of parameters \cite{ren2026driftlite,he2026rne}. \textbf{Our core contribution lies in being the first analytic exploration of this search space}: We show a particular set of \(\mathbf{f}_t\) and \(g_t\) for which \(h^{\mathbf{f},g}_t\) is upper bounded \(\sup_{\mathbf{x}} h^{\mathbf{f},g}_t(\mathbf{x})<\infty\) or upper\(+\)lower bounded \(\sup_{\mathbf{x}} |h^{\mathbf{f},g}_t(\mathbf{x})|<\infty\), with \emph{little or no assumptions} on the expert laws \(\pkt(\mathbf{x}), t\in[0,1]\).

\paragraph{Overview of derivations}
We examine the two-dimensional family of proposal distributions parameterized by \(({\color{red}\knob},g_t^2)\)
\begin{align}
    \label{eq:1-knob-family-in-short-INTRO}
		d\mathbf{x}_t 
        &= \left[- \mathbf{F}_t(\mathbf{x}_t)+ \left(\red{\knob}\tfrac{\noiseparam^2_t}{2}+\tfrac{g^2_t}{2} \right)\nabla_{\mathbf{x}} \log \pPoIE_t(\mathbf{x}_t)\right] \, dt 
        +g_t \, d\mathbf{W}_t
\end{align}
which (along with the target trajectory Eq.~\ref{eq:PoIE-INTRO}) fixes the required importance \logweighingfunction{}
\begin{subequations}
\label{eq:1-knob-family-logweighingfunction-in-short-INTRO}
\begin{align}
        \label{eq:squared-score-term-INTRO}
        h^{\red{\knob}}_t(\mathbf{x}_t)
        &=\tfrac{\noiseparam^2_t}{2}(\red{\knob}-\tfrac1B) \sqnorm{\nabla_{\mathbf{x}} \log \pPoIE_t}
        &(\text{squared-score)}
        \\&\quad {} 
        \label{eq:drift-disagreement-term-INTRO}
        -\tfrac{\noiseparam^2_t}{4\sumbetashort}\sum_{j,k}  \beta_j \beta_k \inlinesqnorm{\nabla_{\mathbf{x}} \log p^{j}_t - \nabla_{\mathbf{x}} \log \pkt} 
        &(\text{drift-disagreement)}
        \\&\quad {}  
        \label{eq:divergence-terms-INTRO}
        + \tfrac{\noiseparam^2_t}{2}\left(-1+\red{\knob}\right) \sum_k \beta_k \nabla_{\mathbf{x}} \cdot \nabla_{\mathbf{x}} \log \pkt
        +(B-1)\nabla_{\mathbf{x}} \cdot \mathbf{F}_t 
        &(\text{divergence terms)}
\end{align}
\end{subequations}
where \(B\doteq\textstyle\sum_k\beta_k\). We give descriptions on the behaviors of the two parameters. First, the parameter \(g_t^2\) does not influence the \logweighingfunction{}. It can hence be seen as implementing a Langevin "gauge". Furthermore, as one takes \(g_t^2\) to \(\infty\), one can expect perfect sampling (Prop.~\ref{prop:CONJECTURE-Langevin-Infinity-MAIN})---but only at the cost of necessitating an increasingly fine Euler--Maruyama step size. Our main results revolve around the remaining knob \({\color{red} \knob}\). When set to the value of \({\color{red} \knob=1/B}\), the squared-score term Eq.~\ref{eq:squared-score-term-INTRO} vanishes leaving the divergence terms Eq.~\ref{eq:divergence-terms-INTRO} 
and the drift-disagreement term Eq.~\ref{eq:drift-disagreement-term-INTRO}. 

\paragraph{Main results}
We assume the divergence terms to be bounded 
    \footnote{Except in the compact setting Prop.~\ref{prop:compact-support-implies-boundedness-MAIN}, where boundedness follows from the compactness and linearity.} 
(RC.~\ref{cond:bounded-score-divergence-MAIN},~\ref{cond:bounded-shared-nonlinear-drift-divergence-MAIN}, Rem.~\ref{rem:score-smooth-probably-holds-for-linear-diffusion-models},~\ref{rem:bounded-F-div-holds-at-all-timesteps-for-linear-diffusion-models}), leaving the behavior of the drift-disagreement term to directly lead to our three main results: First, when all exponents are positive, the \logweighingfunction{} is upper bounded (Prop.~\ref{prop:positive-exponents-implies-above-boundedness-MAIN}). Second, when all target distributions are compact and diffusion models are linear (\(\mathbf{F}_t(\mathbf{x}_t)=\alpha_t \mathbf{x}_t\)), then by Tweedie's identities the \logweighingfunction{} is upper\(+\)lower bounded (Prop.~\ref{prop:compact-support-implies-boundedness-MAIN}). Third, in the particular setting of the annealing of a single diffusion model (\(K=1\)), we again have an upper\(+\)lower bounded \logweighingfunction{}  (Prop.~\ref{prop:one-model-implies-boundedness-MAIN}). All results come with uniqueness (Prop.~\ref{prop:compact-support-implies-boundedness-MAIN},~\ref{prop:one-model-implies-boundedness-MAIN}) or quasi-uniqueness (Corr.~\ref{corr:positive-exponents-result-uniquness-K-param-family}) within an extended \((K+1)\)-parameter family.

\paragraph{Interpretation of restricted setting.}
Recall that following previous works \cite{du2023reduce,skreta_feynman-kac_2025,he2026rne} we restrict our scope to diffusion models (Eq.~\ref{eq:nonlinear-diffusion-model-INTRO}) and the PoIE trajectory (Eq.~\ref{eq:PoIE-INTRO}). This may seem to be a somewhat contrived setting, but it is due to this confinement that we obtain the drift-disagreement term Eq.~\ref{eq:drift-disagreement-term-INTRO} which is key to our results.

\paragraph{Limitations.} Our bounds are instantaneous, i.e., they hold at a fixed time $t$, and in general need not be uniform or integrable over the full interval $t\in[0,1]$. This issue is particularly relevant near $t=1$, where the expert marginals may become increasingly non-smooth and the constants entering the bounds may diverge. Moreover, boundedness of the Feynman--Kac log-weight rate is not an optimality statement: it does not by itself imply the best finite-budget sampling performance according to metrics such as effective sample size, Wasserstein error, or estimator variance. We therefore complement the analytical results with numerical experiments that compare the predicted weight stability with actual sampling quality, including examples for which $\knob=1/B$ is not the empirically optimal choice.

\section{Background: Weighted stochastic differential equations (WSDEs)}\label{sec:preliminaries-WSDEs-MAIN}

\subsection{Fokker--Planck PDE: An evolution of distributions resulting from a WSDE}
WSDEs, more formally Feynman--Kac particle systems, generalize SDEs by tracking not only the particle's location (\(\mathbf{x}_t\)) but also its changing weight (\(w_t\))
\begin{align}
    \begin{cases}
    \label{eq:wSODE-definition-MAIN}
    d\mathbf{x}_t &= \mathbf{f}_t(\mathbf{x}_t) \, dt + g_t \, d\mathbf{W}_t \\
    d \log w_t &= h_t(\mathbf{x}_t) \, dt + \mathbf{j}_t(\mathbf{x}_t)\cdot d\mathbf{W}_t.
    \end{cases}
\end{align}
Equivalent population representations implement the Feynman--Kac potential through branching and killing (or ``cloning'') of particles, a construction widely used in statistical-physics approaches to rare-event sampling \cite{giardina2006direct,giardina2011simulating}.
The distribution of the weight \footnote{More formally \(\rho_t(\mathbf x)\doteq\mathbb E[w_t\,\delta(\mathbf x_t-\mathbf x)]\).} (and not simply the distribution of particles, as in \textit{unweighted} SDEs/ODEs) follows the Fokker--Planck PDE
\begin{equation}
    \partial_t \rho_t(\mathbf{x}) = -\nabla_{\mathbf{x}}\cdot(\rho_t \mathbf{f}_t) + \tfrac{g^2_t}{2} \Delta_{\mathbf{x}} \rho_t + (h_t+\tfrac12\sqnorm{\mathbf{j}_t}-g_t\nabla_{\mathbf{x}}\log\rho_t\cdot \mathbf{j}_t - g_t \nabla_{\mathbf{x}} \cdot \mathbf{j}_t)\rho_t.
    \label{eq:Fokker--Planck-for-WSDEs-MAIN}
\end{equation}
\(\rho_t\) is in general unnormalized \(\int \rho_t(\mathbf{x})\,d\mathbf{x}\neq1\). 

\subsection{Using IS/SMC to sample from a WSDE}
One can approximately sample from these marginals (after normalizing \(\tilde\rho_t(\mathbf{x}) \doteq \rho_t(\mathbf{x})/\int \rho_t(\mathbf{y})\, d\mathbf{y}\)) using algorithms such as importance sampling (IS) or Sequential Monte Carlo (SMC). These algorithms both consist of simultaneously integrating (in Euler--Maruyama fashion) \(N\) particles \((x^{(n)}_t,w^{(n)}_t)_{n=1,\dots,N}\) according to Eq.~\ref{eq:wSODE-definition-MAIN}, then at certain (either deterministic or stochastic) time steps \(t_i\) resampling the particles according to the normalized weights \(W^{(n)}_{t}\doteq w^{(n)}_t/(\sum^{N}_{n=1} w^{(n)}_t)\). Resampling here refers to for example multinomial sampling using the normalized weights as probabilities. If one resamples only once at terminal time \(t=1\), it is referred to as IS, otherwise it is referred to as SMC.

\subsection{Optimizing IS/SMC for efficient sampling}
\cite{he2026rne} showed a large family of WSDEs \((\mathbf{f}_t,g_t,h_t,\mathbf{j}_t)\) which sample from the desired product-of-experts target marginal Eq.~\ref{eq:PoE-MAIN}. Our goal then becomes optimizing over this space to find a WSDEs which sample efficiently. We now define efficiency in the context of our problem.

In importance sampling or sequential Monte Carlo settings, the main goal one pursues is designing a proposal distribution which is as similar as possible to the target distribution, as this leads to more computationally efficient sampling, or more precisely, requiring less number of particles \(N\) to achieve the same sampling fidelity. 

The optimal solution is that of the proposal being equal to the target. In this case one may disregard the weights, i.e., there is no need to perform resampling with respect to the weights. In our context, the ideal WSDE is that with constant weights---where \(h_t\) and \(\mathbf{j}_t\) are both zero functions. Such an ideal (unweighted) SDEs and ODEs always exist. For example, for any sufficiently regular probability trajectory \(\{p_t\}_{t\in[0,1]}\), there exists a vector field \(\mathbf{f}_t\) such that \(d\mathbf{x}_t=\mathbf{f}_t(\mathbf{x}_t)\,dt,\ w_t=1\) results in the particles being distributed according to the trajectory \cite[Thm.~8.3.1]{ambrosio2005gradient}.
But calculating such optimal vector fields is in general intractable \cite[Sec.~3.2]{ren2026driftlite}, and one can only hope to make the proposal distribution as similar to the target as possible. 

The most commonly used measure of discrepancy between the proposal and target in IS/SMC is effective-sample-size (ESS) \(1/(\sum_n (W^{(n)}_t)^2)\). The ESS can be considered the empirical version of the reciprocal of the normalized weight variance \(\text{ESS}/N\approx\mathbb E[w_t]^2/\mathbb E[w^2_t]=1/(1+\mathbb V[w_t]/\mathbb E[w_t]^2)\) and equivalently the Chi-squared divergence between the target and proposal distributions (see \cite[Sec.~8.6]{chopin2020introduction} for an overview of these quantities). In particular, a zero weight variance \(\mathbb V[w_t]=0\) immediately implies equivalence of target and proposal and the maximum value of \(ESS=N\). 

\section{Preliminary lemmas}
\subsection{Langevin gauge}\label{sec:Langevin-gauge}
In this paper we discuss only WSDEs Eq.~\ref{eq:wSODE-definition-MAIN} with no directional weight \(\mathbf{j}_t(\mathbf{x}_t)=0\). Such WSDEs have a \emph{Langevin gauge} 
    \footnote{This precisely holds in the (unweighted) SDE regime as well---it is the basis of the well-known equivalence of \lineardiffusionmodel{}s and flow-matching models (see e.g. \cite{gao2025diffusionsamecoin}).} 
which allows further simplification of the search space. We define the Langevin gauge as the following.
\begin{lemma}\label{prop:langevin-gauge-MAIN}
Say that the following weighted (non-stochastic) differential equation
\begin{align}
    \label{eq:weighted-NONSTOCHASTIC-DE-MAIN}
    d\mathbf{x}_t 
    &= \mathbf{f}_t(\mathbf{x}_t) \, dt\\
    d \log w_t &= h_t(\mathbf{x}_t) \, dt
\end{align}
with \(\mathbf{x}_0\sim\rho_0(\cdot)\) results in a trajectory of weighted particles \(\{\rho_t(\cdot)\}_{t\in[0,1]}\). Then, for any choice of Langevin strength \(g_t\geq0\), the WSDE
\begin{align}
    \begin{cases}
    \label{eq:wSODE-with-zero-directional-weight-MAIN}
    d\mathbf{x}_t 
    &= \mathbf{f}_t(\mathbf{x}_t) \, dt + [\tfrac{g^2_t}{2}\nabla_{\mathbf{x}} \log \rho_t(\mathbf{x}_t)\,dt + g_t \, d\mathbf{W}_t]
    \\
    d \log w_t &= h_t(\mathbf{x}_t) \, dt
    \end{cases}
\end{align}
also results in the same mass trajectory \(\rho_t\).
\LineBetweenStatementAndProof
\begin{proof}
    This can be seen directly from the Fokker--Planck Eq.~\ref{eq:Fokker--Planck-for-WSDEs-MAIN}.
\end{proof}
\end{lemma}
Hence we introduce the short-hand 
\begin{align}\label{eq:Langevin-shorthand-definition}
    \Langevin_{\rho_t,g_t} \doteq [\tfrac{g^2_t}{2}\nabla_{\mathbf{x}} \log \rho_t(\mathbf{x}_t)\,dt + g_t \, d\mathbf{W}_t].
\end{align}

A discrete-time algorithmic analogy of Lemma.~\ref{prop:langevin-gauge-MAIN} would be the fact that running several (say, unadjusted Langevin) MCMC steps after each SMC step does not change the obtained mass distribution (this is the mechanism behind resample-move SMC algorithms \cite[Sec.~4.4]{doucet2009tutorial}).

This lemma, combined with the following proposition, simplifies our search space.

\subsection{Infinite Langevin trivially leads to constant weights}\label{sec:infinite-Langevin}
Furthermore, we show that when one takes the Langevin strength to infinity \(g_t\to\infty\) in Eq.~\ref{eq:wSODE-with-zero-directional-weight-MAIN}, the weights trivially become constant, and the unweighted particles become samples from the target.

\begin{proposition}[Weights become constant as we take the Langevin strength to infinity]\label{prop:CONJECTURE-Langevin-Infinity-MAIN}
    As we take the Langevin strength \(g_t\) to infinity in Eq.~\ref{eq:wSODE-with-zero-directional-weight-MAIN}, the weights converge to a constant, \(w_t\to Z_t/Z_0\), and the distribution of the unweighted particles converges to the normalised target \(\bar\rho_t\) (\(Z_t\) and \(\bar\rho_t\) are defined below).
    \LineBetweenStatementAndProof
    \begin{proof}
        \textbf{Intuition}: Running an arbitrarily large or small number of MCMC steps after a correct SMC step preserves desired marginals.
    
        \textbf{Heuristic argument}: As one takes the Langevin strength \(g_t\) to infinity, \(\mathbf{x}_t\) and \(\mathbf{x}_{t+\Delta t}\) will approach independent samples from \(\bar\rho_t\)  and \(\bar\rho_{t+\Delta t}\). The first consequence of this is that the distribution of unweighted particles (in importance sampling terminology the proposal distribution) approaches the target distribution. The second consequence is that, due to independence, the rate of change of log weight at times \(\mathbf{x}_t\) and \(\mathbf{x}_{t+\Delta t}\) will be nearly independent. Hence the accumulated weight at any time \(t\) converges to a constant:
        \begin{align}
            \log w_t &= 
            \int^t_{s=0} h_s(\mathbf{x}_s) \,ds \xrightarrow{g_t\to\infty} \int^t_{s=0} \mathbb E_{\mathbf{x}_s\sim \bar\rho_s} [h_s(\mathbf{x}_s)] \,ds = \log \frac{Z_t}{Z_0}
            \\
            \text{ where } Z_s&\doteq \int\rho_s(\mathbf{x})\,d\mathbf{x} \quad \text{and}\quad \bar\rho_s\doteq \frac{\rho_s}{Z_s}
        \end{align}
        where last equality is due to Eq.~\ref{eq:rate-of-change-of-normalization-constant-for-general-wsdes}.

        A \textbf{formal proof} is given in App.~\ref{sec:langevin-finite-action} (Thm.~\ref{thm:finite-action} and Cor.~\ref{cor:finite-action-consequences}).
    \end{proof}
\end{proposition}

\paragraph{Precedence of Prop.~\ref{prop:CONJECTURE-Langevin-Infinity-MAIN}}
For annealed Langevin dynamics ($\mathbf f_t\equiv0$), this limit was analysed by \cite{guo2025provable,guo2026complexity}. Our result (Thm.~\ref{thm:finite-action}) generalises theirs, \cite[Thm.~2]{guo2026complexity} for the weights and \cite[Thm.~1]{guo2025provable} for the particles, to an arbitrary drift $\mathbf f_t$, using the same proof method.

\paragraph{Implications of Prop.~\ref{prop:CONJECTURE-Langevin-Infinity-MAIN}}
Prop.~\ref{prop:CONJECTURE-Langevin-Infinity-MAIN} in turn leads to reaching the maximum possible effective-sample-size (Cor.~\ref{cor:finite-action-consequences}), and a proposal distribution (distribution of unweighted particles) which is equal to the target distribution (Thm.~\ref{thm:finite-action}). In other words, one achieves perfectly sample-efficient sampling. However, this only arrives at the computational cost of requiring an increasingly fine Euler--Maruyama integration step (Rem.~\ref{rem:caveat-require-fine-step-size})---it is this that motivates one to optimize over the space of the drift \(f\) as well, which leads to our main results.

\paragraph{Further remarks on Prop.~\ref{prop:CONJECTURE-Langevin-Infinity-MAIN}}
Prop.~\ref{prop:CONJECTURE-Langevin-Infinity-MAIN} has analogues in both statistical physics and SMC. In the former, it corresponds to the quasi-static (slow-switching) limit of Jarzynski's equality \cite{jarzynski_equilibrium_1997}, in which the weights become deterministic. In the latter, it can be interpreted as the limit of a resample-move \cite[Sec.~4.4]{doucet2009tutorial} algorithm when one takes the number of MCMC 'move' steps to infinity---in particular, as \(g_t\to\infty\), we approach the annealed Langevin dynamics seen in \cite{du2023reduce}. 

\paragraph{A practical heuristic}
We also anticipate that under many practical settings, this relationship is roughly monotonic---i.e. a higher \(g_t\) often, but not always, results in lower weight variance. We find this often, but not always, holds in our experiments Sec.~\ref{sec:experiments}. However, a higher Langevin parameter \(g_t\) leads to lower computational efficiency as it necessitates the use of a finer (Euler--Maruyama) time step size, which motivates us to optimize over the space of \(\mathbf{f}_t\) as well---this is our primary objective in this paper.

\section{Problem setting}

\subsection{Diffusion models}\label{sec:nonlinear-diffusion-model-MAIN}
Say we have \(K\) diffusion models defined on \(\mathbb R^d\) 
\begin{equation}\label{eq:nonlinear-diffusion-model-MAIN}
    d\mathbf{x}_t = [-\mathbf{F}_t(\mathbf{x}_t) + \tfrac{\xi_t^2}{2} \nabla_{\mathbf{x}} \log p^{\smash{(k)}}_t(\mathbf{x}_t)]\, dt + \Langevin_{\pkt,g_t}, \quad k=1,\dots,K
\end{equation}
where \(\mathbf{F}_t\) and \(\noiseparam_t\) are shared among the \(K\) models. Each model transports a standard Gaussian
\(\pk_{t=\timezero}=\mathcal N(0,I)\)
to its target \(\pktdata=\pk_{data}\)
along a trajectory of marginals \(\pkt, t\in[0,1]\), the score function \(\nabla_{\mathbf{x}} \log p^{\smash{(k)}}_t(\cdot)\) of which is available. 

Most real-world implementations of diffusion models are \lineardiffusionmodel{}s
\begin{equation}\label{eq:linear-diffusion-model-short-MAIN}
    \mathbf{F}_t(\mathbf{x}_t)=\driftparam_t \mathbf{x}_t,
\end{equation}
since there exists an efficient algorithm to learn the score function \(\nabla_{\mathbf{x}} \log p^{\smash{(k)}}_t(\cdot)\) of such models. Hence, at least at the time of writing, the more general setting \(\mathbf{F}_t(\cdot)\) has more theoretical than practical interest. \footnote{In our setting, such models also have the advantage of being able to be re-parameterized in the space-time \((x,t)\) axes to share the parameters \((\driftparam_t,\noiseparam_t)\) (Eq.~\ref{eq:formula-reparameterizing-linear-diffusion-models})---a condition which we assume throughout this paper. This condition is not strictly necessary to derive computationally tractable sampling methods \cite{lee2025collapse}, but our discussion below benefits from the simplicity and regularity which this provides.}

\subsection{Aim: Product-of-experts}Consider \(K\) \nonlineardiffusionmodel{}s (Eq.~\ref{eq:nonlinear-diffusion-model-MAIN}).
We aim to sample from their \textbf{product-of-experts} (PoE) distribution
\begin{align}\label{eq:PoE-MAIN}
    p^{\PoE} \doteq \frac1Z \prod^K_{k=1} (\pk_{data})^{\beta_k}
\end{align}
for exponents with a positive sum \(\sumbetaksmall>0\). Note that negative exponents are in general allowed (e.g. Prop.~\ref{prop:compact-support-implies-boundedness-MAIN}) unless specified otherwise (e.g. Prop.~\ref{prop:positive-exponents-implies-above-boundedness-MAIN}) \footnote{We assume throughout that \(p^{\PoE}\) and furthermore the PoIE trajectory (Eq.~\ref{eq:PoIE-trajectory-MAIN}) is normalizable at all \(t\in[0,1]\). See \cite{lee2025collapse} for discussion of normalizability of the PoIE trajectory.}. 

\subsection{Particular design choice: Product-of-intermediate-experts}\label{sec:poie-trajectory-MAIN}
In order to do so, we follow the particular design choice of sampling from the \textbf{product-of-intermediate-experts} (PoIE) trajectory \cite{du2023reduce,skreta_feynman-kac_2025} where the marginals at non-terminal times \(t<1\) are likewise multiplied 
\begin{align}
\label{eq:PoIE-trajectory-MAIN}
   \pPoIE_{t}
    &\doteq
    \frac{1}{Z_t}\prod^K_{k=1} (\pkt)^{\beta_k}
    , \quad{t\in[\timezero,\timeone]},
\end{align}
resulting in a trajectory with the boundaries
\begin{align}
    \pPoIE_{t=\timezero}&=\mathcal N\left(0,\frac1{\sumbetaksmall}\cdot I\right),
    \\
    \pPoIE_{t=\timeone}&=p^{\PoE}.
\end{align}

\paragraph{Why this trajectory?}
All current computationally tractable methods \cite{du2023reduce,skreta_feynman-kac_2025,he2026rne,lee2025collapse} sampling from the product-of-experts distribution (Eq.~\ref{eq:PoE-MAIN}) of \(K\) diffusion models (Eq.~\ref{eq:nonlinear-diffusion-model-MAIN}) follow this trajectory. 
This trajectory results in a large set of computationally tractable WSDEs: for almost\footnote{See Appendix.~\ref{sec:TRACTABLE-which-WSDEs-sampling-from-PoIE-of-nonlinear-diffusion-models-ARE-TRACTABLE} on why \(g_t=0\) results in the \logweighingfunction{} being less amenable to computation in general.} any \((\mathbf{f}_t,g_t,\mathbf{j}_t)\), 
there exists a computationally tractable \cite{he2026rne} \(h^{\mathbf{f}_t,g_t,\mathbf{j}_t}_t\) such that the WSDE Eq.~\ref{eq:wSODE-definition-MAIN} samples from the PoE target (Prop.~\ref{prop:all-WSDEs-sampling-from-PoIE-of-nonlinear-diffusion-models}, Appendix.~\ref{sec:TRACTABLE-which-WSDEs-sampling-from-PoIE-of-nonlinear-diffusion-models-ARE-TRACTABLE}~\footnote{\cite{he2026rne} derives this by considering the Radon--Nikodym derivative of path measures. We show (Prop.~\ref{prop:all-WSDEs-sampling-from-PoIE-of-nonlinear-diffusion-models}) another derivation using only the Fokker--Planck PDE (Eq.~\ref{eq:Fokker--Planck-for-WSDEs-MAIN}).}).

Within the context of this large diversity of options, our paper's main contribution lies in finding a WSDE within this family which a-priori gives an upper bound or both an upper and lower bounded \logweighingfunction{}.

\section{Main results: Set-up}

\subsection{The \((1+1)\)-parameter family}

We exposit our main results in context of a \((1+1)\)-parameter family of WSDEs (parameterized by \((\red{\knob},g_t)\)) which all follow the product-of-intermediate-experts trajectory (Eq.~\ref{eq:PoIE-trajectory-MAIN}). The purpose of this narrowing of the scope is two-fold: first, clarity and second, due to this family's correspondence to an accompanying family of backward-in-time noising-like processes (Sec.~\ref{sec:noising-process-intuition-MAIN}) which allows an elegant interpretation of the results. The \((1+1)\)-parameter family is the following \footnote{The shared drift \(-\mathbf{F}_t\) is deliberately kept fixed --- for \lineardiffusionmodel{}s in particular, this choice makes our analysis invariant to the noising schedule; see Remark~\ref{rem:why-fix-shared-drift}.}:
\begin{align}
	&\begin{cases}\label{eq:1-knob-family-in-short-MAIN}
		d\mathbf{x}_t 
        &= \left[- \mathbf{F}_t(\mathbf{x}_t)+ \red{\knob}\tfrac{\noiseparam^2_t}{2}\nabla_{\mathbf{x}} \log \pPoIE_t(\mathbf{x}_t)\right] \, dt + \Langevin_{\pPoIE_t,g_t} \\
        &= \left[- \mathbf{F}_t(\mathbf{x}_t)+ \red{\knob}\tfrac{\noiseparam^2_t}{2}\nabla_{\mathbf{x}} \log \pPoIE_t(\mathbf{x}_t)\right] \, dt + \left[\frac{g^2_t}{2} \nabla_{\mathbf{x}} \log \pPoIE_t(\mathbf{x}_t) \, dt + g_t \, d\mathbf{W}_t\right] \\
		d \log w_t &= h^{\red{\knob}}_t(\mathbf{x}_t) \, dt \\
        \mathbf{x}_\timezero &\sim \pPoIE_{t=\timezero} = \mathcal N\big(0,(1/{\textstyle\sum_k\beta_k})\cdot I\big) \\
		w_\timezero &= 1
	\end{cases}
\end{align}
where \(\nabla_{\mathbf{x}} \log \pPoIE_t(\mathbf{x}_t)=\sum_k \beta_k\nabla_{\mathbf{x}} \log \pkt(\mathbf{x}_t)\) by definition, 
the \(\Langevin_{\pPoIE_t,g_t}\) term is defined as in Sec.~\ref{sec:Langevin-gauge}, 
and the \logweighingfunction{} \(h^{\knob}_t\) is of the form
\begin{tcolorbox}[colback=white, colframe=black, boxsep=0pt, left=-15pt, right=6pt, top=-4pt, bottom=4pt, arc=0pt, outer arc=0pt, boxrule=1pt]
\begin{subequations}
\label{eq:1-knob-family-logweighingfunction-in-short-MAIN}
\begin{align}
        \label{eq:squared-score-term-MAIN}
        h^{\red{\knob}}_t(\mathbf{x}_t)
        &=\tfrac{\noiseparam^2_t}{2}(\red{\knob}-\tfrac1B) \sqnorm{\nabla_{\mathbf{x}} \log \pPoIE_t}
        &(\text{squared-score term)}
        \\&\quad {} 
        \label{eq:drift-disagreement-term-MAIN}
        -\tfrac{\noiseparam^2_t}{4\sumbetashort}\sum_{j,k}  \beta_j \beta_k \inlinesqnorm{\nabla_{\mathbf{x}} \log p^{j}_t - \nabla_{\mathbf{x}} \log \pkt} 
        &(\text{drift-disagreement term)}
        \\&\quad {}  
        \label{eq:score-divergence-term-MAIN}
        + \tfrac{\noiseparam^2_t}{2}\left(-1+\red{\knob}\right) \sum_k \beta_k \nabla_{\mathbf{x}} \cdot \nabla_{\mathbf{x}} \log \pkt
        &(\text{score divergence term)}
        \\&\quad {}
        \label{eq:shared-nonlinear-drift-divergence-term-MAIN}
        +(B-1)\nabla_{\mathbf{x}} \cdot \mathbf{F}_t 
        &(\text{divergence of shared drift)}
    \\
    \nonumber
    \text{ where }&B\doteq\textstyle\sum_k\beta_k.
\end{align}
\end{subequations}
\end{tcolorbox}

\paragraph{The \((K+1)\)-parameter family}

We also briefly discuss the \((K+1)\)-parameter family, where each score can be adjusted separately 
\begin{align}
	&\begin{cases}\label{eq:K-knob-family-in-short-MAIN}
		d\mathbf{x}_t &= \left[- \mathbf{F}_t(\mathbf{x}_t)+ \tfrac{\noiseparam^2_t}{2}\sum_k \red{\knob_k}\beta_k\nabla_{\mathbf{x}} \log \pkt(\mathbf{x}_t)\right] \, dt + \Langevin_{\pPoIE_t,g_t} \\
		d \log w_t &= h^{\red{\{\knob_k\}_k}}_t(\mathbf{x}_t) \, dt
	\end{cases}
\end{align}
with same initial conditions as Eq.~\ref{eq:1-knob-family-in-short-MAIN}.
We show we can still claim some uniqueness results (Corr.~\ref{corr:positive-exponents-result-uniquness-K-param-family}, Prop.~\ref{prop:compact-support-implies-boundedness-MAIN},~\ref{prop:one-model-implies-boundedness-MAIN}) within this larger family.

\subsection{Regularity conditions}\label{sec:regularity-conditions-MAIN}

We give two regularity conditions, RC.~\ref{cond:bounded-score-divergence-MAIN} and RC.~\ref{cond:bounded-shared-nonlinear-drift-divergence-MAIN}. When using \lineardiffusionmodel{}, one of them (RC.~\ref{cond:bounded-shared-nonlinear-drift-divergence-MAIN}) trivially holds at all time \(t\in[0,1]\) (Remark.~\ref{rem:bounded-F-div-holds-at-all-timesteps-for-linear-diffusion-models}, as noted in \cite{skreta_feynman-kac_2025}), and the other (RC.~\ref{cond:bounded-score-divergence-MAIN}) can be expected to hold at all non-terminal time \(t\in[0,1)\) for non-pathological data distributions (Remark.~\ref{rem:score-smooth-probably-holds-for-linear-diffusion-models}).

\begin{condition}[Bounded score divergence at a given time 
    \footnote{This regularity condition bears similarity to other regularity conditions in the Monte Carlo literature. For example it is implied by the Lipschitz-smoothness regularity condition used in e.g. \cite[Condition.L1]{durmus2017nonasymptotic}.}]
    \label{cond:bounded-score-divergence-MAIN}
    All \(k\) models have bounded score divergence 
    \begin{equation}
        \sup_{\mathbf{x}\in\mathbb R^d} \left|\nabla_{\mathbf{x}} \cdot \nabla_{\mathbf{x}} \log \pkt(\mathbf{x})\right| < \infty.
    \end{equation}
\end{condition}

\begin{remark}[RC.~\ref{cond:bounded-score-divergence-MAIN} most likely will hold for all \(t\in[0,1)\) when using \lineardiffusionmodel{}]\label{rem:score-smooth-probably-holds-for-linear-diffusion-models}
    We argue that when using \lineardiffusionmodel{}s (Eq.~\ref{eq:linear-diffusion-model-short-MAIN}), regularity condition.~\ref{cond:bounded-score-divergence-MAIN} can be reasonably expected to hold at all \(t\in[0,1)\), except for highly pathological target distributions \(\pktdata(\cdot)\) (see Appendix.~\ref{sec:regularity-condition-bounded-score-divergence} for justification). However, it may not hold at \(t=1\) even for commonly encountered target distributions, e.g. compact distributions (Sec.~\ref{sec:t-uparrow-one-case-MAIN} for consequences).
\end{remark}

\begin{condition}[Bounded shared drift divergence at a given time]\label{cond:bounded-shared-nonlinear-drift-divergence-MAIN}
    The vector field \(\mathbf{F}_t(\mathbf{x}_t)\) shared among \(k\) models (Eq.~\ref{eq:nonlinear-diffusion-model-MAIN}) have bounded divergence
    \begin{equation}
        \sup_{\mathbf{x}\in\mathbb R^d} |\nabla_{\mathbf{x}} \cdot \mathbf{F}_t(\mathbf{x})|<\infty.
    \end{equation}
\end{condition}

\begin{remark}[RC.~\ref{cond:bounded-shared-nonlinear-drift-divergence-MAIN} holds at all time when using \lineardiffusionmodel{}s \cite{skreta_feynman-kac_2025}]\label{rem:bounded-F-div-holds-at-all-timesteps-for-linear-diffusion-models}
    For \lineardiffusionmodel{}s (\(\mathbf{F}_t(\mathbf{x}_t)=\driftparam_t\mathbf{x}_t\)), we have \(\nabla_{\mathbf{x}}\cdot \mathbf{F}_t (\mathbf{x}) = \driftparam_td\) where \(d\) is the dimension of our diffusion model \(d=\text{dim}(\mathbf{x}_t)\).
\end{remark}

\section{Main results: Bounded \logweighingfunction{}}\label{sec:main-results-MAIN}
In this section we examine under which conditions we have a bounded \logweighingfunction{} \(\sup_{\mathbf{x}}h^{\knob}_t(\mathbf{x})=M_t<\infty\) for a given time \(t\). 

\subsection{Result 1: Positive exponents \(\implies\) upper boundedness.}\label{sec:positive-exponents-MAIN}

\paragraph{Statistical-physics interpretation.} At a fixed time $t$, it is useful to introduce the dimensionless effective energies
\[
    U_t^{(k)}(x)\doteq-\log p_t^{(k)}(x),
    \qquad
    U_t^{\mathrm{PoIE}}(x)
    \doteq-\log p_t^{\mathrm{PoIE}}(x)
    =
    \sum_k\beta_k U_t^{(k)}(x)+\mathrm{const}.
\]
The score is then the corresponding statistical force, $\nabla\log p_t^{\mathrm{PoIE}}=-\nabla U_t^{\mathrm{PoIE}}$, and $\lambda$ controls the strength with which the proposal follows this force. Thus, for fixed Langevin gauge $g_t$, increasing $\lambda$ strengthens attraction toward low-$U_t^{\mathrm{PoIE}}$ regions and tends to make the proposal narrower.

The Feynman--Kac rate $h_t^\lambda$ has a complementary population-dynamics interpretation: up to a spatially constant shift, positive $h_t^\lambda(x)$ corresponds to local replication of particle mass and negative $h_t^\lambda(x)$ to killing, closely related to the branching or cloning representations used in rare-event statistical mechanics \cite{giardina2006direct,giardina2011simulating,garbaczewski2024killing}. In Eq.~\ref{eq:1-knob-family-logweighingfunction-in-short-MAIN}, the squared-score term is therefore a force-squared contribution to this branching rate. For $\lambda>1/B$ its coefficient is positive, so regions with large statistical force can generate arbitrarily large positive branching. At the critical value $\lambda=1/B$ this channel cancels exactly. Moreover, when all $\beta_k>0$, the remaining drift-disagreement term is a negative weighted variance of the expert forces: disagreement between experts can only increase killing, never produce runaway branching. In this sense, $\lambda=1/B$ is the strongest confining, or ``narrowest,'' proposal in this family for which the potentially unbounded force-squared branching mechanism is universally screened.

\paragraph{Relation to nonequilibrium work reweighting.} There is also a useful analogy with nonequilibrium work relations. The accumulated Feynman--Kac weight has the exponential form
\[
    w_t
    =
    \exp\!\left(
        \int_0^t h_s(X_s)\,ds
    \right),
\]
so spatial and pathwise fluctuations of $h_t$ play a role analogous to fluctuations of the work entering exponential reweighting identities such as Jarzynski's equality \cite{jarzynski_equilibrium_1997}. Rare trajectories with large positive accumulated rate can therefore dominate the weighted ensemble, just as rare low-work trajectories can dominate exponential work averages. The two controls discussed in this paper act on this problem in complementary ways. Increasing the Langevin gauge $g_t$ enhances relaxation toward the instantaneous target and, in the $g_t\to\infty$ limit of Proposition~\ref{prop:CONJECTURE-Langevin-Infinity-MAIN}, suppresses accumulated weight fluctuations. In contrast, the choice $\lambda=1/B$ acts locally on the Feynman--Kac rate itself: it removes the potentially unbounded force-squared contribution before any large-$g_t$ relaxation limit is taken. Thus $g_t$ controls mixing, whereas $\lambda$ controls the local structure of the required reweighting.

Let us now get back to systematic derivations.
Recall that a high \(\knob\) results in locations of particles (i.e., the proposal distribution) which are tightly clustered while a low \(\knob\) results in those spread apart.
When all exponents are positive, \(\knob=1/B\) is the `narrowest' proposal distribution in the family which still guarantees us an upper bounded \logweighingfunction{}.

\begin{proposition}[Positive exponents \(\implies\) upper boundedness.]\label{prop:positive-exponents-implies-above-boundedness-MAIN}
    (Under RC.~\ref{cond:bounded-score-divergence-MAIN},~\ref{cond:bounded-shared-nonlinear-drift-divergence-MAIN})
    
    When all exponents \(\beta_k\) are positive, then \(\knob\leq1/B\) produces a \logweighingfunction{} that is upper bounded,
    i.e., there exists \(M_t\in\mathbb R\) such that 
    \begin{equation}
        h^{\knob}_t(\mathbf{x}) < M_t < \infty
    \end{equation}
    for all \(x\).
    
    Whereas for \(\knob>1/B\), upper unboundedness may occur.
    \LineBetweenStatementAndProof
    \begin{proof}
        \textbf{Upper boundedness}: 
        
        \begin{itemize}
            \item \textbf{Squared-score term (Eq.~\ref{eq:squared-score-term-MAIN})}: Contributes non-positively due to its non-positive coefficient \((\noiseparam^2_t/2)(\knob-1/B)\).
            \item \textbf{Drift-disagreement term (Eq.~\ref{eq:drift-disagreement-term-MAIN})}: Contributes non-positively due to its negative coefficient  \(-(\noiseparam^2_t/4B)(\beta_j\beta_k)\). This is where positive exponents are required.
            \item \textbf{Score divergence term and shared drift divergence term (Eq.~\ref{eq:score-divergence-term-MAIN},~\ref{eq:shared-nonlinear-drift-divergence-term-MAIN})}: Assumed to be bounded.
        \end{itemize}
        
        \textbf{Above-\textit{un}boundedness:} 
        
        \(\pktdata(\mathbf{x})=\mathcal N(0,I)\) for all \(k\) (which leads to \(\pkt(\mathbf{x}) = \mathcal N(0,(\mu^2_t+\sigma^2_t)\cdot I)\) for some \(\sigma_t\) for all \(k\) (Eq.~\ref{eq:linear-interpolant})) is an example of a data distribution which satisfies RC.~\ref{cond:bounded-score-divergence-MAIN}, but gives above-unboundedness for \(\knob>1/B\).
    \end{proof}
\end{proposition}

\begin{corollary}[Uniqueness within the \((K+1)\)-parameter family]
\label{corr:positive-exponents-result-uniquness-K-param-family}
    In the \((K+1)\)-parameter family Eq.~\ref{eq:K-knob-family-in-short-MAIN}, one can make a similar "narrowest-possible" statement: For any set of positive exponents \(\{\beta_k>0\}_{k=1,\dots,K}\), the choice of knobs \(\knob_k=1/B,\forall k\) is the unique choice for which (under RC.~\ref{cond:bounded-score-divergence-MAIN},~\ref{cond:bounded-shared-nonlinear-drift-divergence-MAIN}) we are guaranteed an upper-bound \textit{and} moreover if the experts coincide (\(\pkt=\pjt,\forall k,j\)), then we get a lower-bound as well (see Cor.~\ref{corr:all-experts-agree-implies-boundedness-MAIN}).
    \LineBetweenStatementAndProof
    \begin{proof}
        Since we assume RC.~\ref{cond:bounded-score-divergence-MAIN},~\ref{cond:bounded-shared-nonlinear-drift-divergence-MAIN}, we need only examine the squared-score term (Eq.~\ref{eq:squared-score-term-MAIN}) and the drift-disagreement term (Eq.~\ref{eq:drift-disagreement-term-MAIN}). One can show that the set of knobs \((\knob_1,\dots\knob_K)\) which guarantee upper-boundedness form a paraboloid. The set of knobs \((\knob_1,\dots\knob_K)\) which an upper and lower bound when the experts coincide forms a line \(\sum_k \beta_k \knob_k=1\). The paraboloid meets this line tangentially at \(\knob_k=1/B\).
    \end{proof}
\end{corollary}

\begin{remark}[A particularly clean case: \(\sumbetaksmall=1,\ \beta_k>0\)]\label{rem:clean-case-sumbetak-is-one}
    An important special case is that when the exponents are both all positive and normalized \(\sumbetaksmall=1\)---here \(\knob=1/B=1\) gives an upper bound without the divergence regularity conditions RC.~\ref{cond:bounded-score-divergence-MAIN},~\ref{cond:bounded-shared-nonlinear-drift-divergence-MAIN}. 
\end{remark}

\begin{remark}[Another particularly clean case: Linear diffusion model \(\wedge\) \(\sumbetaksmall\leq1\)\(\wedge\)\(\beta_k>0\)]\label{rem:clean-case-sumbek-leq-one-and-linear-diffusion-model}
    When assuming \lineardiffusionmodel{}s (Eq.~\ref{eq:linear-diffusion-model-short-MAIN}) and \(\sumbetaksmall\leq1\) we can obtain an upper bound on the \logweighingfunction{} without requiring RC.~\ref{cond:bounded-shared-nonlinear-drift-divergence-MAIN} and RC.~\ref{cond:bounded-score-divergence-MAIN} by setting \(\knob=1\).
\end{remark}

For a different set of exponents, we obtain a lower bound instead. This in particular pertains to what the literature calls the classifier-free guidance \cite{ho_classifier-free_2022} setting: \(\beta_1>1, \beta_2=1-\beta_1\).
\begin{corollary}[Analogous result in the other direction]\label{cor:reverse-holder-result}
    (Under RC.~\ref{cond:bounded-score-divergence-MAIN},~\ref{cond:bounded-shared-nonlinear-drift-divergence-MAIN})
    If, instead of all exponents being positive, exactly one exponent is positive (\(\beta_1>0>\beta_k\) for \(k>1\)), then \(\knob\geq1/B\) produces a \textit{lower} bounded \logweighingfunction{}.
\end{corollary}

\begin{remark}
    In the normalized exponents (\(\sum_k\beta_k=1\)) case, Prop.~\ref{prop:positive-exponents-implies-above-boundedness-MAIN} and Corr.~\ref{cor:reverse-holder-result} can be explained as consequences of Hölder's inequality and reverse Hölder's inequality, respectively, applied to a construction that exploits the models' shared noising process (Appendix.~\ref{sec:discrete-time-result}). 
\end{remark}

\subsection{Result 2: Compact data distributions \(\implies\) upper\(+\)lower boundedness}
A stronger result is available if we assume linear diffusion models  \(\mathbf{F}_t(\mathbf{x}_t)=\driftparam_t\mathbf{x}_t\) (Eq.~\ref{eq:linear-diffusion-model-short-MAIN}) and if the \(K\) target distributions \(\pktdata\) are compact.

\begin{proposition}[Compact data distributions \(\wedge\) \Lineardiffusionmodel{} \(\implies\) upper\(+\)lower boundedness]\label{prop:compact-support-implies-boundedness-MAIN}
    If we have linear diffusion models \(\mathbf{F}_t(\mathbf{x}_t)=\driftparam_t\mathbf{x}_t\), and all \(K\) target distributions \(\pk_{data}\) are compact, then \(\knob=1/B\) results in an upper-and-lower bounded \logweighingfunction{} \(h^{\knob=1/B}_t(\mathbf{x})\) at every non-terminal time \(t\in[\timezero,\timeone)\),
    i.e., for all \(t<1\), there exists \(m_t,M_t\in\mathbb R\) such that 
    \begin{equation}
        -\infty < m_t < h^{\knob=1/B}_t(\mathbf{x}) < M_t < \infty
    \end{equation}
    for all \(\mathbf{x}\).

    Furthermore, \(\knob=1/B\) in the \((1+1)\)-parameter family is equivalent to \(\knob_k=1/B,\forall k\) in the \((K+1)\)-parameter family. One can show that this is the \textbf{unique} set of knobs in the \((K+1)\)-parameter family Eq.~\ref{eq:K-knob-family-in-short-MAIN} which guarantees upper and lower boundedness.
    \LineBetweenStatementAndProof
    \begin{proof}
        \textbf{Boundedness}: 
        
        \begin{itemize}
            \item \textbf{Squared-score term (Eq.~\ref{eq:squared-score-term-MAIN})}: Vanishes when \(\knob=1/B\).
            \item \textbf{Drift-disagreement term (Eq.~\ref{eq:drift-disagreement-term-MAIN})}: Can be written using Tweedie's (Prop.~\ref{prop:Tweedies-identity}) as \(\propto \inlinesqnorm{\mathbb E^j[\mathbf{x}_1|\mathbf{x}_t]-\mathbb E^k[\mathbf{x}_1|\mathbf{x}_t]}\) where conditional expectation is taken with respect to the joint distribution defined in Eq.~\ref{eq:linear-interpolant}. Each expectation is confined to a compact region, hence the distance between them is bounded.
            \item \textbf{Score divergence term (Eq.~\ref{eq:score-divergence-term-MAIN})}: Can be written using Tweedie's (Prop.~\ref{prop:Laplacian-is-Covariance}) as the trace of covariance \(\propto \sum_k \beta_k \Tr\Cov^k[\mathbf{x}_1|\mathbf{x}_t]+(\text{constant})\) where conditional covariance is taken with respect to the joint distribution defined in Eq.~\ref{eq:linear-interpolant}. Each covariance is again bounded due to compactness.
            \item \textbf{Shared drift divergence term (Eq.~\ref{eq:shared-nonlinear-drift-divergence-term-MAIN})}: Is a constant: \(\nabla_{\mathbf{x}} \cdot \mathbf{F}_t(\mathbf{x}_t) = \driftparam_t d\) where \(d\) is the dimension of \(\mathbf{x}_t\).
        \end{itemize}
        Both applications of Tweedie's identities (Prop.~\ref{prop:Tweedies-identity},~\ref{prop:Laplacian-is-Covariance}) exploit the fact that \(p^k_{t}\) for \(t<1\) is an isotropic Gaussian convolution of \(p^k_{t=1}\) when using \lineardiffusionmodel{}s (see Appendix.~\ref{sec:linear-diffusion-model}). For the full proof, including for the uniqueness statement, see Prop.~\ref{prop:compact-support-implies-boundedness}.

    \end{proof}
\end{proposition}

\subsection{Result 3: One diffusion model (\(K=1\)) \(\implies\) upper\(+\)lower boundedness}
This follows trivially from regularity conditions~\ref{cond:bounded-score-divergence-MAIN} and \ref{cond:bounded-shared-nonlinear-drift-divergence-MAIN}

\begin{proposition}[One diffusion model (\(K=1\)) \(\implies\) upper\(+\)lower boundedness] \label{prop:one-model-implies-boundedness-MAIN}
    (Under RC.~\ref{cond:bounded-score-divergence-MAIN},~\ref{cond:bounded-shared-nonlinear-drift-divergence-MAIN})
    
    If we have one diffusion model (\(K=1\)), then \(\knob=1/B\) results in an above-and-below bounded \logweighingfunction{} \(h^{\knob=1/B}_t(\mathbf{x})\).

    Furthermore, \(1/B\) is the unique value in the \((1+1)\)-parameter family that guarantees this property for all experts satisfying RC.~\ref{cond:bounded-score-divergence-MAIN},~\ref{cond:bounded-shared-nonlinear-drift-divergence-MAIN}.
    \LineBetweenStatementAndProof
    \begin{proof}
    
        \textbf{Above-boundedness}: 
        \begin{itemize}
            \item \textbf{Squared-score term (Eq.~\ref{eq:squared-score-term-MAIN})}: Vanishes when \(\knob=1/B, K=1\).
            \item \textbf{Drift-disagreement term (Eq.~\ref{eq:drift-disagreement-term-MAIN})}: Vanishes when \(K=1\).
            \item \textbf{Score divergence term and shared drift divergence term (Eq.~\ref{eq:score-divergence-term-MAIN},~\ref{eq:shared-nonlinear-drift-divergence-term-MAIN})}: Assumed to be bounded.
        \end{itemize}

    \textbf{Uniqueness}: Any Gaussian \(p^1_{t}\) gives an unbounded square-score term which vanishes iff \(\knob=1/B\).
    \end{proof}
\end{proposition}

More generally, in the multiple diffusion model setting \(K>1\), if all expert marginals coincide \(\pkt=\pjt\), then the drift-disagreement term Eq.~\ref{eq:drift-disagreement-term-MAIN} vanishes and the setting becomes identical to the \(K=1\) setting, resulting in the following.
\begin{corollary}
    [All experts agree \(\implies\) upper\(+\)lower boundedness] \label{corr:all-experts-agree-implies-boundedness-MAIN}
    (Under RC.~\ref{cond:bounded-score-divergence-MAIN},~\ref{cond:bounded-shared-nonlinear-drift-divergence-MAIN})

    If all experts coincide \(\pkt=\pjt\), then \(\knob=1/B\) and only \(\knob=1/B\) guarantees an above and below bounded \logweighingfunction{} \(h^{\knob=1/B}_t(\mathbf{x})\).
\end{corollary}

The vanishing of the drift-disagreement term Eq.~\ref{eq:drift-disagreement-term-MAIN} can be interpreted more generally. For example, if the experts agree along certain directions, corresponding components in the drift-disagreement term will cancel 
\footnote{For a concrete example, consider \(K\) data distributions lying on a common affine subspace. When using \lineardiffusionmodel{}s, which have intermediate timestep marginals which are isotropic Gaussian convolutions of the data distributions, the score differences then have no component in the directions orthogonal to the subspace. 
More generally, one can consider data distributions that are product measures over a common orthogonal decomposition of \(\mathbb R^d\), with one factor shared across all \(K\) experts. We leave the exploration of whether such a modeling assumption is useful in the context of real-world high-dimensional data distributions such as images, or in latent spaces, as further work.}. 
Alternatively, if the \(K\) marginals have Gaussian tail decay with a shared decay rate, then the drift-disagreement term (Eq.~\ref{eq:drift-disagreement-term-MAIN}) is upper\(+\)lower bounded---Prop.~\ref{prop:compact-support-implies-boundedness-MAIN} is a consequence of this. 

\section{Experiments}\label{sec:experiments}

We do not have sufficient reason to a-priori believe that an upper-bound nor even an upper\(+\)lower bound will result in higher sampling efficiency---these claims must rather be empirically verified. More specifically, we aim to test the following two hypotheses on two different parameters; the knob which affects the \logweighingfunction{} (\(\knob\) in Eq.~\ref{eq:1-knob-family-in-short-MAIN}), and the Langevin gauge term which keeps the \logweighingfunction{} constant (\(g_t\) in Eq.~\ref{eq:1-knob-family-in-short-MAIN}).

\paragraph{Two hypotheses to test}
More specifically, based on the discussions in Sec.~\ref{sec:main-results-MAIN} and Prop.~\ref{prop:CONJECTURE-Langevin-Infinity-MAIN} respectively, we aim to test the following two hypotheses.
\begin{hypothesis}\label{hyp:knob=1/B-gives-best-performance}
    The choice \(\knob=1/B\) (in the \((1+1)\)-parameter family Eq.~\ref{eq:1-knob-family-in-short-MAIN}) results in the most efficient sampling \emph{under fixed (Euler--Maruyama integration step size) compute budget}.
\end{hypothesis}
\begin{hypothesis}\label{hyp:higher-Langevin-gives-best-performance}
    A higher Langevin parameter (\(g_t\) in Eq.~\ref{eq:wSODE-with-zero-directional-weight-MAIN}) results in more efficient sampling \emph{given enough compute budget (in the form of sufficiently fine Euler--Maruyama integration step size)}. 
\end{hypothesis}
Here, we define 'efficient sampling' as resulting in better distributional distance metrics (such as Wasserstein distance) when given access to the same number of particles \footnote{Because if one can afford to take the number of particles to infinity, one can expect, except for pathological cases, the SMC algorithm to converge to the true distribution under any choice of \(\knob\) and \(g_t\).} and the same Euler--Maruyama time step size \footnote{Because if one can afford to take the Euler--Maruyama time step size to zero, then one can take a very high value of \(g_t\) (in Eq.~\ref{eq:1-knob-family-in-short-MAIN}) which, according to Prop.~\ref{prop:CONJECTURE-Langevin-Infinity-MAIN}, will result in the SMC algorithm trivially outputting the true distribution.}.

We use \lineardiffusionmodel{}s (Eq.~\ref{eq:linear-diffusion-model-short-MAIN}) in all experiments \footnote{Recall that this resolves the two regularity conditions due to Remarks.~\ref{rem:score-smooth-probably-holds-for-linear-diffusion-models} and Remarks.~\ref{rem:bounded-F-div-holds-at-all-timesteps-for-linear-diffusion-models}.}.

\subsection{Experimental details}\label{sec:experimental-details-MAIN}
\paragraph{The grid search}

We run a grid search over the two hyperparameters: the knob \(\knob\) in Eq.~\ref{eq:1-knob-family-in-short-MAIN} (which influences the  \logweighingfunction{} \(h^{\knob}_t(\cdot)\)) and the Langevin gauge knob \(g_t\) in Eq.~\ref{eq:1-knob-family-in-short-MAIN} (which leaves the \logweighingfunction{} intact Lemma.~\ref{prop:langevin-gauge-MAIN}, and when taken to infinity trivially results in perfect sampling Prop.~\ref{prop:CONJECTURE-Langevin-Infinity-MAIN}). 

\paragraph{Technical detail: reparameterization}
For ease of comparison, we re-write the two knobs \((\knob,\, g_t)\) of our grid-search
as \((\knob B,\,  \zeta_t \doteq g_t/\noiseparam_t)\), where \(B\doteq\sum_k\beta_k\) and \(\noiseparam_t\) (Eq.~\ref{eq:nonlinear-diffusion-model-MAIN}). 
The reasons are as follows. First, the reparameterization \(\knob \to \knob B\) results in the choice of \(\knob=1/B\) proposed in Hypothesis.~\ref{hyp:knob=1/B-gives-best-performance} staying one \(\knob B=1\) across settings. Second, recall that \(\noiseparam_t\) is determined (via Eq.~\ref{eq:formula-for-drift-parameter-and-noise-parameter}) by the (arbitrary) choice of space-time parameterization \((\mu_t,\sigma_t)\) in Eq.~\ref{eq:linear-interpolant}, commonly called the noising schedule. Unlike \(g_t\), the rescaled Langevin strength \(\zeta_t\doteq g_t/\noiseparam_t\) is invariant to this choice (Remark~\ref{rem:why-fix-shared-drift}, Eq.~\ref{eq:1-knob-family-schedule-free}), which makes results comparable across pre-trained models with different schedules. We also choose a time-constant \(\zeta_t=\zeta\) for simplicity.

\begin{figure}[htbp]
    \centering
        \includegraphics[width=1.0\linewidth]{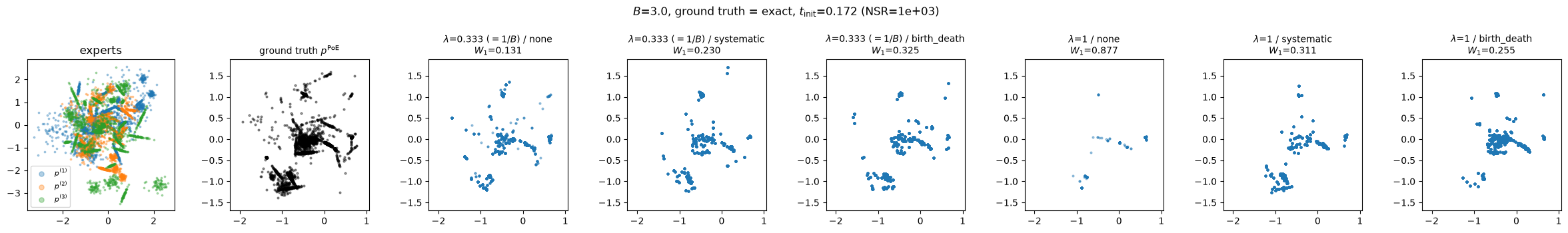}
    \captionsetup{width=0.9\linewidth}
    
    \caption{A \(K=3\) \textbf{Mixture-of-Gaussians} settings with \(\beta_i=1\) and \(B=\sum^K_{k=1}\beta_k=3\). We compare \(\knob=1/B\) proposed in our paper to \(\knob=1\) proposed in \cite{skreta_feynman-kac_2025}. Interestingly, we see that our \(\knob\) with \textit{no SMC resampling} achieved best Wasserstein-1 metric. This is likely due to resampling reducing particle diversity, and we expect this superiority to be an artifact of the low number of particles used in this toy setting.
    }
    \label{fig:MoG_particles}
\end{figure}

\paragraph{The experimental settings}
The grid search is performed in four different experimental settings used in previous works (\cite{skreta_feynman-kac_2025} and \cite{he2026rne}): 
alanine dipeptide (ALDP) (Fig.~\ref{fig:ALDP-beta-2.666}), LJ-13 (Fig.~\ref{fig:LJ13-beta-2}, Fig.~\ref{fig:LJ13-beta-1.333}), Mixture-of-Gaussians (MoG) (Fig.~\ref{fig:MoG_particles}, Fig.~\ref{fig:MoG_metrics}) and dual-target molecular generation (Fig.~\ref{fig:sbdd-dual-target-molecule-generation}). 
Three of the four settings (ALDP, LJ-13, MoG) are especially useful in that it allows the calculation of explicit distributional distance metrics (such as Wasserstein distance) since we have access to the samples from the true distribution. In the other setting (dual-target molecular generation) we use proxy metrics (e.g. sum of binding scores of the generated small molecule to each of the two target proteins) motivated by the application domain. 

\begin{table}[htb]
    \centering
    \begin{threeparttable}
    \begin{tabular}{|l|l|l|}
    \hline
    \text{Experimental setting} & \textit{No. diffusion models (\(K\))} & \textit{W1, W2, TV calculable} \\ \hline
    \textbf{ALDP}                             & 1                                     & Yes                            \\ \hline
    \textbf{LJ-13}                            & 1                                     & Yes                            \\ \hline
    \textbf{MoG}                              & \(\geq1\)                             & Yes                            \\ \hline
    \textbf{Dual-target mol. generation} & 2                                     & No                             \\ \hline
    \end{tabular}
    \caption{A summary of the four experimental settings used in this paper. All are based on settings tested in two previous works \cite{skreta_feynman-kac_2025} and \cite{he2026rne}.}
    \end{threeparttable}
\end{table}

For further details on experimental settings, including the definitions of metrics used (Energy-Total-Variation, Distance-Total-Variation, and Wasserstein-2 distance) see Appendix.~\ref{sec:further-experimental-details}, \cite{skreta_feynman-kac_2025} and \cite{he2026rne}.

\subsection{Results and interpretations}\label{sec:experimental-results}
\begin{figure}[htbp]
    \centering
    \includegraphics[width=1.0\linewidth]{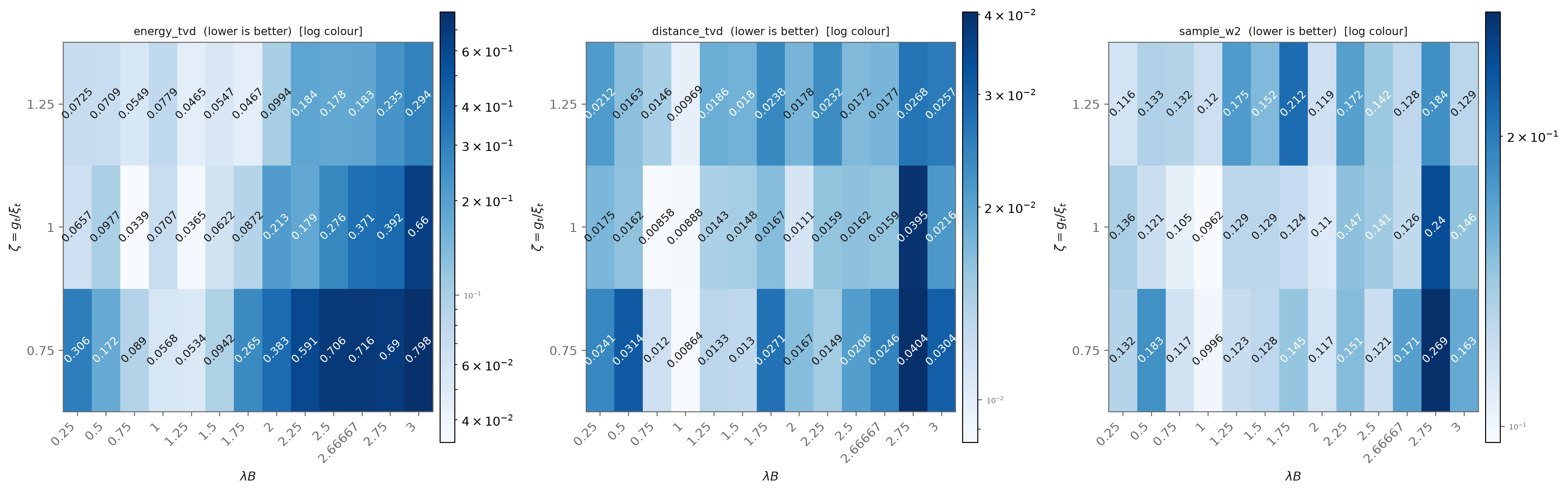} 
    \captionsetup{width=0.9\linewidth}
    \caption{\textbf{ALDP} (for which \(K=1\)) with exponent \(\beta_1=B=8/3\). One sees that a knob \(\knob\) near our proposed \(\knob=1/B\) (i.e., \(\knob B=1\)) often achieves better total-variation distance and Wasserstein-2 distance than \(\knob=1\) proposed in \cite{skreta_feynman-kac_2025}.}
    \label{fig:ALDP-beta-2.666}
\end{figure}

\begin{figure}[htbp]
    \centering
    \includegraphics[width=1.0\linewidth]{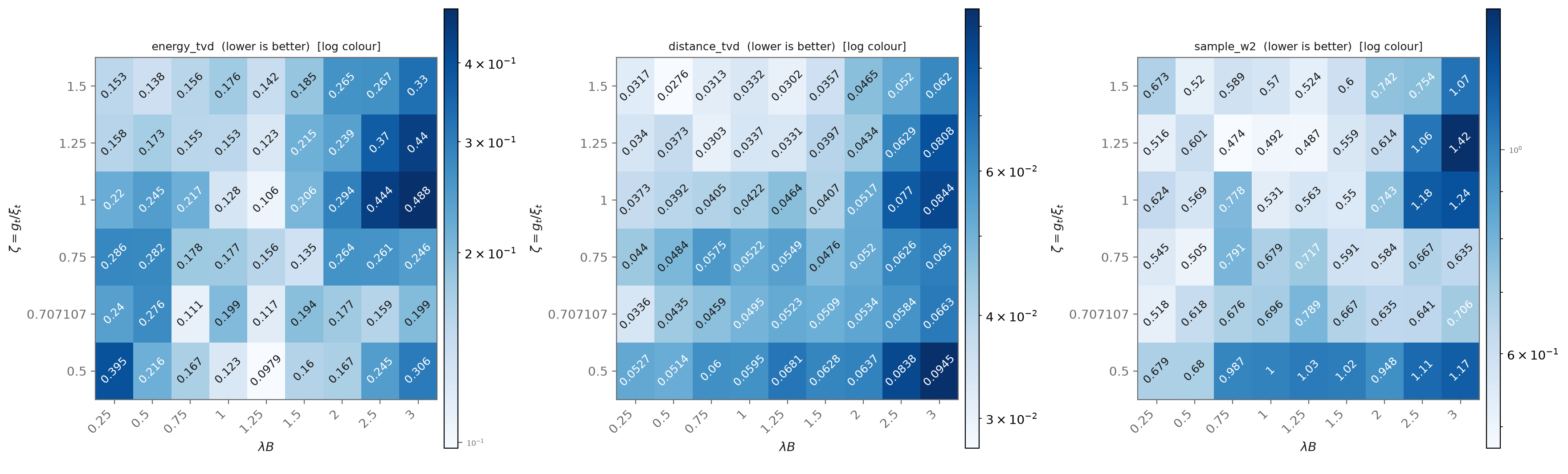} 
    \captionsetup{width=0.9\linewidth}
    \caption{\textbf{LJ-13} (for which \(K=1\)) with exponent \(\beta_1=B=2\). One sees that a knob \(\knob\) near our proposed \(\knob=1/B\) often achieves better total-variation distance and Wasserstein-2 distance than \(\knob=1\) proposed in \cite{skreta_feynman-kac_2025}.}
    \label{fig:LJ13-beta-2}

    \includegraphics[width=1.0\linewidth]{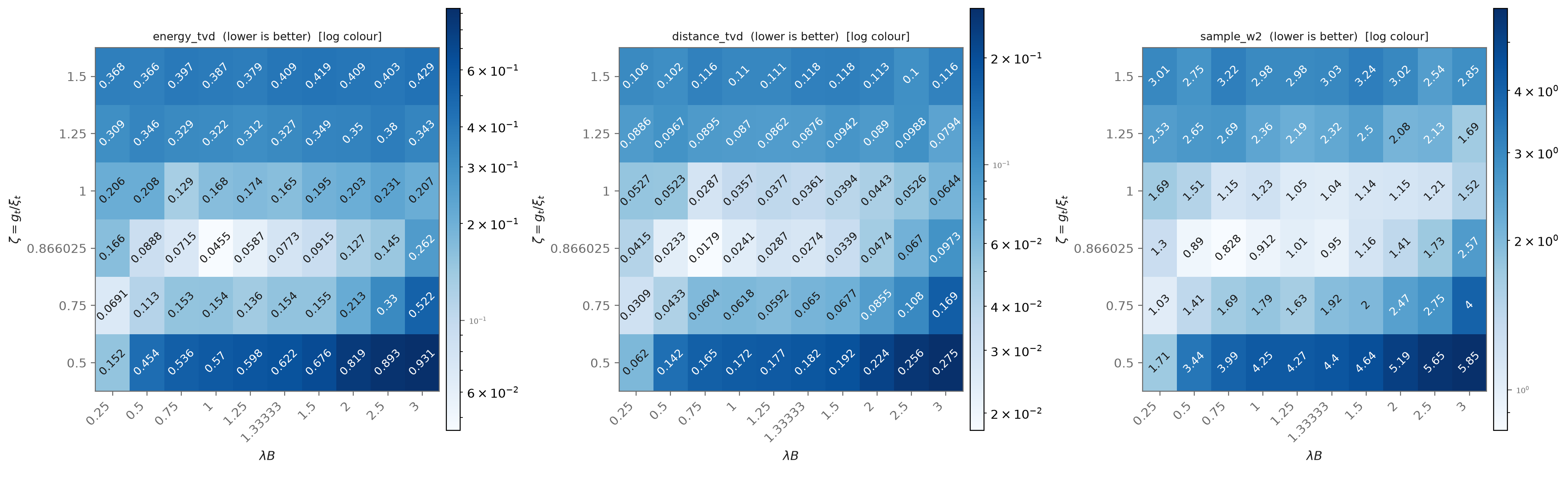} 
    \captionsetup{width=0.9\linewidth}
    \caption{\textbf{LJ-13} (for which \(K=1\)) with exponent \(\beta_1=B=4/3\). Results are similar as the \(\beta_1=2\) (Fig.~\ref{fig:LJ13-beta-2}), except the currently unexplained phenomena of a higher Langevin parameter (\(\zeta=g_t/\noiseparam_t\), the y-axis) resulting in worse metrics is more pronounced.}
    \label{fig:LJ13-beta-1.333}
\end{figure}

\begin{figure}[htbp]
    \centering
    \includegraphics[width=1.0\linewidth]{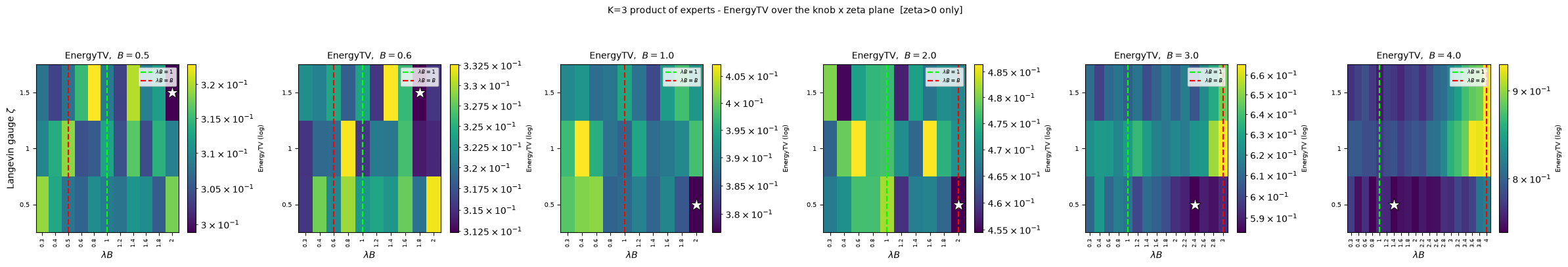} 
    
    \includegraphics[width=1.0\linewidth]{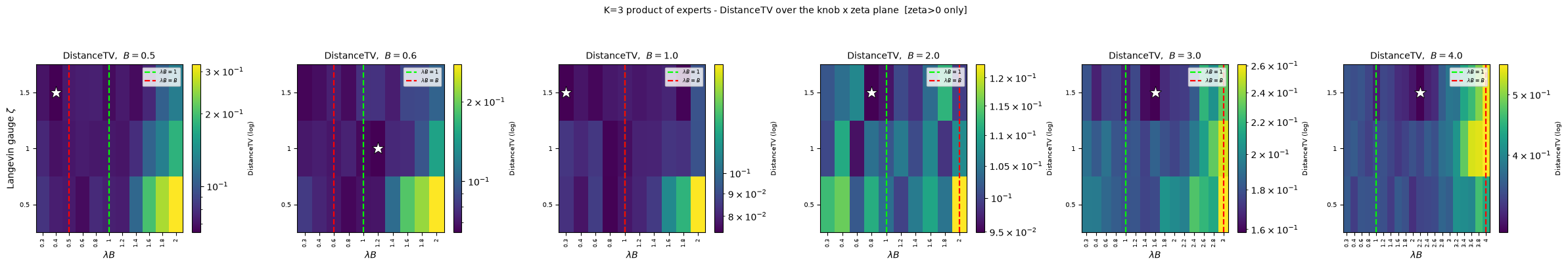} 
    
    \includegraphics[width=1.0\linewidth]{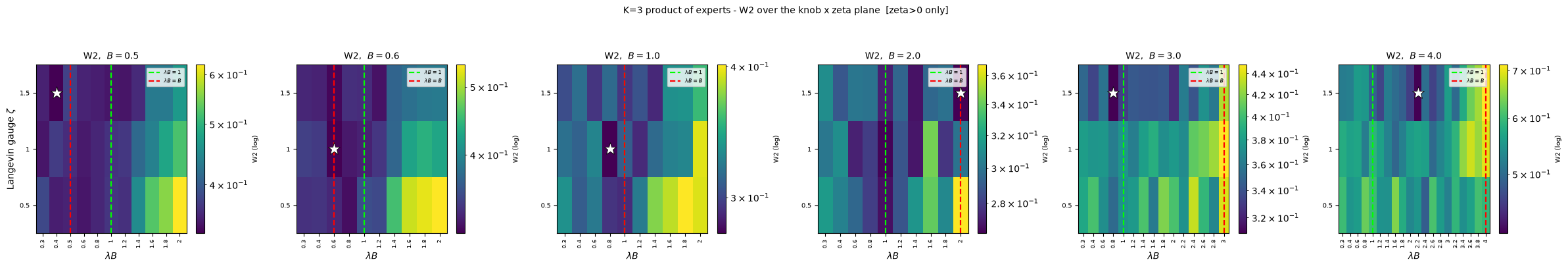} 
    \captionsetup{width=0.9\linewidth}
    
    \caption{A \(K=3\) \textbf{Mixture-of-Gaussians} settings with exponents \(\beta_1=\beta_2=\beta_3\) with 
    \(\sumbetaksmall=B\in\{0.5, 0.6, 1, 2, 3, 4\}\). 
    This figure reports the Energy-Total-Variation metric, Distance-Total-Variation metric, and Wasserstein-2 metric. We observe a high amount of variance in the Energy-TV metric. In the latter two metrics, we observe that a knob \(\lambda\) between \(1/B\) and \(1\) often achieves best metrics. Also, a high Langevin parameter (here defined as \(\zeta\)) in general achieves better metrics.}
    \label{fig:MoG_metrics}
\end{figure}

\begin{figure}[htbp]
    \centering
    \includegraphics[width=1.0\linewidth]{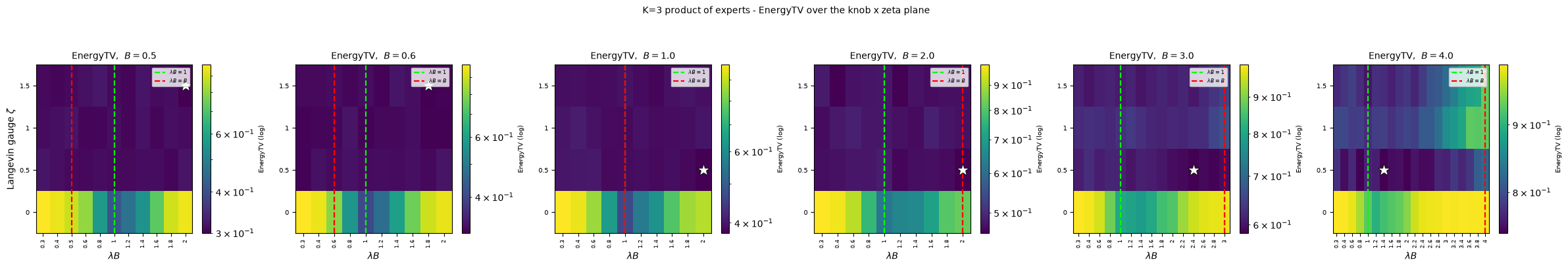} 
    
    \includegraphics[width=1.0\linewidth]{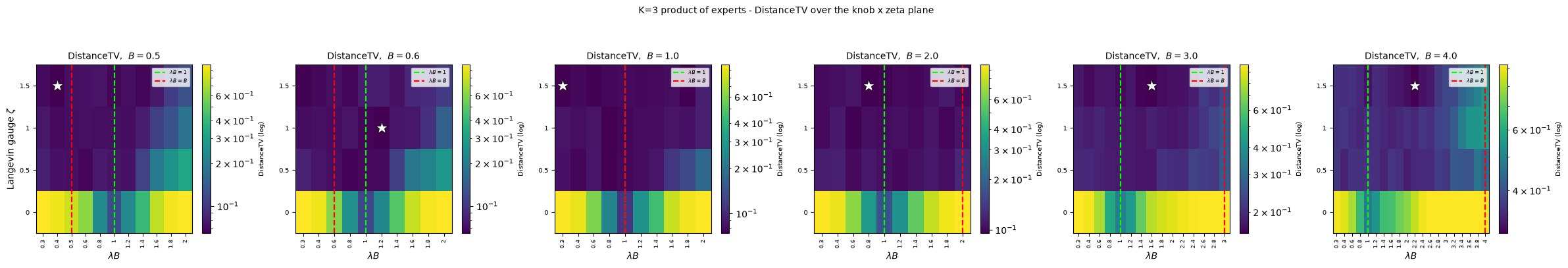} 
    
    \includegraphics[width=1.0\linewidth]{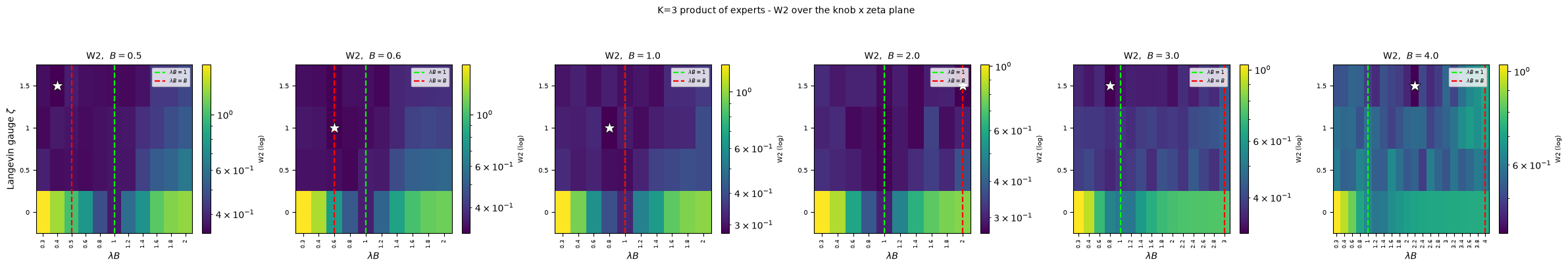} 
    \captionsetup{width=0.9\linewidth}
    
    \caption{Fig.~\ref{fig:MoG_metrics} with an additional value of Langevin parameter \(\zeta\) plotted: \(\zeta=0\). (Fig.~\ref{fig:MoG_metrics} was plotted without \(\zeta=0\) to make clear the differences between the nonzero \(\zeta\) settings, the values of which are closer together in general.) We see that in the ODE (\(\zeta=0\)) regime, picking a \(\knob\) near \(1/B\) has even greater benefit than in the SDE (\(\zeta>0\)) regime.}
    \label{fig:MoG_metrics_with_zero_langevin_row}
\end{figure}

\begin{figure}[htbp]
    \centering
    \includegraphics[width=1.0\linewidth]{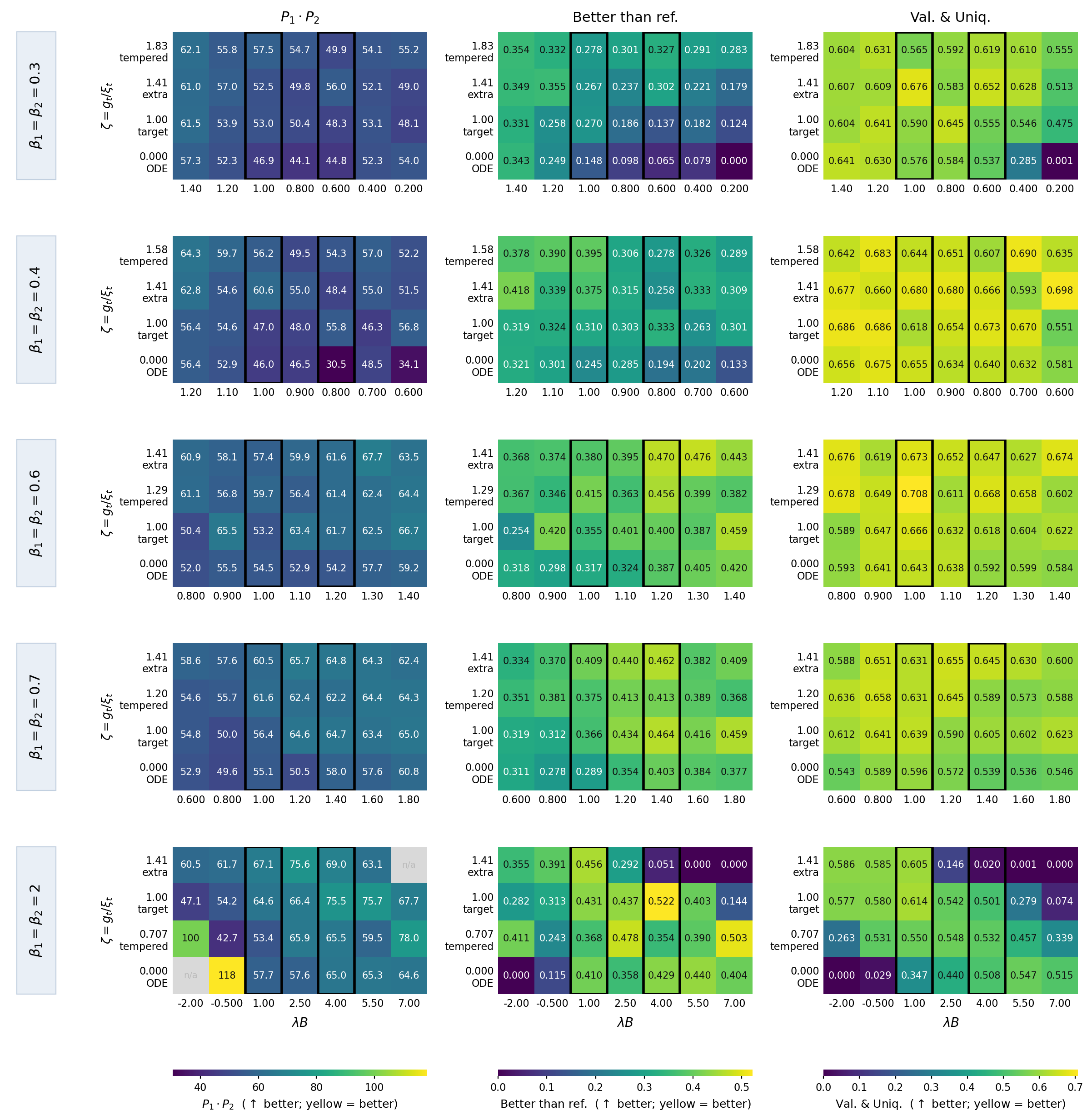} 
    
    \captionsetup{width=1.0\linewidth}
    
    \caption{Dual-target molecular generation task introduced in \cite{skreta_feynman-kac_2025}. Run over multiple \(\beta_k\)s, values of \(\knob\) and Langevin parameter \(\zeta=g_t/\noiseparam_t\). In this experimental setting, it is harder to make out the superior sampling method as there are multiple metrics. For example, we see that a higher \(\beta_k\) in general results in a better (higher) product of docking scores \(P_1\cdot P_2\) (as reported in \cite{skreta_feynman-kac_2025}) but at the cost of more unstable sampling and sensitivity to choice of \((\knob,\zeta)\); The proportion of invalid (not corresponding to a physically realizable molecule) or non-unique molecules among those generated is extremely high for certain \((\knob,\zeta)\) settings when one chooses a high \(\beta_k\). Note also that for high \(\beta_k=2\), the choice of \(\knob=1/B\) resulted in the widest range of \(\zeta\) for which sampling was stable (high Val. \& Uniq. values). The 'tempered' value of \(\zeta\) refers to \(\zeta=1/\sqrt {\beta_k}\) introduced in \cite{skreta_feynman-kac_2025}.}
    \label{fig:sbdd-dual-target-molecule-generation}
\end{figure}

\paragraph{Results: Hypothesis.~\ref{hyp:higher-Langevin-gives-best-performance}}
Out of the two hypotheses, the results paint a less ambiguous picture for Hypothesis.~\ref{hyp:higher-Langevin-gives-best-performance}: \emph{In most settings (Figs.~\ref{fig:ALDP-beta-2.666},~\ref{fig:LJ13-beta-2},~\ref{fig:MoG_metrics},~\ref{fig:MoG_metrics_with_zero_langevin_row},~\ref{fig:sbdd-dual-target-molecule-generation}) we see a higher Langevin parameter \(g_t\) result in better metrics}. Two notable exceptions are Fig.~\ref{fig:LJ13-beta-1.333} and the bottom row (\(\beta_1=\beta_2=2\)) of Fig.~\ref{fig:sbdd-dual-target-molecule-generation}. We leave investigation of this as further work.

\paragraph{Results: Hypothesis.~\ref{hyp:knob=1/B-gives-best-performance}}
Hypothesis.~\ref{hyp:knob=1/B-gives-best-performance} holds with a more varying degree among the experimental settings. In general, the knob \(\knob=1/B\) demonstrates clear efficacy in the single model \(K=1\) settings (Fig.~\ref{fig:ALDP-beta-2.666},~\ref{fig:LJ13-beta-2},~\ref{fig:LJ13-beta-1.333}) and the non-stochastic ODE (\(g_t=0\) in Eq.~\ref{eq:1-knob-family-in-short-MAIN}) settings (Fig.~\ref{fig:MoG_metrics_with_zero_langevin_row}), but less so in the stochastic (\(g_t>0\)) multiple models (\(K>1\)) settings (Fig.~\ref{fig:MoG_metrics}).
\emph{One possible explanation for \(\knob=1/B\) having clearer superiority in the single diffusion model setting \(K=1\) than in the \(K>1\) setting is that the latter we are only guaranteed an upper bound (due to Prop.~\ref{prop:positive-exponents-implies-above-boundedness-MAIN}), whereas in the former we have both an upper and lower bound (due to Prop.~\ref{prop:one-model-implies-boundedness-MAIN}).}

We must also note that a reliable violation of Hypothesis.~\ref{hyp:knob=1/B-gives-best-performance} occurs in the Mixture-of-Gaussians (\(K=3\)) setting. In the \textbf{ODE} regime (i.e., \(g_t=0(=\zeta)\) in Eq.~\ref{eq:1-knob-family-in-short-MAIN} and \(\zeta=0\) in the figures), we see a clear out-performance of \(\knob=1/B\) compared to other values of \(\knob\) (see the \(\zeta=0\) rows of heatmaps in Fig.~\ref{fig:MoG_metrics_with_zero_langevin_row}). This result is in line with Hypothesis.~\ref{hyp:knob=1/B-gives-best-performance}. 
However, in the \textbf{SDE} regime (\(g_t>0\), \(\zeta>0\))  (Fig.~\ref{fig:MoG_metrics}), a knob 
\(\knob \in (1,1/B)\) (when \(B<1\)) or \(\knob\in(1/B,1)\) (when \(B>1\)) 
often out-performs \(\knob=1/B\)~\footnote{See the location of the white stars (the best performed setting) in relation to the green line (\(\knob=1/B\)) in Fig.~\ref{fig:MoG_metrics}}. One possible conjecture is that the score divergence term (Eq.~\ref{eq:score-divergence-term-MAIN}), which we assume to be upper\(+\)lower bounded (Rem.~\ref{rem:score-smooth-probably-holds-for-linear-diffusion-models}) may have a more detrimental effect on sampling efficiency than our theory suggested
    \footnote{To elaborate, the magnitude of the score divergence term (Eq.~\ref{eq:score-divergence-term-MAIN}) is minimized at \(\knob=1\) (due to coefficient \((-1+\knob)\)). (Note that \(\knob=1\) is also the choice of \(\knob\) exposited in \cite{skreta_feynman-kac_2025}.) Recall that we assert that one can reasonably assume the score divergence to be upper\(+\)lower bounded at any non-terminal time steps \(t<1\) for most (non-pathological) data distributions (Remark.~\ref{rem:score-smooth-probably-holds-for-linear-diffusion-models}). This motivated us to work under the assumption that one can 'ignore' this score divergence term, and focus on flattening the sum of the squared-score term (Eq.~\ref{eq:squared-score-term-MAIN}) and the drift-disagreement-term (Eq.~\ref{eq:drift-disagreement-term-MAIN})---hence leading to Prop.~\ref{prop:positive-exponents-implies-above-boundedness-MAIN},~\ref{prop:compact-support-implies-boundedness-MAIN},~\ref{prop:one-model-implies-boundedness-MAIN} and the Hypothesis.~\ref{hyp:knob=1/B-gives-best-performance}. However, the empirical results above suggest that a knob between the two values \(1/B\) and \(1\)---i.e., a compromise between the magnitudes of the squared-score term and the score divergence term---leads to more efficient sampling in this Mixture-of-Gaussians setting.}
.
We leave a more detailed investigation of this as further work.

\section{Related works and scope}

\paragraph{WSDEs and SMC in the context of deep generative models}
Weighted stochastic differential equations, or more formally Feynman--Kac particle systems, and sequential Monte Carlo algorithms have been used in conjunction with pre-trained deep generative models to enable sampling from distributions other than what the model was originally designed and trained to sample from (\(p_{data}(\mathbf{x})\)). This includes reward-tilted sampling (\(\propto p_{data}(\mathbf{x})r(\mathbf{x})\)), product-of-experts (\(\prod_k(p^k_{data})^{\beta_k}\)), and annealing (\((p_{data})^{\beta}\)). We deal with the latter two in our paper.

\paragraph{Sampling from the product of diffusion models}
The problem of sampling from the weighted product of data distributions of several diffusion models, or the product-of-experts (\textbf{PoE}) marginal
\begin{equation}\tag{Eq.~\ref{eq:PoE-MAIN}}
    p^{\PoE} \doteq \frac1Z \prod^K_{k=1} (\pk_{data})^{\beta_k}
\end{equation}
has first been explored in \cite{du2023reduce}. They noted that the naive approach of simply replacing the score \(\nabla_{\mathbf{x}} \log p_t(\mathbf{x}_t)\) by \(\sum_k \beta_k \nabla_{\mathbf{x}} \log \pkt(\mathbf{x}_t)\) in the original diffusion model SDE does not correctly sample from \(p^{PoE}\). Hence, they introduced the product-of-intermediate-experts (\textbf{PoIE}) trajectory 
\begin{align}\tag{Eq.~\ref{eq:PoIE-trajectory-MAIN}}
   \pPoIE_{t}
    \doteq
    \frac{1}{Z_t}\prod^K_{k=1} (\pkt)^{\beta_k} \ \text{ with }\  \sumbetak>0
\end{align}
and sampled from this using annealed Langevin dynamics, which amounts to running multiple Langevin mixing steps at each (discretized) timestep.

\cite{skreta_feynman-kac_2025} was the first to derive a WSDE which samples from the PoIE trajectory, and the PoE marginal more generally. Their WSDE has the merit of having a \logweighingfunction{} without a vector field divergence term, which was thought to require expensive auto-differentiation based calculation to implement in practice. However, \cite{he2026rne} noted that any WSDE with any vector field divergence term \(\nabla_{\mathbf{x}} \cdot \mathbf{f}_t(\mathbf{x})\) in a WSDE \logweighingfunction{} can be calculated without auto-differentiation by using the identity Eq.~\ref{eq:divergence-calculation-trick-backward-forward-ito-integral}.
It is worth noting that this identity cannot be applied in the context of weighted ODEs (\(g_t=0\)).

\cite{he2026rne} was also the first to use directional weights (\(\mathbf{j}_t(\mathbf{x}_t)\cdot d\mathbf{W}_t\) term in \(d\log w_t\)) in the context of this problem. We do not explore this degree of freedom in our paper.

\cite{lee2025collapse} pointed out the potential of unnormalizable marginals arising when defining the PoIE trajectory. As amelioration they proposed the generalized PoIE trajectory with time varying exponents \(\beta_{k,t}\) which we do not explore in this paper.

The repeated appearance \cite{skreta_feynman-kac_2025,he2026rne,lee2025collapse} of the seemingly arbitrary decision to sampling from the PoIE trajectory (in the aim of sampling from the PoE marginal) can be explained by the fact that the PoIE trajectory results in a particularly simple expression of \(\partial_t \log \pPoIE_t(\mathbf{x})\) (due to \(=\sum_k \beta_k\,\partial_t \log \pkt(\mathbf{x})\)) and hence the \logweighingfunction{} (Prop.~\ref{prop:all-WSDEs-sampling-from-PoIE-of-nonlinear-diffusion-models}). 

We present a re-derivation of the family of WSDEs in \cite{he2026rne} which, unlike the derivation in \cite{he2026rne}, does not invoke path measures but simply the Fokker--Planck PDE. We also present an errata of  \cite[Prop~D.3]{skreta_feynman-kac_2025} (Remark.~\ref{rem:skreta-errata})

\paragraph{Sampling from an annealed version of a single diffusion model (\(K=1\))}
One of our main results is a WSDE which (in the linear diffusion model setting Remark.~\ref{rem:score-smooth-probably-holds-for-linear-diffusion-models}) has an upper\(+\)lower bounded \logweighingfunction{} when \(K=1\). The \(K=1\) setting is also referred to as sampling from an annealed diffusion model, as we are sampling from the target distribution powered to some exponent (Eq.~\ref{eq:PoE-MAIN} with \(K=1\)).
There exists a well-known method for calculating the log-likelihood of a non-stochastic (ODE) generative model:
One integrates the divergence of the score throughout the generation path. One can use this calculated log-likelihood for importance sampling from the annealed version of the original flow-matching model. We note that this importance sampling scheme is an instance of the knob \(\knob=1/B\) exposited in our paper. To elaborate, the above importance sampling scheme using the divergence-based calculation of the log-likelihood becomes equivalent to our Eq.~\ref{eq:1-knob-family-in-short-MAIN} with \(\knob=1/B\) and with zero Langevin parameter \(g_t=0\). 
Hence our contribution in this context is two-fold: noting the existence of a stochastic generalization, and conjecturing that, in the \lineardiffusionmodel{} context, the importance weight can be expected to be upper\(+\)lower bounded for non-pathological data distributions at all non-terminal times  (i.e., conjecturing that regularity condition.~\ref{cond:bounded-score-divergence-MAIN} holds for non-pathological data distributions at all non-terminal times; see Remark.~\ref{rem:score-smooth-probably-holds-for-linear-diffusion-models}).

\section{Discussion}

\subsection{Summary of contributions}
\paragraph{Primary contributions}
Our main contribution is deriving WSDEs which sample from a product of diffusion models (Eq.~\ref{eq:nonlinear-diffusion-model-MAIN}) with a \logweighingfunction{} which is either upper bounded (Prop.~\ref{prop:positive-exponents-implies-above-boundedness-MAIN}), or both upper and lower bounded (Prop.~\ref{prop:compact-support-implies-boundedness-MAIN}~\ref{prop:one-model-implies-boundedness-MAIN}). In the setting of the \lineardiffusionmodel{}, one can expect the regularity conditions necessary for Prop.~\ref{prop:positive-exponents-implies-above-boundedness-MAIN},~\ref{prop:one-model-implies-boundedness-MAIN} to hold at all non-terminal time \(t<1\) for all but highly pathological data distributions \(\{\pktdata\}_{k=1,\dots,K}\) (Remarks.~\ref{rem:score-smooth-probably-holds-for-linear-diffusion-models},~\ref{rem:bounded-F-div-holds-at-all-timesteps-for-linear-diffusion-models},~\ref{rem:compact-support-result-doesnt-hold-at-time-one}).

\paragraph{Secondary contributions}
We show that taking the Langevin strength to infinity makes the weights constant for an arbitrary drift (Prop.~\ref{prop:CONJECTURE-Langevin-Infinity-MAIN}), generalising a result of \cite[Thm.~2]{guo2026complexity} for annealed Langevin dynamics. This result is relevant to not only the problem of sampling from the product-of-experts of generative models but in any other contexts for which the score of the marginal generated by the WSDE (i.e., \(\nabla_{\mathbf{x}} \log \rho_t(\mathbf{x}_t)\) in Eq.~\ref{eq:wSODE-with-zero-directional-weight-MAIN}) is known. This includes the reward-skewed diffusion model sampling setting in \cite{skreta_feynman-kac_2025}.

We unify and simplify existing works in the literature. We show that \cite{he2026rne}'s work can be derived using the Fokker--Planck PDE without mention of path measure derivatives (Eq.~\ref{eq:translation-from-He26-to-our-notation}). We also show an errata of \cite{skreta_feynman-kac_2025} (Rem.~\ref{rem:skreta-errata}) using Lemma.~\ref{prop:langevin-gauge-MAIN}.

\subsection{Limitations of theoretical results}
\label{sec:t-uparrow-one-case-MAIN}
\paragraph{Lack of time uniform boundedness: \(t\uparrow1\) limit}

Note that the compact setting upper\(+\)lower bound only holds at non-terminal times \(t<1\) (for proof, see Remark.~\ref{rem:compact-support-result-doesnt-hold-at-time-one}). More specifically, the gap between the upper and lower bound grows to infinity as \(t\uparrow1\). The intuitive reason is that the magnitudes of both the score and score divergence takes larger magnitudes in some regions as \(t\uparrow1\) until at \(t=1\) it is undefined in some regions. This may pose problems if one, for example, wishes to use the bounds to use existing results in SMC literature which assume such bounds (such as SMC convergence results briefly exposited in Appendix.~\ref{sec:links-to-SMC-convergence-results}). Similar bound degeneracy can be stated for the other two results (Prop.~\ref{prop:positive-exponents-implies-above-boundedness-MAIN}, Prop.~\ref{prop:one-model-implies-boundedness-MAIN}) if the data distribution is sufficiently non-smooth. 
We leave a more detailed investigation of this phenomena as further work.

\paragraph{Uniqueness only within a family}
Another limitation is that we show uniqueness (Corr.~\ref{corr:positive-exponents-result-uniquness-K-param-family}, Props.~\ref{prop:compact-support-implies-boundedness-MAIN}~\ref{prop:one-model-implies-boundedness-MAIN}) of the three main results (Prop.~\ref{prop:positive-exponents-implies-above-boundedness-MAIN},~\ref{prop:compact-support-implies-boundedness-MAIN},~\ref{prop:one-model-implies-boundedness-MAIN}) only within a restricted \((K+1)\)-parameter family (Eq.~\ref{eq:K-knob-family-in-short-MAIN}).

\subsection{Limitations of experimental results and practical limitations}

As discussed in Sec.~\ref{sec:experimental-results}, the value of \(\knob=1/B\) consistently demonstrated effectiveness over other values in the \(K=1\) setting and in the zero diffusive coefficient setting, but in other settings, in particular the Mixture-of-Gaussians setting elaborated in Sec.~\ref{sec:experimental-results}, we see a value of \(\knob\) between \(1/B\) and \(1\) (the value exposited in \cite{skreta_feynman-kac_2025}) out-performing \(\knob=1/B\). We gave a tentative explanation of these phenomena---in the \(K>1\) setting, \(\knob=1/B\) only gives an upper-bound (Prop.~\ref{prop:positive-exponents-implies-above-boundedness-MAIN}) but in the \(K=1\) setting, it provides both an upper and a lower bound (Prop.~\ref{prop:one-model-implies-boundedness-MAIN}). We leave a more detailed investigation of this as further work.

The WSDEs exposited in our paper in general requires the evaluation of the divergence of a vector field \footnote{More precisely, we need to evaluate the divergence of the score function \(\nabla_{\mathbf{x}}\cdot  \nabla_{\mathbf{x}} \log \pkt(\cdot)\) and, in the case of non-linear (\(\mathbf{F}_t\) not of the form Eq.~\ref{eq:linear-diffusion-model-short-MAIN}) diffusion models, the divergence of the shared drift \(\mathbf{F}_t\). See Eq.~\ref{eq:1-knob-family-in-short-MAIN}.}, which is parameterized by a neural network in most practical settings.
Following \cite{he2026rne}, we bypass the computationally expensive and potentially high variance explicit calculation involving auto-differentiation, and instead implicitly evaluate them using the identity Eq.~\ref{eq:divergence-calculation-trick-backward-forward-ito-integral}. Hence we are prone to the numerical instabilities involved in the use of Eq.~\ref{eq:divergence-calculation-trick-backward-forward-ito-integral} which has been expounded upon in \cite{he2026rne}. 

A practical limitation seen across methods (\cite{du2023reduce,skreta_feynman-kac_2025,he2026rne,lee2025collapse} as well as this work) sampling from the product-of-experts target distribution Eq.~\ref{eq:PoE-MAIN}, and by other methods that modify the original data distributions, such as reward-tilted sampling, is that of distribution shift. 
Such methods must evaluate the learned function approximator (here, the scores \(\nabla_{\mathbf{x}}\log\pkt(\cdot)\)) at points \(\mathbf{x}_t\) lying in regions rarely visited during training. We conjecture that, in our PoE (Eq.~\ref{eq:PoE-MAIN}) / PoIE (Eq.~\ref{eq:PoIE-trajectory-MAIN}) setting, this shift is generally more severe in high-dimensional settings and when \(\sumbetaksmall\neq1\), and that either can degrade sample quality. There is room for further work in both learning algorithms and inference algorithms to ameliorate this problem.

\section*{Acknowledgements}
The authors thanks Hamidreza Behjoo for comments and advice on early drafts of this paper. MC acknowledges support of KAIST Graduate School of AI during his sabbatical visit, where this collaboration started in December 2025-January 2026.

\section*{Disclaimer on the use of large language models}
Large language models (LLMs) were used in this work in three ways: to assist with writing and debugging code, to help run and organize experiments, and to help derive and present the proofs in Appendices~\ref{sec:noising-process-intuition-MAIN} and~\ref{sec:proof-of-Langevin-Infinity}. All LLM-generated code, experimental scripts, and proof steps were reviewed, tested, and verified by the authors, who take full responsibility for the correctness of all reported results and claims. The LLMs did not contribute to the conceptual framing, research questions, or the decision of which results to report.

\bibliography{}
	\bibliographystyle{tmlr}

\appendix

\section{Background: Weighted stochastic differential equations and the Fokker--Planck PDE}
\subsection{Fokker--Planck PDE}
The Fokker--Planck PDE describes the mass distribution obtained from an arbitrary weighted stochastic differential equation (WSDE).
\begin{proposition}[Fokker--Planck PDE for WSDEs]\label{prop:Fokker--Planck-for-WSDEs}
    The WSDE
    \begin{align}
        \begin{cases}\label{eq:wSODE-definition}
        d\mathbf{x}_t &= \mathbf{f}_t(\mathbf{x}_t) \, dt + g_t \, d\mathbf{W}_t \\
        d \log w_t &= h_t(\mathbf{x}_t) \, dt + \mathbf{j}_t(\mathbf{x}_t)\cdot d\mathbf{W}_t
        \\
        \mathbf{x}_0 &\sim p_0\\
        w_\timezero &= 1.
        \end{cases}
    \end{align}
    results in the (in general unnormalized) distribution of mass \(\rho_t(\mathbf{x}), t>0\) satisfying 
    \begin{align}
    \partial_t \rho_t(\mathbf{x}) &= -\nabla_{\mathbf{x}}\cdot(\rho_t \mathbf{f}_t) + \tfrac{g^2_t}{2} \Delta_{\mathbf{x}} \rho_t + (h_t+\tfrac12\sqnorm{\mathbf{j}_t}-g_t\nabla_{\mathbf{x}}\log\rho_t\cdot \mathbf{j}_t - g_t \nabla_{\mathbf{x}} \cdot \mathbf{j}_t)\rho_t, \\
    \label{eq:Fokker--Planck-for-WSDEs}
    \rho_0&=p_0.
    \end{align}
\end{proposition}

Note that unlike SDEs, the marginals in question is not the distribution of particles \(\mathbf{x}_t\) resulting from the WSDE. It is rather the distribution of particles \textit{weighted} by their mass \(w_t\). In other words, it describes the evolving distribution of \textit{mass} itself.

The log Fokker--Planck PDE 
\begin{align}\label{eq:log-Fokker--Planck-for-WSDEs}
\partial_t \log \rho_t(\mathbf{x}) = \frac{\partial_t \rho_t(\mathbf{x})}{\rho_t(\mathbf{x})} &= -\nabla_{\mathbf{x}} \log \rho_t \cdot \mathbf{f}_t - \nabla_{\mathbf{x}} \cdot \mathbf{f}_t + \tfrac{g^2_t}{2} (\nabla_{\mathbf{x}} \cdot \nabla_{\mathbf{x}} \log \rho_t+\sqnorm{\nabla_{\mathbf{x}} \log \rho_t}) 
    \\ \nonumber
    &\quad {} + (h_t+\tfrac12\sqnorm{\mathbf{j}_t}-g_t\nabla_{\mathbf{x}}\log \rho_t\cdot \mathbf{j}_t - g_t \nabla_{\mathbf{x}} \cdot \mathbf{j}_t)
\end{align}
comes useful later.

\paragraph{Criterion for normalized mass distribution}
The distribution \(\rho_t\) of mass resulting from a WSDE is in general unnormalized \(\int \rho_t(\mathbf{x}) d\mathbf{x}\neq1\). 

It is only normalized \(\int \rho_t(\mathbf{x}) d\mathbf{x} = 1\) if the following integral is zero at all time steps 
\begin{equation}\label{eq:normalization-criterion}
    \int \left(h_s(\mathbf{x})+ \frac12 |\mathbf{j}_s(\mathbf{x})|^2\right)\rho_s(\mathbf{x})\,d\mathbf{x}=0, \quad \forall s\in[0,t].
\end{equation}
This is because this is the rate of change of normalization constant
\begin{equation}\label{eq:rate-of-change-of-normalization-constant-for-general-wsdes}
    \frac{d}{dt} \int \rho_t(\mathbf{x})\,d\mathbf{x} = \int \left(h_t(\mathbf{x})+ \frac12 |\mathbf{j}_t(\mathbf{x})|^2\right)\rho_t(\mathbf{x})\,d\mathbf{x}.
\end{equation}

\subsection{WSDEs with zero directional weight}

In this paper we only deal with WSDEs with zero directional weight (\(\mathbf{j}_t(\mathbf{x})=0\) in Eq.~\ref{eq:wSODE-definition}).

There are several properties worth noting with such WSDEs.

For example, there is a Langevin gauge (which is crucially often practically implementable in practice in our context) which does not change the resulting marginal of the particles' mass (Lemma.~\ref{prop:langevin-gauge-MAIN}). 

Lemma.~\ref{prop:langevin-gauge-MAIN} also allows us to derive an errata in \cite{skreta_feynman-kac_2025}.
\begin{remark}\label{rem:skreta-errata}
Prop~D.3 in \cite{skreta_feynman-kac_2025} seemingly violates Lemma.~\ref{prop:langevin-gauge-MAIN}. This stems from a mistake in their Eq.~213. For a correct derivation, one should replace \(\sigma^2_t\) in Eq.~213 with \(\frac{\beta+(1-\beta)2a}{\beta}\sigma^2_t\), which is the correct way to apply their Prop.~C.8 with the modified diffusive coefficient \(\frac{\beta+(1-\beta)2a}{\beta}\sigma^2_t\).
\end{remark}

Lemma.~\ref{prop:langevin-gauge-MAIN} also allows us to simplify our search space of WSDEs by grouping together WSDEs which differ only by a Langevin term.

This is justified by Prop.~\ref{prop:CONJECTURE-Langevin-Infinity-MAIN}.

\section{Background: Several definitions of a diffusion model}
\label{sec:several-definitions-diffusion-models}

We show several definitions of diffusion models with increasing generality.

\subsection{\Lineardiffusionmodel{}: A linear drift term \(+\) a score term}\label{sec:linear-diffusion-model}

This is the narrowest definition of diffusion models, but nonetheless encompasses most diffusion models \cite{ho_denoising_2020,song_score-based_2021} and flow-matching models \cite{lipman_flow_2023} used in practice. It is also referred to as a linear Gaussian interpolant model in the literature \cite{albergo_stochastic_2025}. 

The rigidity allows us to derive strong results with very little assumptions on the data distributions \(\pk_{t=\timeone}\). All results seen in the main text hold under this definition.

We can either define \lineardiffusionmodel{} by its marginals or by its sampling SDE. Either results in the same family. 

\textbf{Marginal-based definition:}
A \lineardiffusionmodel{} follows marginals which are isotropic Gaussian convolutions of the target (data) distribution (Def.~\ref{def:linear-diffusion-model}) up to a rescaling in the \(x\)-axis.
\textbf{SDE-based definition:}
Equivalently, a \lineardiffusionmodel{} follows an SDE with a drift term composed of a linear drift term and a score term (Eq.~\ref{eq:linear-diffusion-model}).

\begin{definition}[\Lineardiffusionmodel{}]\label{def:linear-diffusion-model}
    A \lineardiffusionmodel{} is an SDE or ODE which follows a trajectory from a standard Gaussian \(\pk_{t=\timezero}(\mathbf{x}_0)=\mathcal N(0,I)\) to the target \(\pk_{t=\timeone}(\mathbf{x}_1)\) following a trajectory of intermediate distributions \(\pkt(\mathbf{x}_t)\) which are isotropic Gaussian convolutions of the target distribution up to a scale in the \(x\)-axis.
    \begin{align}\label{eq:linear-interpolant}
            \mathbf{x}_t \overset{d}{=} \mu_t \cdot \mathbf{x}_1 + \sigma_t \cdot z, &\quad {}  \mathbf{x}_1\sim \pktdata(\cdot),\quad z\sim\mathcal N(0,I),
    \end{align}
    where \(\mu_t\) and \(\sigma_t\) are strictly increasing and decreasing functions with boundary conditions
    \begin{align}
            (\mu_{\timeone},\sigma_{\timeone})=(1,0) \text{ and } &(\mu_\timezero,\sigma_\timezero)=(0,1).\label{eq:noising-schedule-boundary-conditions}
    \end{align}
\end{definition}

The marginals \(\pkt(\mathbf{x}_t)\) defined by Eq.~\ref{eq:linear-interpolant} can then be sampled from an SDE (or ODE) in this SDE family:
\begin{align}
\begin{cases} \label{eq:linear-diffusion-model}
d\mathbf{x}_t 
&=\left(- \driftparam_t \mathbf{x}_t + \frac{\noiseparam_t^2}{2} \nabla \log p_t(\mathbf{x}_t) \right) dt + \left(\frac{g_t^2}{2}  \nabla \log p_t(\mathbf{x}_t) \, dt+ g_t\, d\mathbf{W}_t\right) 
\\
&= \left(- \driftparam_t \mathbf{x}_t + \frac{\noiseparam_t^2}{2} \nabla \log p_t(\mathbf{x}_t) \right) dt + \Langevin_{(p_t, g_t)} 
\\
\mathbf{x}_{\timezero} &\sim p_{\timezero} = \mathcal{N}(0,I).
\end{cases}
\end{align}
where 
\begin{equation}\label{eq:formula-for-drift-parameter-and-noise-parameter}
-\driftparam_t = \frac{d}{dt} \log \mu_t, \quad \noiseparam^2_t = -\mu^2_t \frac{d}{dt} ({\sigma^2_t}/{\mu^2_t}).
\end{equation}

The choice of the \emph{noising schedule} \((\mu_t,\sigma_t)\) (and hence also that of \((\driftparam_t,\noiseparam_t)\)) is arbitrary, because \lineardiffusionmodel{}s with different noising schedules can be reparameterized in the \((x,t)\) axes to share their schedules with no additional training. Hence our paper assumes the \(K\) \lineardiffusionmodel{}s to share the same noising schedule. 

The formula for reparameterization is
\begin{align}\label{eq:formula-reparameterizing-linear-diffusion-models}
    \NSR^{(1)}(t^{(1)}) &= \NSR^{(2)}(t^{(2)}), 
    \\
    {\mathbf{x}_{t^{(1)}}^{(1)}}/{\mu^{(1)}_{t^{(1)}}} &= \mathbf{x}_{t^{(2)}}^{(2)}/\mu^{(2)}_{t^{(2)}}
\end{align}
where
\begin{equation}\label{eq:definition-noise-signal-ratio}
    \NSR(t) \doteq \frac{\sigma^2_t}{\mu^2_t}
\end{equation}
is defined as the noise-signal ratio.

\subsection{Nonlinear diffusion model: A general drift term \(+\) a score term}\label{sec:nonlinear-diffusion-model}

This definition of a diffusion model was used by several related works \cite{skreta_feynman-kac_2025,he2026rne}. 

\begin{definition}[\Nonlineardiffusionmodel{}s]
    A \nonlineardiffusionmodel{} maps an initial distribution \(\pk_{t=\timezero}(\mathbf{x}_0)\) to the target distribution \(\pk_{t=\timeone}(\mathbf{x}_1)\) through the trajectory \(\{\pkt\}_{t\in[0,1]}\) using SDE of the form
        \footnote{The minus in front of \(\mathbf{F}_t\) is a relic from the fact that such models can be defined as the reverse process of the so called 'noising process' \(d\mathbf{x}_\tau = \mathbf{F}_\tau(\mathbf{x}_\tau) d\tau + \noiseparam_\tau d\mathbf{W}_\tau\).}
    \begin{align}
    \begin{cases}\label{eq:nonlinear-diffusion-models}
    d\mathbf{x}_t 
    &=\left(- \mathbf{F}_t(\mathbf{x}_t)+ \frac{\noiseparam_t^2}{2} \nabla \log p_t(\mathbf{x}_t) \right) dt + \left(\frac{g_t^2}{2}  \nabla \log p_t(\mathbf{x}_t) \, dt+ g_t\, d\mathbf{W}_t\right) 
    \\
    &= \left(- \mathbf{F}_t(\mathbf{x}_t)+ \frac{\noiseparam_t^2}{2} \nabla \log p_t(\mathbf{x}_t) \right) dt + \Langevin_{(p_t, g_t)} 
    \\
    \mathbf{x}_{\timezero} &\sim p_{\timezero} = \mathcal{N}(0,I).
    \end{cases}
    \end{align}
    with degree of freedom \(g_t\geq0\).
\end{definition}

We obtain \lineardiffusionmodel{}s (Sec.~\ref{sec:linear-diffusion-model}) when the nonlinear drift term \(\mathbf{F}_t(\mathbf{x}_t)\) is linear \(\driftparam_t\mathbf{x}_t\).

Previous works on product-of-experts use this definition \cite{skreta_feynman-kac_2025, he2026rne}. They further assume that the \(K\) diffusion models share the same nonlinear drift function \(\mathbf{F}_t(\mathbf{x}_t)\) and the diffusion coefficient \(\noiseparam_t\).

\begin{remark}[Why we restrict to a fixed shared drift \(-\mathbf{F}_t\)]\label{rem:why-fix-shared-drift}
In the families Eq.~\ref{eq:1-knob-family-in-short-MAIN} and Eq.~\ref{eq:K-knob-family-in-short-MAIN}, the knob acts only on the score term, while the shared drift \(-\mathbf{F}_t\) is kept fixed. This choice serves two purposes.

\emph{(i) A simple \logweighingfunction{}.} With \(-\mathbf{F}_t\) fixed, \(\mathbf{F}_t\) enters the \logweighingfunction{} only through the divergence term \((B-1)\nabla_{\mathbf{x}}\cdot\mathbf{F}_t\) (Eq.~\ref{eq:shared-nonlinear-drift-divergence-term-MAIN}). For \lineardiffusionmodel{}s (\(\mathbf{F}_t(\mathbf{x}_t)=\driftparam_t\mathbf{x}_t\)) this term is the constant \((B-1)\driftparam_t d\). Our upper bounds (Sec.~\ref{sec:main-results-MAIN}) rely on this structure.

\emph{(ii) Invariance to the noising schedule.} For \lineardiffusionmodel{}s, keeping \(-\driftparam_t\mathbf{x}_t\) fixed also makes the family invariant to the (arbitrary) choice of noising schedule \((\mu_t,\sigma_t)\) (Eq.~\ref{eq:linear-interpolant}). To see this, change to the schedule-free coordinates
\begin{equation}\label{eq:schedule-free-coordinates}
    \mathbf{y}\doteq\mathbf{x}/\mu_t, \qquad u\doteq-\NSR(t)
\end{equation}
(cf.\ Eq.~\ref{eq:formula-reparameterizing-linear-diffusion-models}), where the clock \(u\) increases with \(t\). The term \(-\driftparam_t\mathbf{x}_t\,dt\) is exactly absorbed by the change of variables, as \(-\driftparam_t=\tfrac{d}{dt}\log\mu_t\) (Eq.~\ref{eq:formula-for-drift-parameter-and-noise-parameter}). Using also \(\noiseparam_t^2\,dt=\mu_t^2\,du\), Eq.~\ref{eq:1-knob-family-in-short-MAIN} becomes
\begin{equation}\label{eq:1-knob-family-schedule-free}
    d\mathbf{y}=\tfrac{\knob+\zeta_t^2}{2}\,\nabla_{\mathbf{y}}\log q^{\textrm{PoIE}}_u(\mathbf{y})\,du+\zeta_t\,d\tilde{\mathbf{W}}_u,
    \qquad \zeta_t\doteq g_t/\noiseparam_t,
\end{equation}
where \(\tilde{\mathbf{W}}_u\) is a standard Brownian motion in the clock \(u\) and \(q^{\textrm{PoIE}}_u\) is the PoIE trajectory (Eq.~\ref{eq:PoIE-trajectory-MAIN}) expressed in \(\mathbf{y}\). Eq.~\ref{eq:1-knob-family-schedule-free} depends on the schedule only through the time change \(t\mapsto u\). Hence both the knob \(\knob\) and the rescaled Langevin strength \(\zeta_t\) have a schedule-independent meaning, whereas \(g_t\) itself does not. A knob on \(-\driftparam_t\mathbf{x}_t\) would instead leave a residual term proportional to \(\tfrac{d}{dt}\log\mu_t\), whose meaning depends on the schedule.
\end{remark}

\section{Problem setting}

\subsection{Problem setting: sampling from the product-of-experts}
\label{sec:poe}

We assume we have access to \(K\) diffusion models of the form in Sec.~\ref{sec:nonlinear-diffusion-model}, with the \(K\) models share the vector field \(\mathbf{F}_t(\mathbf{x}_t)\) as well as the diffusion coefficient \(\noiseparam_t\).
Our aim is to sample from the weighted product of their final time-step marginals, which we call the \textbf{product-of-experts} target
\begin{align}
    p^{\PoE} &\doteq \frac{1}{Z} \prod_k (\pk_{data})^{\beta_k},
\end{align}
where \(\sumbetak>0\) and \(Z\) is the normalization constant.

\subsection{Design choice: product-of-intermediate-experts trajectory}\label{sec:poie}

Previous approaches chose to follow a key design decision. They chose \cite{du2023reduce,skreta_feynman-kac_2025,he2026rne,lee2025collapse} to sample along what we call the \textbf{product-of-intermediate-experts} trajectory (or generalizations such as the time varying exponents \(\beta_{k,t}\) \cite{lee2025collapse}), defined by multiplying the intermediate marginals of the \(K\) models at every timestep
\begin{align}\label{eq:PoIE-marginals}
    \pPoIE_{t}
    \doteq \frac{1}{Z_t}
    \prod^K_{k=1} (\pkt)^{\beta_k}
    , \quad{t\in[\timezero,\timeone]}.
\end{align}
which in particular has the starting point of a Gaussian and the endpoint of the data/target distribution
\begin{align}
    \pPoIE_{t=\timezero}&=\mathcal N(0,\,\frac{1}{\textstyle\sum_k\beta_k} \cdot I)
    \\
    \pPoIE_{t=\timeone}&=p^{\PoE}.
\end{align}
We also choose the PoIE trajectory (with time-constant exponents \(\beta_k\)) in our work, and leave generalizations to other choices of trajectories as further work. 

\subsection{All WSDEs sampling from PoIE}

The PoIE trajectory (Sec.~\ref{sec:poie}) results in a particularly clean expression of \(\partial_t \log \pPoIE_t(\mathbf{x})\)
\begin{align}\label{eq:general-poie-log-partial-derivative}
    \partial_t \log \pPoIE_t(\mathbf{x}) = -\partial_t \log Z_t + \sum_k \beta_k \partial_t \log \pkt(\mathbf{x}),
\end{align}
which in turn leads to a particularly large family of computationally tractable of WSDEs which sample from it.
This clarifies why it has been the common design choice among previous papers sampling from the product of diffusion models \cite{skreta_feynman-kac_2025,he2026rne,lee2025collapse}.

\subsubsection{All WSDEs sampling from PoIE of \nonlineardiffusionmodel{}s}

For consistency with previous literature \cite{skreta_feynman-kac_2025,he2026rne}, let us use the definition of the diffusion model given in Sec.~\ref{sec:nonlinear-diffusion-model}, i.e. \nonlineardiffusionmodel{}s. 

We can categorize all WSDEs (of the form Eq.~\ref{eq:wSODE-definition}) which sample from the PoIE trajectory of \(K\) \nonlineardiffusionmodel{}s by plugging in their Eq.~\ref{eq:general-poie-log-partial-derivative} into the Fokker--Planck PDE (Prop.~\ref{prop:Fokker--Planck-for-WSDEs}) then rearranging. Here, we fix \((\mathbf{f}_t,g_t,\mathbf{j}_t)\) then observe the \(h_t\) required to sample from the PoIE trajectory.

\begin{proposition}\label{prop:all-WSDEs-sampling-from-PoIE-of-nonlinear-diffusion-models}
[All WSDEs which sample from PoIE of \nonlineardiffusionmodel{}s]
Given \(K\) \nonlineardiffusionmodel{}s (Eq.~\ref{eq:nonlinear-diffusion-models}), their PoIE trajectory (Eq.~\ref{eq:PoIE-marginals}) can be sampled by the following family of WSDEs (Eq.~\ref{eq:wSODE-definition}).
    \begin{align}
        \begin{cases} \label{eq:all-WSDEs-sampling-from-PoIE-of-nonlinear-diffusion-models}
            d\mathbf{x}_t 
            &= 
            \proposaldrift(\mathbf{x}) \, dt + \Langevin_{(\pPoIE_t, \red{g_t})}
            \\
            &=
            \proposaldrift(\mathbf{x}) \, dt + \left[\frac{\red{g_t^2}}{2}
            \nabla_{\mathbf{x}}\log \pPoIE_t(\mathbf{x}) \, dt
            + {\red{g_t}}\, d\mathbf{W}_t \right]
            \\ 
            d\log w_t &=  h^{\mathbf{f}_t,g_t,\mathbf{j}_t}_t(\mathbf{x}_t)\,dt + \mathbf{j}_t(\mathbf{x}_t)\,d\mathbf{W}_t
        \end{cases}
    \end{align}
    with initial conditions 
    \(\mathbf{x}_0 \sim \mathcal N(0,\frac{1}{\sumbetak} \cdot I) = \pPoIE_\timezero\), 
    \(w_\timezero = 1\), and the \logweighingfunction{}
    \begin{align}\label{eq:log-weighing-function-of-all-WSDEs-sampling-from-PoIE-of-nonlinear-diffusion-models}
        h^{\mathbf{f}_t,g_t,\mathbf{j}_t}_t(\mathbf{x}_t)&\doteq
        \nabla_{\mathbf{x}} \log \pPoIE_t \cdot \mathbf{F}_t(\mathbf{x}_t) +  (\textstyle\sum_k\beta_k) \nabla_{\mathbf{x}} \cdot \mathbf{F}_t(\mathbf{x}_t)
        \\&\quad {}  \nonumber 
        - \tfrac{\noiseparam^2_t}{2} \sum_k \beta_k \left(\sqnorm{\nabla_{\mathbf{x}} \log \pkt} + \nabla_{\mathbf{x}} \cdot \nabla_{\mathbf{x}} \log \pkt \right)
        \\&\quad {}  \nonumber 
        +\nabla_{\mathbf{x}} \log \pPoIE \cdot \proposaldrift + \nabla_{\mathbf{x}} \cdot \proposaldrift
        \\&\quad {}  \nonumber 
        + \red{g_t} \left(\nabla_{\mathbf{x}} \log \pPoIE \cdot \mathbf{j}_t + \nabla_{\mathbf{x}} \cdot \mathbf{j}_t \right)- \tfrac12 \sqnorm{\mathbf{j}_t} 
    \end{align}
    \LineBetweenStatementAndProof
    \begin{proof}
        One can take the time derivative of the product-of-intermediate-experts marginals Eq.~\ref{eq:PoIE-marginals}, plug in the \(K\) individual Fokker--Planck equations of the \(K\) diffusion models (obtained by plugging in Eq.~\ref{eq:nonlinear-diffusion-models} to Eq.~\ref{eq:Fokker--Planck-for-WSDEs}), then observe that this is equal to the Fokker--Planck equation arising from Eq.~\ref{eq:log-weighing-function-of-all-WSDEs-sampling-from-PoIE-of-nonlinear-diffusion-models}. To be more precise, following this line of derivation will yield a normalized \logweighingfunction{} \(\tilde h^{\mathbf{f}_t,g_t,\mathbf{j}_t}_t(\cdot)=h^{\mathbf{f}_t,g_t,\mathbf{j}_t}_t(\cdot)-\partial_t \log Z_t\) with an extra \textit{intractable} constant term \(-\partial_t \log Z_t\). This normalization factor can be omitted whilst preserving the correctness, as usually done in SMC.
    \end{proof}
\end{proposition}

\paragraph{Translation to \cite{he2026rne}'s notation}

The family of WSDEs obtained here is equivalent to the WSDEs obtained using the method of \cite{he2026rne} with the PoIE trajectory.
The translation from their notation to the one in Prop.~\ref{prop:all-WSDEs-sampling-from-PoIE-of-nonlinear-diffusion-models} is
\begin{align}
    \begin{cases}\label{eq:translation-from-He26-to-our-notation}
    -\mathbf{a}^{\text{He26}}_t &\equiv \mathbf{f}^{\text{Ours}}_t+\tfrac{(g^{\text{Ours}}_t)^2}{2}\nabla_{\mathbf{x}} \log \pPoIE_t, 
    \\
    \epsilon^{\text{He26}}_t &\equiv g^{\text{Ours}}_t
    \\
    \mathbf{b}^{\text{He26}}_t &\equiv  \mathbf{a}^{\text{He}}_t- g^{\text{Ours}}_t \mathbf{j}^{\text{Ours}}_t + (g^{\text{Ours}}_t)^2\nabla_{\mathbf{x}}\log\pPoIE_t, 
    \\
    \mathbf{f}^{\text{He26}}_t &\equiv \mathbf{F}^{\text{Ours}}_t.
    \end{cases}
\end{align}

But our derivation did not need to invoke path measures and Radon--Nikodym derivatives thereof, but only used the definition of the PoIE marginals and the Fokker--Planck PDE.

As shown in \cite{he2026rne}, almost all members of this family can be sampled from computationally relatively efficiently (see Appendix.~\ref{sec:TRACTABLE-which-WSDEs-sampling-from-PoIE-of-nonlinear-diffusion-models-ARE-TRACTABLE}). In particular, if we use a stochastic differential equation \(g_t>0\), then we can sample from any choice of \(\mathbf{f}_t,\mathbf{j}_t\) for which we can obtain function evaluations of, without needing to calculate the divergences of \(\mathbf{f}_t,\mathbf{j}_t,\mathbf{F}_t\).

\section{Computational tractability of WSDEs in Prop.~\ref{prop:all-WSDEs-sampling-from-PoIE-of-nonlinear-diffusion-models}}
\label{sec:practical-tricks-to-sample-from-WSDE}\label{sec:TRACTABLE-which-WSDEs-sampling-from-PoIE-of-nonlinear-diffusion-models-ARE-TRACTABLE}

\cite{he2026rne} generalized the results in \cite{skreta_feynman-kac_2025} by showing a much larger family of WSDEs can be used to sample from the product-of-experts of diffusion models. Their main innovation related to a novel method of computing the integral of a vector field divergence term in an Ito integral (such as \(\nabla_{\mathbf{x}}\cdot \mathbf{f}_t(\cdot)\), \(\nabla_{\mathbf{x}}\cdot \mathbf{F}_t(\cdot)\) and \(\nabla_{\mathbf{x}}\cdot \mathbf{j}_t(\cdot)\) in Eq.~\ref{eq:log-weighing-function-of-all-WSDEs-sampling-from-PoIE-of-nonlinear-diffusion-models}). The naive method will consist of using computationally expensive auto-differentiation based methods, such as Hutchinson's estimators. \cite{he2026rne} showed that we can instead use a particular identity 
\begin{proposition}
    \begin{align}\label{eq:divergence-calculation-trick-backward-forward-ito-integral}
        \int^{t_1}_{t_2} g_t\,\nabla_{\mathbf{x}} \cdot v_t(\mathbf{x}_t) \,\overset{\rightarrow}{dt} &= \int^{t_1}_{t_2} \,v_t(\mathbf{x}_t) \cdot \overset{\leftarrow}{d\mathbf{W}_t} - \int^{t_1}_{t_2} \,v_t(\mathbf{x}_t) \cdot \overset{\rightarrow}{d\mathbf{W}_t},
    \end{align}
    holds for any Ito process \(d\mathbf{x}_t = \mathbf{f}_t(\mathbf{x}_t)\,dt + g_t \, d\mathbf{W}_t\) and any vector field \(v_t\).
\end{proposition}
to instead implicitly calculate the vector field divergence term (\(\nabla_{\mathbf{x}}\cdot \mathbf{f}_t(\cdot)\), \(\nabla_{\mathbf{x}}\cdot \mathbf{F}_t(\cdot)\) and \(\nabla_{\mathbf{x}}\cdot \mathbf{j}_t(\cdot)\)).

This in turn implies that for \textit{any} choice of vector fields \(\mathbf{F}_t, \mathbf{f}_t, \mathbf{j}_t\) and a positive diffusive coefficient \(g_t>0\), we simulate the WSDE in Prop.~\ref{prop:all-WSDEs-sampling-from-PoIE-of-nonlinear-diffusion-models} without expensive auto-differentiation based algorithms.
Note however, that using this identity (Eq.~\ref{eq:divergence-calculation-trick-backward-forward-ito-integral}) has the disadvantage of potential numerical instability which is discussed in \cite{he2026rne}.
For \(g_t=0\), one cannot use Eq.~\ref{eq:divergence-calculation-trick-backward-forward-ito-integral}, hence one must resort to calculating divergence terms explicitly (e.g. using auto-differentiation), or choosing a WSDE without divergence terms (\(\knob=1\) in Eq.~\ref{eq:1-knob-family-logweighingfunction-in-short-MAIN}, as introduced in \cite{skreta_feynman-kac_2025}).

\section{Further remarks on regularity condition.~\ref{cond:bounded-score-divergence-MAIN}}\label{sec:regularity-condition-bounded-score-divergence}

Recall regularity condition.~\ref{cond:bounded-score-divergence-MAIN} used in Prop.~\ref{prop:positive-exponents-implies-above-boundedness-MAIN} and Prop.~\ref{prop:one-model-implies-boundedness-MAIN}:
\begin{condition}[Bounded score divergence at a given time
    ]
    \label{cond:bounded-score-divergence}
    All \(k\) models have bounded score divergence 
    \begin{equation}
        \sup_{\mathbf{x}\in\mathbb R^d} \left|\nabla_{\mathbf{x}} \cdot \nabla_{\mathbf{x}} \log \pkt(\mathbf{x})\right| < \infty.
    \end{equation}
\end{condition}

We propose that when using \lineardiffusionmodel{}s (Eq.~\ref{eq:linear-diffusion-model-short-MAIN}), this regularity condition can be expected to hold at all \(t\in[0,1)\) for non-pathological data (target) distributions.
We now justify why. Recall that for \lineardiffusionmodel{}s, the non-terminal time marginals (\(\pkt(\cdot)\) for \(t<1\)) are Gaussian convolutions of the target marginal \(\pktdata\) (see Eq.~\ref{eq:linear-interpolant}). This immediately gives us a lower bound---using Tweedie's identity (Eq.~\ref{eq:Laplacian-is-Covariance}) we have \(\nabla_{\mathbf{x}} \cdot \nabla_{\mathbf{x}} \log p(\mathbf{x}_t)\geq-d/\sigma^2_t\). Hence this regularity condition becomes equivalent to only demanding that we have an upper bound \(\sup_{\mathbf{x}} \nabla_{\mathbf{x}} \cdot \nabla_{\mathbf{x}} \log \pkt < \infty, \forall t\in[0,1)\). Although it is not hard to find a counterexample \(p\) for which \(p'=N(0,I)*p\) does not satisfy \(\sup_{\mathbf{x}} \nabla_{\mathbf{x}} \cdot \nabla_{\mathbf{x}} \log p' < \infty\), we conjecture such distributions are contrived enough to not be encountered in practice. 

In particular, a \lineardiffusionmodel{}  with a compact data distribution \(\pktdata\) is \textbf{sufficient} for RC.~\ref{cond:bounded-score-divergence} to hold in \(\forall t\in[0,1)\) (this is a part of the proof of Prop.~\ref{prop:compact-support-implies-boundedness-MAIN} in Appendix.~\ref{sec:proof-of-compact-support-result}).

\section{Proof of compact support result Prop.~\ref{prop:compact-support-implies-boundedness-MAIN}}\label{sec:proof-of-compact-support-result}

\paragraph{Compact support data distributions leads to boundedness}

\begin{proposition}[Prop.~\ref{prop:compact-support-implies-boundedness-MAIN}]\label{prop:compact-support-implies-boundedness}
    
    If we have linear diffusion models \(\mathbf{F}_t(\mathbf{x}_t)=\driftparam_t\mathbf{x}_t\), and all \(K\) target distributions \(\pk_{data}\) are compact, then \(\knob=1/B\) results in an upper-and-lower bounded \logweighingfunction{} \(h^{\knob=1/B}_t(\mathbf{x})\) (Eq.~\ref{eq:1-knob-family-logweighingfunction-in-short-MAIN}) at every non-terminal time \(t\in[\timezero,\timeone)\),
    i.e., for all \(t<1\), there exist \(m_t,M_t\in\mathbb R\) such that 
    \begin{equation}
        -\infty < m_t < h^{\knob=1/B}_t(\mathbf{x}) < M_t < \infty
    \end{equation}
    for all \(\mathbf{x}\).

    Furthermore, \(\knob=1/B\) in the \((1+1)\)-parameter family is equivalent to \(\knob_k=1/B,\forall k\) in the \((K+1)\)-parameter family. One can show that this is the \textbf{unique} set of knobs in the \((K+1)\)-parameter family Eq.~\ref{eq:K-knob-family-in-short-MAIN} which guarantees upper and lower boundedness.
    \LineBetweenStatementAndProof
     
    \begin{proof}
        Recall that a \lineardiffusionmodel{} has intermediate time marginals \(\pkt\) which are isotropic Gaussian convolutions of the final time marginal (the data distribution) \(\pktdata\) (Sec.~\ref{sec:linear-diffusion-model}). 
        This \textit{allows} us to rewrite the score and its divergence using the Tweedie's identities
        \begin{equation}\tag{Prop.~\ref{prop:Tweedies-identity}}
            \nabla_{\mathbf{x}} \log \pkt(\mathbf{x}) = \frac{\mu_t \mathbb E^k[\mathbf{x}_1|\mathbf{x}_t=x]-x}{\sigma^2_t}
        \end{equation}
        and
        \begin{align}\tag{Prop.~\ref{prop:Laplacian-is-Covariance}}
            \nabla_{\mathbf{x}} \cdot \nabla_{\mathbf{x}} \log \pkt(\mathbf{x}) 
            &= \Tr \nabla_{\mathbf{x}}^2 \log \pkt(\mathbf{x}) 
            \\
            &= \frac{\mu^2_t \, \Tr \Cov^k [\mathbf{x}_1 | \mathbf{x}_t=x] }{\sigma^4_t} - \frac{ d}{\sigma^2_t},
        \end{align}
        where the expectation is taken with respect to the joint distribution described by Eq.~\ref{eq:linear-interpolant}: \(\mathbf{x}_1\sim\pktdata(\cdot),\ \mathbf{x}_t \overset{d}{=} \mu_t \cdot \mathbf{x}_1 + \sigma_t \cdot z, \ \ z\sim\mathcal N(0,I)\).

        The \logweighingfunction{} Eq.~\ref{eq:1-knob-family-logweighingfunction-in-short-MAIN} with \(\mathbf{F}_t(\mathbf{x}_t)=\driftparam_t\mathbf{x}_t\)
        \begin{equation}
        \tag{\ref{eq:1-knob-family-logweighingfunction-in-short-MAIN} with Eq.~\ref{eq:linear-diffusion-model-short-MAIN}}
            \begin{aligned}
                h^{\red{\knob}}_t(\mathbf{x}_t) 
                    &=\tfrac{\noiseparam^2_t}{2}(\red{\knob}-\tfrac1B) \sqnorm{\nabla_{\mathbf{x}} \log \pPoIE_t(\mathbf{x}_t)}
                    \\&\quad {} 
                    -\tfrac{\noiseparam^2_t}{4\sumbetashort}\sum_{j,k}  \beta_j \beta_k \inlinesqnorm{\nabla_{\mathbf{x}} \log p^{j}_t(\mathbf{x}_t) - \nabla_{\mathbf{x}} \log \pkt(\mathbf{x}_t)} 
                    \\&\quad {}  
                    + \tfrac{\noiseparam^2_t}{2}\left(-1+\red{\knob}\right) \sum_k \beta_k \nabla_{\mathbf{x}} \cdot \nabla_{\mathbf{x}} \log \pkt(\mathbf{x}_t)
                    \\&\quad {}  
                    + \driftparam_t(B-1)d 
            \end{aligned}
        \end{equation}
        can be written in terms of these identities to obtain
        \begin{equation}
        \label{eq:1-knob-family-logweighingfunction-Tweedies-rewritten}
            \begin{aligned}
                h^{\red{\knob}}_t(\mathbf{x}_t) 
                    &=\tfrac{\noiseparam^2_t}{2}(\red{\knob}-\tfrac1B) \sqnorm{\sum_k \beta_k \frac{\mu_t \mathbb E^k[\mathbf{x}_1|\mathbf{x}_t]-\mathbf{x}_t}{\sigma^2_t}}
                    &(\text{squared-score term)}
                    \\&\quad {} 
                    -\frac{\noiseparam^2_t}{4\sumbetashort} \frac{\mu^2_t}{\sigma^4_t} \sum_{j,k}  \beta_j \beta_k \sqnorm{\mathbb E^j[\mathbf{x}_1|\mathbf{x}_t]-\mathbb E^k[\mathbf{x}_1|\mathbf{x}_t]} 
                    &(\text{expectation-disagreement term)}
                    \\&\quad {}  
                    + \frac{\noiseparam^2_t}{2} \frac{\mu^2_t}{\sigma^4_t}  \left(-1+\red{\knob}\right) \sum_k \beta_k \Tr \Cov^k [\mathbf{x}_1 | \mathbf{x}_t] 
                    &(\text{trace of covariances term)}
                    \\&\quad {}  
                    + \driftparam_t(B-1)d 
                    - \frac{\noiseparam^2_t}{2} \frac{d}{\sigma^2_t}  \left(-1+\red{\knob}\right) B
                    &(\text{one additional constant w.r.t }\mathbf{x}_t).
            \end{aligned}
        \end{equation}
        
        Recall that when \(\knob=1/B\), the squared-score term vanishes, leaving us with the other three terms.
        Both the conditional expectation and conditional covariance are bounded in the sense that \(\mathbf{x}_1\) can take values only in a compact region. 
        More precisely, define the diameters \(D_{k}\doteq \sup_{\mathbf{x},\mathbf{y}\in\text{supp}(\pktdata)}\lVert \mathbf{x}-\mathbf{y}\rVert\) and \(D_{j,k}\doteq \sup_{\mathbf{x},\mathbf{y}\in\text{supp}(\pktdata)\cup \text{supp}(p^{(j)}_{t=1})}\lVert \mathbf{x}-\mathbf{y}\rVert\). Then
        \begin{equation}\label{eq:something-leq-Djk}
            0\leq \sqnorm{\mathbb E^j[\mathbf{x}_1|\mathbf{x}_t]-\mathbb E^k[\mathbf{x}_1|\mathbf{x}_t]} \leq {D^2_{j,k}}
        \end{equation}
        and
        \begin{equation}\label{eq:something-leq-Dk}
            0 \leq \Tr \Cov^k [\mathbf{x}_1 | \mathbf{x}_t] \leq \frac{D^2_k}2.
        \end{equation}
        For simplicity let us now use \(D\), defined as the maximum of all \(D_i\)s and \(D_{i,j}\)s---in other words the diameter of all the data distributions combined.
        
        Eq.~\ref{eq:something-leq-Djk} and Eq.~\ref{eq:something-leq-Dk} gives us the above and below boundedness of \(h^{\knob\red{=1/B}}_t\). For all \(x\), we have
        \begin{equation}
            -\infty<m_t<h^{\knob\red{=1/B}}_t(\mathbf{x})<M_t<\infty
        \end{equation}
        for some \(m_t,M_t\) for which 
        \begin{equation}\label{eq:final-result-bound-compact-support}
        |M_t-m_t| \leq  \frac{\noiseparam^2_t\mu^2_t}{\sigma^4_t}\left(\frac {B'^2}{4B} + \frac{|1-B|\,B'}{4B}\right)D^2
        \end{equation}
        where \(B'\doteq\sum_k|\beta_k|\)
        \paragraph{Uniqueness:} Write \(L\doteq\sum_k\knob_k\beta_k\) and \(\mathbf{m}_k(\mathbf{x})\doteq\mathbb E^k[\mathbf{x}_1|\mathbf{x}_t=\mathbf{x}]\) where conditional expectation is taken with respect to the joint distribution defined in Eq.~\ref{eq:linear-interpolant}. Substituting Tweedie's identity \(\nabla_{\mathbf{x}}\log\pkt=(\mu_tm_k-\mathbf{x})/\sigma_t^2\) into the \logweighingfunction{} of the \((K+1)\)-parameter family (obtained by plugging in \(\mathbf{j}_t=0, g_t=0, \mathbf{f}_t=- \mathbf{F}_t(\mathbf{x}_t)+ \tfrac{\noiseparam^2_t}{2}\sum_k \red{\knob_k}\beta_k\nabla_{\mathbf{x}} \log \pkt(\mathbf{x}_t)\) from Eq.~\ref{eq:K-knob-family-in-short-MAIN} into Eq.~\ref{eq:log-weighing-function-of-all-WSDEs-sampling-from-PoIE-of-nonlinear-diffusion-models}) gives \(h^{\{\knob_k\}_k}_t(\mathbf{x})=\tfrac{\noiseparam_t^2}{2\sigma_t^4}\big(B(L-1)\sqnorm{\mathbf{x}}+\mu_t\langle \mathbf{x},\textstyle\sum_k\beta_kc_k\,\mathbf{m}_k(\mathbf{x})\rangle\big)+(\text{bounded})\), where \(c_k\doteq2-L-B\knob_k\) and both \(\lVert \mathbf{m}_k\rVert\) and the remainder are bounded. If \(L\neq1\) (in the \((1+1)\)-parameter family, exactly \(\knob\neq1/B\)), the quadratic term dominates for any compact data, e.g.\ \(\pktdata=p_c,\forall k\). If \(L=1\) but \(\knob_j\neq1/B\) for some \(j\) with \(\beta_j\neq0\), then \(\sum_k\beta_kc_k=0\) and \(c_j\neq0\). Take translated copies \(\pktdata(\bullet)=p_c(\bullet-\mathbf{a}_k)\) with \(\mathbf{a}_j=\mathbf{e}\) a unit vector and \(\mathbf{a}_k=0\) otherwise. If we observe a noisy point \(\mathbf{x}_t=r\mathbf{e}'\) far out in direction \(\mathbf{e}'\), the posterior over the denoised point \(\mathbf{x}_1\) accumulates on the part of the support that reaches furthest in that direction, i.e., \(\langle \mathbf{e}',\mathbf{m}_k(r\mathbf{e}')\rangle\to\langle \mathbf{a}_k,\mathbf{e}'\rangle+\max_{y\in\operatorname{supp}p_c}\langle \mathbf{y},\mathbf{e}'\rangle\) as \(r\to\infty\). The \(p_c\)-dependent parts then cancel, so \(h^{\{\knob_k\}_k}_t(\pm r\mathbf{e})=\tfrac{\noiseparam_t^2\mu_t}{2\sigma_t^4}\,r\,(\pm\beta_jc_j+o(1))+O(1)\), which is unbounded both above and below. For \(p_c=\delta_0\) this is exactly affine in \(\mathbf{x}\).
    \end{proof}
\end{proposition}

\subsection{Remarks on the compact support result}\label{sec:remarks-comments-on-the-compact-support-result}

The following is an important limitation of Prop.~\ref{prop:compact-support-implies-boundedness-MAIN}.

\begin{remark}[Compact support result bound diverges as \(t\uparrow1\)]\label{rem:compact-support-result-doesnt-hold-at-time-one}
    See the \logweighingfunction{} Eq.~\ref{eq:1-knob-family-logweighingfunction-Tweedies-rewritten} and the resulting bound Eq.~\ref{eq:final-result-bound-compact-support}. \(\frac{\noiseparam^2_t \mu^2_t}{\sigma^4_t}\) is unbounded as \(t\uparrow1\)

    This can be seen by rewriting
    \(\frac{\noiseparam^2_t \mu^2_t}{\sigma^4_t} = \tfrac{d}{dt}\SNR_t\) where \(\SNR_t\) (referred to as signal-noise-ratio in the literature) is \(\SNR_t\doteq\tfrac{\mu^2_t}{\sigma^2_t}\). Then \(\int^1_{t=1-\epsilon} \frac{\noiseparam^2_t \mu^2_t}{\sigma^4_t} \, dt = \SNR_{t=1}-\SNR_{t=1-\epsilon}\) is infinity due to the boundary condition \(\mu_{t=1}=1\) and \(\sigma_{t=1}=0\) in Eq.~\ref{eq:noising-schedule-boundary-conditions}.

    Hence the width \((M_t-m_t)\) of our bound  Eq.~\ref{eq:final-result-bound-compact-support} is unbounded as \(t\uparrow1\), 
    and in particular it diverges fast enough to be not integrable \(\int^1_{t=1-\epsilon} (M_t-m_t)\,dt=\infty\).
\end{remark}

The above remark intuitively makes sense, because both the score and the divergence of the score will, loosely speaking, take infinite values. Hence the \logweighingfunction{}, composed of these values, can also diverge.

\section{Preliminaries: Tweedie's Identities}
\label{sec:Tweedie-identity}

\paragraph{Relevance to our setting.}
Because the \lineardiffusionmodel{} setting (Eq.~\ref{eq:linear-interpolant}) results in Gaussian-convolved marginals, we can apply these identities in this setting.

    \begin{lemma}[Tweedie's identity, first derivative \cite{efron2011tweedie, luo_diffusion_2021}]\label{prop:Tweedies-identity}
        Let \(\mathbf{x}_1\) be a random variable, and define \(\mathbf{x}_t \doteq \mu_t  \mathbf{x}_1 + \sigma_t \cdot z\) where \(z \sim \mathcal N(0,I)\).
        Then, the score of \(\mathbf{x}_t\) satisfies 
        \begin{equation}\label{eq:Tweedies-identity}
            \nabla_{\mathbf{x}} \log p(\mathbf{x}_t) = \frac{\mu_t \, \mathbb E[\mathbf{x}_1 | \mathbf{x}_t] - \mathbf{x}_t}{\sigma^2_t}
        \end{equation}
        
        \begin{proof}
            \begin{align}
                &\quad {}  \ \,
                \nabla_{\mathbf{x}_t} \log p(\mathbf{x}_t) 
                \\
                &= \frac{1}{p(\mathbf{x}_t)} \nabla_{\mathbf{x}_t} p(\mathbf{x}_t) 
                \\
                &= \frac{1}{p(\mathbf{x}_t)} \nabla_{\mathbf{x}_t} \int p(\mathbf{x}_1)p(\mathbf{x}_t|\mathbf{x}_1)\ d\mathbf{x}_1 
                \\
                &= \frac{1}{p(\mathbf{x}_t)} \int p(\mathbf{x}_1)\nabla_{\mathbf{x}_t} p(\mathbf{x}_t|\mathbf{x}_1)\ d\mathbf{x}_1
                \\
                &= \frac{1}{p(\mathbf{x}_t)} \int p(\mathbf{x}_1) p(\mathbf{x}_t|\mathbf{x}_1) \nabla_{\mathbf{x}_t} \log p(\mathbf{x}_t|\mathbf{x}_1)\ d\mathbf{x}_1
                \\
                &= \int p(\mathbf{x}_1|\mathbf{x}_t) \nabla_{\mathbf{x}_t} \log p(\mathbf{x}_t|\mathbf{x}_1)\ d\mathbf{x}_1
                \\
                &= \int p(\mathbf{x}_1|\mathbf{x}_t) \frac{\mu_t \cdot \mathbf{x}_1 - \mathbf{x}_t}{\sigma^2_t}\ d\mathbf{x}_1
                \\
                &= \frac{\mu_t \, \mathbb E[\mathbf{x}_1 | \mathbf{x}_t] - \mathbf{x}_t}{\sigma^2_t}
            \end{align}
        \end{proof}
    \end{lemma}
    Intuitively, the score points in the direction of the most likely location of the original signal.

    \begin{lemma}[Tweedie's identity, second derivative]\label{prop:Laplacian-is-Covariance}
        Let \(\mathbf{x}_1\) be a random variable, and \(\mathbf{x}_t\) to be defined as \(\mathbf{x}_t \doteq \mu_t  \mathbf{x}_1 + \sigma_t \cdot z\) where \(z \sim \mathcal N(0,I)\).
        Then, the log-hessian of \(p(\mathbf{x}_t)\) satisfies 
        \begin{align}
            \nabla_{\mathbf{x}}^2 \log p(\mathbf{x}_t) 
            &= \frac{\mu_t \, \nabla_{\mathbf{x}}^\top\mathbb E[\mathbf{x}_1 | \mathbf{x}_t] }{\sigma^2_t} - \frac{ I}{\sigma^2_t} 
            \\
            &= \frac{\mu^2_t \, \Cov [\mathbf{x}_1 | \mathbf{x}_t] }{\sigma^4_t} - \frac{ I}{\sigma^2_t} 
        \end{align}

        and hence
        \begin{align}
            \nabla_{\mathbf{x}} \cdot \nabla_{\mathbf{x}} \log p(\mathbf{x}_t) 
            &= \Tr \nabla_{\mathbf{x}}^2 \log p(\mathbf{x}_t) 
            \\ \label{eq:Laplacian-is-Covariance}
            &= \frac{\mu^2_t \, \Tr \Cov [\mathbf{x}_1 | \mathbf{x}_t] }{\sigma^4_t} - \frac{ d}{\sigma^2_t} 
        \end{align}
        where \(d\) is the dimension.
    \end{lemma}

\section{The \(t\downarrow0\) limit}\label{sec:the-t-downarrow-zero-limit}

In this section we discuss how the \logweighingfunction{} behaves at the \(t\downarrow0\) limit. In particular, we show the following.

\begin{proposition}
    Consider the \((1+1)\)-parameter family of WSDEs (Eq.~\ref{eq:1-knob-family-in-short-MAIN}) sampling from the product-of-intermediate-experts trajectory (Eq.~\ref{eq:PoIE-trajectory-MAIN}) of \(K\) \lineardiffusionmodel{}s (linearity chosen for simplicity and ubiquity in practice) (Eq.~\ref{eq:nonlinear-diffusion-model-MAIN} with Eq.~\ref{eq:linear-diffusion-model-short-MAIN}). 

    Then, for a reasonable \lineardiffusionmodel{} noising schedule \((\mu_t,\sigma_t)\) (see Eq.~\ref{eq:linear-interpolant}), the \logweighingfunction{} converges point-wise to a constant function iff \(\knob=1/B\). If \(\knob\neq1/B\), then 
    the \logweighingfunction{} approaches a quadratic function with unbounded curvature as \(t\downarrow0\).
\end{proposition}

Recall the \logweighingfunction{} for \lineardiffusionmodel{}s (Eq.~\ref{eq:1-knob-family-logweighingfunction-in-short-MAIN} with Eq.~\ref{eq:linear-diffusion-model-short-MAIN})
\begin{equation}
    \nonumber
    \begin{aligned}
        h^{\red{\knob}}_t(\mathbf{x}) 
            &=\noiseparam^2_t \Bigg(\tfrac12(\red{\knob}-\tfrac1B) \sqnorm{\nabla_{\mathbf{x}} \log \pPoIE_t(\mathbf{x})}
            &(\text{squared-score term)}
            \\&\quad {} 
            -\tfrac{1}{4\sumbetashort}\sum_{j,k}  \beta_j \beta_k \inlinesqnorm{\nabla_{\mathbf{x}} \log p^{j}_t(\mathbf{x}) - \nabla_{\mathbf{x}} \log \pkt(\mathbf{x})} 
            &(\text{drift-disagreement term)}
            \\&\quad {}  
            + \tfrac12\left(-1+\red{\knob}\right) \sum_k \beta_k \nabla_{\mathbf{x}} \cdot \nabla_{\mathbf{x}} \log \pkt(\mathbf{x})\Bigg)
            &(\text{score divergence term)}
            \\&\quad {}  
            + \driftparam_t(B-1)d 
            &(\text{constant w.r.t }\mathbf{x}_t)
        \\\big(\text{where }&B\doteq\textstyle\sum_k\beta_k\big)
    \end{aligned}
\end{equation}

Recall that all \(K\) marginals \(\pkt\) approach the standard Gaussian \(\mathcal N(0,I)\) as \(t\downarrow0\) (see definition of \lineardiffusionmodel{} Eq.~\ref{eq:linear-interpolant}). Therefore, if \(\knob=1/B\) then the terms inside the large brackets obviously converge to an \(\mathbf{x}\)-constant function as \(t\downarrow0\), and if \(\knob\neq1/B\), it does not \footnote{In fact if \(\knob \neq 1/B\), then due to \(\lim_{t\downarrow0}\noiseparam_t=\infty\) in the squared-score term, the \logweighingfunction{} becomes an infinitely sharp quadratic function as \(t\downarrow0\).}. 

What worries us is that two (hyperparameter) coefficients (from the definition of a linear diffusion model Def.~\ref{def:linear-diffusion-model}) diverges to infinity:

\begin{remark}\label{lemma:two-coefficients-diverge-to-infinity}
The two \lineardiffusionmodel{} hyperparameters' integrals diverge near zero. For every \(\epsilon\in(0,1)\), we have
\begin{equation}
    \int_0^\epsilon \driftparam_t \, dt = -\infty
    \quad\text{and}\quad
    \int_0^\epsilon \noiseparam^2_t \, dt = \infty.
\end{equation}
In particular, neither coefficient is bounded on any interval \((0,\epsilon)\), i.e.\
\(\liminf_{t\downarrow0}\driftparam_t=-\infty\) and \(\limsup_{t\downarrow0}\noiseparam^2_t=\infty\).
(The limits \(\lim_{t\downarrow0}\driftparam_t\) and \(\lim_{t\downarrow0}\noiseparam^2_t\) themselves need not exist.)
\LineBetweenStatementAndProof
\begin{proof}
    This follows from the boundary condition we impose on \((\mu_t,\sigma_t)\) at time \(t=0\) (Eq.~\ref{eq:noising-schedule-boundary-conditions}).
    
\end{proof}
\end{remark}

Fortunately, this can be resolved by first rewriting the \logweighingfunction{} in terms of Tweedie's identities (Appendix.~\ref{sec:Tweedie-identity}):

\begin{subequations}
    \begin{align}
        h^{\red{\knob}}_t(\mathbf{x}) 
            &=\tfrac{\noiseparam^2_t}{2}(\red{\knob}-\tfrac1B) \sqnorm{\sum_k \beta_k \frac{\mu_t \mathbb E^k[\mathbf{x}_1|\mathbf{x}_t=x]-x}{\sigma^2_t}}
            &(\text{squared-score term)}
            \label{eq:squared-score-term-t-downarrowzero}
            \\&\quad {} 
            -\frac{\noiseparam^2_t}{4\sumbetashort} \frac{\mu^2_t}{\sigma^4_t} \sum_{j,k}  \beta_j \beta_k \sqnorm{\mathbb E^j[\mathbf{x}_1|\mathbf{x}_t=x]-\mathbb E^k[\mathbf{x}_1|\mathbf{x}_t=x]} 
            &(\text{expectation-disagreement term)}
            \label{eq:exp-disagree-term-t-downarrowzero}
            \\&\quad {}  
            + \frac{\noiseparam^2_t}{2} \frac{\mu^2_t}{\sigma^4_t}  \left(-1+\red{\knob}\right) \sum_k \beta_k \Tr \Cov^k [\mathbf{x}_1 | \mathbf{x}_t=x] 
            &(\text{trace of covariances term)}
            \label{eq:trace-covariance-term-t-downarrowzero}
            \\&\quad {}  
            + \driftparam_t(B-1)d 
            - \frac{\noiseparam^2_t}{2} \frac{d}{\sigma^2_t}  \left(-1+\red{\knob}\right) B
            &(\text{one additional constant w.r.t }\mathbf{x}_t).
    \end{align}
\end{subequations}
where the expectations \(\mathbb E^k\) are taken with respect to the joint distribution described by Eq.~\ref{eq:linear-interpolant} of the \(k\)-th model. Recall that constant can effectively be ignored as they do not affect the SMC procedure. The remaining terms can be shown to be well-behaving if we set \(\knob=1/B\) and choose a 'good' noising schedule.

\paragraph{Expectation-disagreement term and the trace of covariances term}
The term \(\frac{\noiseparam^2_t \mu^2_t}{\sigma^4_t}\) can also be made to converge to a finite value, in particular zero, as \(t\downarrow0\) by choosing a suitable noising schedule. 
(This can be seen more easily by rewriting \(\frac{\noiseparam^2_t \mu^2_t}{\sigma^4_t} = \tfrac{d}{dt}\SNR_t\) where \(\SNR_t\) (referred to as signal-noise-ratio in the literature) is defined as \(\SNR_t\doteq\tfrac{\mu^2_t}{\sigma^2_t}\).)

To finish things off, one can show that the expectation and covariance point-wise converge to constants as \(t\downarrow0\): 
\begin{equation}
    \forall x,\ \lim_{t\downarrow0}\mathbb E^k[\mathbf{x}_1|\mathbf{x}_t=\mathbf{x}]=\int \mathbf{x}_1 \,p^{\smash{(k)}}_{t=1}(\mathbf{x}_1)\,d\mathbf{x}_1 \doteq \mathbb E^k[\mathbf{x}_1]
\end{equation}
and 
\begin{equation}
   \forall x,\ \lim_{t\downarrow0} \Tr \Cov^k[\mathbf{x}_1|\mathbf{x}_t=\mathbf{x}]= \int \inlinesqnorm{\mathbf{x}_1-\mathbb E^k[\mathbf{x}_1]} \ p^{\smash{(k)}}_{t=1}(\mathbf{x}_1)\,d\mathbf{x}_1 \doteq \Tr \Cov^k[\mathbf{x}_1]. 
\end{equation}
This can be shown if the data distributions \(\pktdata\) have finite first and second moments respectively.

Hence the expectation-disagreement term (Eq.~\ref{eq:exp-disagree-term-t-downarrowzero}) and the trace-covariance term (Eq.~\ref{eq:trace-covariance-term-t-downarrowzero}) can be made to converge to constants as \(t\downarrow0\) by choosing a suitable noising schedule.

\paragraph{Conclusion}

Under mild regularity conditions, one can use 'good' noising schedule to ensure a well behaving \logweighingfunction{} (for \(\knob=1/B\)) at time zero. Meanwhile for \(\knob\neq1/B\), this is not possible as the \logweighingfunction{} approaches a quadratic function with unbounded curvature as \(t\downarrow0\), for any noising schedule because of the \(\noiseparam^2_t\) coefficient (see Remark.~\ref{lemma:two-coefficients-diverge-to-infinity}) in Eq.~\ref{eq:squared-score-term-t-downarrowzero}. 

This misbehavior of the \logweighingfunction{} for \(\knob\neq1/B\) may be why
\cite{skreta_feynman-kac_2025} (which only uses \(\knob=1\)) found sampling instability during the early timesteps. They mitigated this by 
starting the Euler--Maruyama integration not
from time zero, but from some small finite time above zero (\(t=0.01\))
    \footnote{In their LJ-13 experiments, the authors mention \cite[Appendix.~F.3 'Training details']{skreta_feynman-kac_2025} that starting integration not from a fully-noised state, but rather from \(t_{start}=0.99\) (which is equivalent to \(t_{start}=0.01\) in our time convention) "help[ed] significantly to reduce the variance of the results over different runs". An analogous detail was implicitly implemented in their drug-design experiments: the diffusion model they use, TargetDiff \cite{guan20233d}, starts from a finite Noise-Signal-Ratio (Eq.~\ref{eq:definition-noise-signal-ratio}) value---a common implementation detail for diffusion models in practice---hence effectively starting generation from some time \(t_{start}>0\) in our convention.}.  
This ad-hoc fix breaks asymptotic correctness of our WSDEs. 

However, other values of \(\knob\) can still be useful in practice as one can use \(\knob=1/B\) during the early phases of generation, then switch to a different \(\knob\) later on.
As an example, one can use \(\knob=1/B\) early in generation (when RC.~\ref{cond:bounded-score-divergence-MAIN} likely holds) and switching to \(\knob=1\) later on (when RC.~\ref{cond:bounded-score-divergence-MAIN} may not hold).

\section{Links to SMC convergence results}\label{sec:links-to-SMC-convergence-results}

Bounds on the \logweighingfunction{} of a WSDE can be used to derive bounds on the convergence rate of the SMC algorithm applied to said WSDE.
For examples of results which follow from an upper bound, see e.g. \cite[Chp.~11.2, 11.3]{chopin2020introduction} and for examples of results which follow from having an upper\(+\)lower bound, see e.g. \cite[Chp.~11.4]{chopin2020introduction} and \cite[Sec.~3.5.2, Condition (G)]{moral2004feynman}. 
However, we anticipate the bound divergence at \(t\uparrow1\) discussed in Sec.~\ref{sec:t-uparrow-one-case-MAIN} to be a technical challenge in applying our bounds (Prop.~\ref{prop:positive-exponents-implies-above-boundedness-MAIN},~\ref{prop:compact-support-implies-boundedness-MAIN},~\ref{prop:one-model-implies-boundedness-MAIN}) for this purpose. We demonstrate a proof sketch below in the compact support setting
    \footnote{The compact data distributions \(\pktdata\) with a linear diffusion model setting (Prop.~\ref{prop:compact-support-implies-boundedness-MAIN}) is useful in this effort because we obtain bounds in all non-terminal times \(t<1\) \textit{without} any regularity conditions. In contrast, we cannot guarantee this for Prop.~\ref{prop:positive-exponents-implies-above-boundedness-MAIN},~\ref{prop:one-model-implies-boundedness-MAIN}. For example, even if a data distribution \(\pktdata\) satisfies the score smoothness condition RC.~\ref{cond:bounded-score-divergence-MAIN} we cannot in general guarantee that the other marginals \(\pk_{t<1}\) will satisfy RC.~\ref{cond:bounded-score-divergence-MAIN}, even if we restrict ourselves to the \lineardiffusionmodel{} (Eq.~\ref{eq:linear-diffusion-model-short-MAIN}) setting.}
, and leave a more formal investigation as further work. 

\paragraph{Proof sketch in the compact support setting}

Our compact support result (Prop.~\ref{prop:compact-support-implies-boundedness-MAIN}) gives an upper and lower bound on the \logweighingfunction{} (Eq.~\ref{eq:1-knob-family-logweighingfunction-in-short-MAIN}) at all non-terminal times \(t<1\): 
\begin{align}\label{eq:bounded-h-lambda-general-form}
    -\infty<m_t&<h^{\knob=1/B}_t(\mathbf{x}_t)<M_t<\infty, \quad\text{ for some }m_t, M_t\in\mathbb R.
\end{align}
To turn this into the form of Condition~(G) in \cite[Sec.~3.5.2]{moral2004feynman} one could define a time discretization scheme \(0<t_{0}<t_1<\dots<t_N<1\) 
    \footnote{We cannot choose the final time step to be \(t_{N}=1\) 
    because the gap between the upper and lower bounds in Prop.~\ref{prop:compact-support-implies-boundedness-MAIN} grows unbounded as \(t\uparrow1\) (as discussed in Sec.~\ref{sec:t-uparrow-one-case-MAIN} and Remark.~\ref{rem:compact-support-result-doesnt-hold-at-time-one}). Choosing \(t_0=0\) also results in some technical difficulties, likely more resolvable than that of \(t_{N}=1\); we briefly discuss it in Appendix.~\ref{sec:the-t-downarrow-zero-limit}.}
and define the interval (log) weighing function
\begin{equation}
    G_{n}(\mathbf{x}_{t_{n-1}:t_{n}})\doteq\textstyle\int^{t_{n}}_{t'=t_{n-1}} h^{\knob=1/B}_{t'}(\mathbf{x}_{t'})\, d{t'}= \log w_{t_n} - \log w_{t_{n-1}}
\end{equation}
and observe that 
    \footnote{This also immediately implies a bound on the effective-sample-size as well.}
\begin{equation}\label{eq:bounded-G-general-form}
    -\infty<m^{\text{c}}_{t_{n-1}:t_{n}}<G_{n}(\mathbf{x}_{t_{n-1}:t_{n}})<M^{\text{c}}_{t_{n-1}:t_{n}}<\infty
\end{equation}
for some finite \(m^{\text{c}}_{t_{n-1}:t_{n}}\) and \(M^{\text{c}}_{t_{n-1}:t_{n}}\) 
    \footnote{Recall that in our paper we confined our scope to WSDEs with no directional weight (Eq.~\ref{eq:wSODE-with-zero-directional-weight-MAIN} with \(\mathbf{j}_t(\mathbf{x})=0\))---this choice allowed us to turn bounds on the integrand "\(h\)" to an almost sure bound on the integral "\(G\)" as long as "\(h\)" is integrable---hence linking our bounds with the wider SMC literature.}
    \footnote{Strictly speaking, Condition~(G) in \cite[Sec.~3.5.2]{moral2004feynman} regards importance weight functions \(G_n(\mathbf{x}_n)\) which are functions of \(\mathbf{x}_n\), whereas our \(G_n\) is a function of the path \(\mathbf{x}_{n-1}:\mathbf{x}_n\). We leave this detail as further work.},
as long as \(t_n<1\).
This bounded log-weight accumulation function \(G_{n}\) allows us to apply any results which follow from Condition~(G) in \cite{moral2004feynman} to our compact support setting for any time interval strictly contained between zero and one: \(0<t_{\text{start}}<t_{\text{final}}<1\).

\paragraph{Limitations of SMC convergence results}
Such SMC convergence results must be interpreted with caution. For example, for the SMC convergence results quoted above \cite[Chp.~11.2, 11.3, 11.4]{chopin2020introduction}, \cite[Sec.3.5.2, Condition (G)]{moral2004feynman} an upper bound or an upper\(+\)lower bound is not a necessary condition, but rather a sufficient and convenient one \cite{chopin2020introduction,moral2004feynman}. Also, such convergence results cannot be expected to guarantee superiority of sampling efficiency of WSDEs in practice---such statements will need to be determined empirically (Sec.~\ref{sec:experiments}).

\section{Prop.~\ref{prop:positive-exponents-implies-above-boundedness-MAIN}'s connection to Hölder's inequality}\label{sec:noising-process-intuition-MAIN}

This section provides an alternative (informal) proof of Prop.~\ref{prop:positive-exponents-implies-above-boundedness-MAIN} and Corr.~\ref{cor:reverse-holder-result}, and in particular why it holds without regularity conditions 
when \(\sumbetaksmall=1\). We show that in this setting Prop.~\ref{prop:positive-exponents-implies-above-boundedness-MAIN} and Corr.~\ref{cor:reverse-holder-result} become an application of Hölder's inequality and reverse \cite[Thm.~2.2]{cheung2001generalizations} Hölder's inequality, respectively.

\subsection{Noising process analogue of the \((1+1)\)-parameter family}

It is well known \cite{sohl-dickstein_deep_2015,song_score-based_2021} that a diffusion model 
\begin{equation}\tag{Eq.~\ref{eq:nonlinear-diffusion-model-MAIN}}
    d\mathbf{x}_t = [-\mathbf{F}_t(\mathbf{x}_t) + \tfrac{\xi_t^2}{2} \nabla_{\mathbf{x}} \log p_t(\mathbf{x}_t)]\, dt + \Langevin_{\pkt,g_t}
\end{equation}
shares its trajectory of marginals \(\{p_t\}_{t\in[0,1]}\) with a backward-in-time (so-called) noising process 
    \footnote{In fact, it was initially conceived \cite{sohl-dickstein_deep_2015} as the time-reversal of this noising process. In many recent expositions (e.g. \cite{holderrieth2025introduction}) of linear diffusion models Eq.~\ref{eq:linear-diffusion-model} (and the equivalent \cite{gao2025diffusionsamecoin} flow-matching models) this noising process is often omitted or considered to be an incidental peripheral detail. In this Appendix section, the regularity offered by the existence of this \textit{shared} (among \(K\) diffusion models) noising process can be seen as playing a crucial role in allowing
    (through the definitions Eq.~\ref{eq:discrete-time-target}, Eq.~\ref{eq:discrete-time-proposal}) 
    a simple application of Hölder's inequality to result in a-priori lower or upper bounds Propositions.~\ref{prop:positive-exponents-implies-above-boundedness-DISCRETE},~\ref{cor:reverse-holder-result-DISCRETE}.}
\begin{equation}\label{eq:nonlinear-noising-process-MAIN}
    d\mathbf{x}_\tau = \mathbf{F}_\tau(\mathbf{x}_\tau) \, d\tau + \noiseparam_\tau\, d\mathbf{W}_\tau, \quad\quad \tau=1-t\in[0,1].
\end{equation}

Likewise, the \((1+1)\)-parameter family we have examined (Eq.~\ref{eq:1-knob-family-in-short-MAIN})
\begin{align}
	&\begin{cases}%
		d\mathbf{x}_t &= \left[- \mathbf{F}_t(\mathbf{x}_t)+ \red{\knob}\tfrac{\noiseparam^2_t}{2}\nabla_{\mathbf{x}} \log \pPoIE_t(\mathbf{x}_t)\right] \, dt + \Langevin_{\pPoIE_t,g_t} \\
		d \log w_t &= h^{\red{\knob}}_t(\mathbf{x}_t) \, dt \\
	\end{cases}
\end{align}
shares its trajectory of marginals \(\{\pPoIE_t\}_{t\in[0,1]}\) (Eq.~\ref{eq:PoIE-trajectory-MAIN}) with a backward-in-time noising(-like)
    \footnote{It differs from a conventional definition of a noising process because it is a \textit{weighted} SDE.}
process with a \logweighingfunction{} with a minus sign
\begin{align}\label{eq:analogous-noising-process-MAIN}
	&\begin{cases}
		d\mathbf{x}_\tau
        &=  \mathbf{F}_\tau(\mathbf{x}_\tau)  \, d\tau  + \sqrt{\red{\knob}} \,\noiseparam_\tau\, d\mathbf{W}_\tau \\
		d \log w_\tau &= -h^{\red{\knob}}_\tau(\mathbf{x}_\tau) \, d\tau. \\
	\end{cases}
\end{align}

Due to time reversal the sign (in front of \(h^\knob_\tau\) in Eq.~\ref{eq:analogous-noising-process-MAIN}) flips.
Hence we can say Prop.~\ref{prop:positive-exponents-implies-above-boundedness-MAIN} that when the exponents \(\beta_k\) are positive, then \(\knob=1/B\) is the largest diffusion coefficient which guarantees (not an upper bounded, but) a \textit{lower} bounded \logweighingfunction{}.

By examining the definition of the product-of-intermediate-experts trajectory \(\{\pPoIE_t\}_{t\in[0,1]}\) (Eq.~\ref{eq:PoIE-trajectory-MAIN}), we can draw a parallel to the following discrete time analogue.

\subsection{Noising process toy example in discrete time setting}\label{sec:discrete-time-result-MAIN}
Consider \(K\) distributions \(\pk\). Let 
\begin{align}\label{eq:discrete-time-target}
     p^{target}(\mathbf{x}) &= \frac{1}{Z_{target}} \prod^K_{k=1} \left(\pk * \mathcal N(0, I)\right)^{\beta_k}
     \\
     \label{eq:discrete-time-proposal}
     p^{proposal}_\knob(\mathbf{x}) &= \frac{1}{Z_{proposal}} \bigg(\prod^K_{k=1} (\pk)^{\beta_k}\bigg)*\mathcal N(0, \knob\cdot I), \quad \knob>0.
\end{align}
In other words, the \(target\) distribution is created by first convolving then taking the product, whereas the \(proposal\) distribution is created by first taking the product then convolving. 

Then define the importance weight function
\begin{equation}
    w^\knob(\mathbf{x}) \doteq \log \frac{p^{target}(\mathbf{x})}{p^{proposal}_{\knob}(\mathbf{x})}.
\end{equation}
\(w^\knob(\mathbf{x})\) is analogous to \(-h^\knob(\mathbf{x})\) in Eq.~\ref{eq:analogous-noising-process-MAIN}.

We can state the following
\begin{proposition}[Positive exponents \(\implies\) lower-boundedness.]\label{prop:positive-exponents-implies-above-boundedness-DISCRETE}
    When all exponents \(\beta_k\) are \textbf{normalized} (\(\sumbetaksmall=1\)) and \textbf{positive} (\(\beta_k>0\)), then \(\knob\leq1/B\) produces a \(w^{\knob}(\mathbf{x})\) that is lower-bounded
    \begin{equation}
        -\infty<\inf_{\mathbf{x}} w^\knob(\mathbf{x}).
    \end{equation}    
    
\end{proposition}

\begin{proposition}\label{cor:reverse-holder-result-DISCRETE}
    When all exponents \(\beta_k\) are \textbf{normalized} (\(\sumbetaksmall=1\)) and only one is positive with the rest negative (\(\beta_1>0>\beta_k,\forall k>1\)), then \(\knob\geq1/B\) produces a \(w^{\knob}(\mathbf{x})\) that is upper-bounded
        \begin{equation}
            \sup_{\mathbf{x}} w^\knob(\mathbf{x}) <\infty.
        \end{equation}   
\end{proposition}

These two results are discrete time analogues of Prop.~\ref{prop:positive-exponents-implies-above-boundedness-MAIN} and Corr.~\ref{cor:reverse-holder-result}, respectively \footnote{Where the presence of upper and lower boundedness are now swapped due to \(w^\knob(\mathbf{x})\) being analogous to the sign flipped \(-h^\knob(\mathbf{x})\) rather than \(h^\knob(\mathbf{x})\).}.

Both results are due to a simple application of Hölder's inequality (respectively, the reverse Hölder inequality). We give the proof of the first.

\begin{proof}[Proof of Prop.~\ref{prop:positive-exponents-implies-above-boundedness-DISCRETE}]
    Write \(g_s\) for the density of \(\mathcal N(0,sI)\) and \(q\doteq\prod_k(\pk)^{\beta_k}\). Since \(\sumbetaksmall=1\), we may write \(g_1=\prod_kg_1^{\beta_k}\), and the generalised Hölder inequality \(\int\prod_kf_k^{\beta_k}\leq\prod_k\bigl(\int f_k\bigr)^{\beta_k}\) (for \(f_k\geq0\), \(\beta_k>0\), \(\sumbetaksmall=1\)), applied with \(f_k(\mathbf{x})=\pk(\mathbf{x})\,g_1(\mathbf{y}-\mathbf{x})\), gives for every \(\mathbf{y}\)
    \begin{align}\label{eq:DISCRETE-holder-step}
        \bigl(q*\mathcal N(0,I)\bigr)(\mathbf{y})
        &=\int\prod_k\bigl(\pk(\mathbf{x})\,g_1(\mathbf{y}-\mathbf{x})\bigr)^{\beta_k}d\mathbf{x}
        \;\leq\;\prod_k\Bigl(\int\pk(\mathbf{x})\,g_1(\mathbf{y}-\mathbf{x})\,d\mathbf{x}\Bigr)^{\beta_k}\\
        &=Z_{target}\,p^{target}(\mathbf{y}). \nonumber 
    \end{align}
    Since \(\knob\leq1/B=1\), we have \(e^{-\|\mathbf z\|^2/(2\knob)}\leq e^{-\|\mathbf z\|^2/2}\), i.e.\ \(g_\knob\leq\knob^{-d/2}g_1\). Hence \(Z_{proposal}\,p^{proposal}_\knob=q*g_\knob\leq\knob^{-d/2}\,q*g_1\leq\knob^{-d/2}Z_{target}\,p^{target}\), so that \(w^\knob(\mathbf{x})\geq\log(Z_{proposal}/Z_{target})+\tfrac d2\log\knob>-\infty\) for all \(\mathbf{x}\).
\end{proof}

Moreover, we now show (Appendix.~\ref{sec:discrete-time-result}) that Proposition.~\ref{prop:positive-exponents-implies-above-boundedness-DISCRETE} can be extended to the \(\sumbetaksmall\neq1\) setting, given that we accept regularity conditions on the magnitude of the second derivative \(\nabla^2_{\mathbf{x}} \log p^{(k)}(\cdot)\). (This is analogous to how in the continuous time setting, the smoothness conditions~\ref{cond:bounded-score-divergence-MAIN},~\ref{cond:bounded-shared-nonlinear-drift-divergence-MAIN} were required except when in the normalized setting (Remark.~\ref{rem:clean-case-sumbetak-is-one}).)

\subsection{Proof of Proposition.~\ref{prop:positive-exponents-implies-above-boundedness-DISCRETE} in both the normalised (\(\sum_k\beta_k=1\)) and unnormalised exponents (\(\sum_k\beta_k\neq1\)) settings}
\label{sec:discrete-time-result}

In short, one assumes a bound on curvature and uses a Taylor expansion to derive inequalities

The \(\sumbetaksmall=1\) case was proved in Sec.~\ref{sec:discrete-time-result-MAIN} with one
application of Hölder's inequality (Eq.~\ref{eq:DISCRETE-holder-step}) and one pointwise comparison of
Gaussian kernels, with no regularity condition. When \(\sumbetaksmall\neq1\) the same two steps leave
a residual term, and controlling it
requires one half of a two-sided curvature bound on the log-likelihoods, the half depending on which
side of \(1\) the exponent sum \(\sumbetakshort\doteq\sumbetaksmall\) falls. This is exactly
analogous to the fact that Prop.~\ref{prop:positive-exponents-implies-above-boundedness-MAIN} holds
without regularity conditions~\ref{cond:bounded-score-divergence-MAIN},~\ref{cond:bounded-shared-nonlinear-drift-divergence-MAIN}
only when \(\sumbetaksmall=1\), and requires an upper bound on the Laplacian when \(\sumbetaksmall<1\) and a lower bound when \(\sumbetaksmall>1\).

\begin{condition}[One-sided log-likelihood curvature; needed only if \(\sumbetaksmall\neq1\)]\label{cond:DISCRETE-one-sided-curvature}
    Each \(\pk\) satisfies
    \begin{align}
        \text{(i) }\ \ &\nabla^2_{\mathbf{x}} \log\pk \preceq \kappa I \ \text{ for some } \kappa<1,
        &&\text{if } \sumbetakshort<1;\\
        \text{(ii) }\ \ &\nabla^2_{\mathbf{x}} \log\pk \succeq -L I \ \text{ for some } L<\infty,
        &&\text{if } \sumbetakshort>1.
    \end{align}
\end{condition}

Note that if we had defined the \(p^{target}\) to be convolved by \(\mathcal N(0,\epsilon^2\cdot I)\) rather than \(\mathcal N(0, I)\), then our required condition RC.~\ref{cond:DISCRETE-one-sided-curvature} would be 'for some \(\kappa < 1/\epsilon^2\)', and the critical knob \(\knob\leq\knob^*\doteq\epsilon^2(1/B)\).

\begin{proposition}[Quantitative form of Prop.~\ref{prop:positive-exponents-implies-above-boundedness-DISCRETE}]\label{prop:DISCRETE-curvature-implies-above-boundedness}
    Let RC.~\ref{cond:DISCRETE-one-sided-curvature} hold, and let all exponents be positive \(\beta_k>0\). Then for every \(\knob\leq1/\sumbetakshort\),
    \begin{equation}
        \inf_{\mathbf{x}\in\mathbb R^d} w^\knob(\mathbf{x}) \;>\; -\infty,
    \end{equation}
    with
    \begin{equation}
        \sup_{\mathbf{x}} \frac{p^{proposal}_\knob(\mathbf{x})}{p^{target}(\mathbf{x})}
        \;\leq\; \frac{Z_{target}}{Z_{proposal}}\cdot
        \begin{cases}
            \knob^{-d/2}, & \sumbetakshort=1,\\[6pt]
            \dfrac{1}{(\sqrt\knob\,\sumbetakshort)^{d}\,(1-\kappa)^{(1-\sumbetakshort)d/2}}, & \sumbetakshort<1,\\[10pt]
            \dfrac{(1+L)^{(\sumbetakshort-1)d/2}}{\knob^{d/2}}, & \sumbetakshort>1.
        \end{cases}
    \end{equation}
\end{proposition}

The \(K\) models play no role beyond the single Hölder step: the statement is really one about a
single density raised to a single exponent, which is the following lemma.

\begin{lemma}[Single-distribution bound]\label{lem:DISCRETE-single-distribution}
    Let \(q\) be positive with \(\log q\in C^2(\mathbb R^d)\), let \(b>0\), and let \(\knob\leq1/b\). Then
    \begin{equation}\label{eq:DISCRETE-single-distribution-ratio}
        \mathcal R(\mathbf{y})\doteq\frac{\bigl(q^{b}*\mathcal N(0,\knob I)\bigr)(\mathbf{y})}{\bigl(q*\mathcal N(0,I)\bigr)^{b}(\mathbf{y})}
    \end{equation}
    satisfies
    \begin{equation}
        \sup_{\mathbf{y}\in\mathbb R^d}\mathcal R(\mathbf{y})\;\leq\;
        \begin{cases}
            \knob^{-d/2}, & b=1, \ \text{ with no further assumption},\\[6pt]
            \dfrac{1}{(\sqrt\knob\,b)^{d}\,(1-\kappa)^{(1-b)d/2}},
            & b<1 \ \text{ and } \ \nabla^2_{\mathbf{x}} \log q\preceq\kappa I \ \text{ with } \kappa<1,\\[10pt]
            \dfrac{(1+L)^{(b-1)d/2}}{\knob^{d/2}},
            & b>1, \ \ \nabla^2_{\mathbf{x}} \log q\succeq-L I, \ \text{ and } \ \int q<\infty,
        \end{cases}
    \end{equation}
    and each right-hand side is finite. No integrability assumption on \(q\) is needed in the first
    two cases.
\end{lemma}

\begin{proof}
    \emph{Step 1: the noise level enters only through the kernel.} \(\knob\leq1/b\) is exactly
    \(\tfrac{1}{2\knob}\geq\tfrac{b}{2}\), so \(g_{\knob}(z)\leq C\,g_1(z)^{b}\) for all \(z\), with
    \(C=(2\pi)^{b d/2}(2\pi\knob)^{-d/2}\). This is the only place \(\knob\leq1/b\) is used.

    \emph{Step 2: the peak cancels.} Put \(\Phi_{\mathbf{y}}(\mathbf{x})\doteq\log q(\mathbf{x})-\tfrac12\|\mathbf{y}-\mathbf{x}\|^2\), so that
    \(q(\mathbf{x})g_1(\mathbf{y}-\mathbf{x})=(2\pi)^{-d/2}e^{\Phi_{\mathbf{y}}(\mathbf{x})}\) and \(\nabla^2\Phi_{\mathbf{y}}=\nabla^2\log q-I\).
    Write \(m_{\mathbf{y}}\doteq\sup_{\mathbf{x}}\Phi_{\mathbf{y}}(\mathbf{x})\) and \(w_{\mathbf{y}}\doteq\int e^{\Phi_{\mathbf{y}}(\mathbf{x})-m_{\mathbf{y}}}d\mathbf{x}\). The denominator of
    \eqref{eq:DISCRETE-single-distribution-ratio} is \(\bigl((2\pi)^{-d/2}e^{m_{\mathbf{y}}}w_{\mathbf{y}}\bigr)^{b}\), while
    Step 1 bounds the numerator by \(C(2\pi)^{-b d/2}e^{b m_{\mathbf{y}}}\int e^{b(\Phi_{\mathbf{y}}(\mathbf{x})-m_{\mathbf{y}})}d\mathbf{x}\), so both
    \(e^{m_{\mathbf{y}}}\) and the factors of \(2\pi\) cancel exactly:
    \begin{equation}\label{eq:DISCRETE-width-identity}
        \mathcal R(\mathbf{y})\;\leq\;\frac{C}{w_{\mathbf{y}}^{b}}\int e^{b(\Phi_{\mathbf{y}}(\mathbf{x})-m_{\mathbf{y}})}d\mathbf{x}.
    \end{equation}

    \emph{Step 3: \(b=1\).} The integral is \(w_{\mathbf{y}}\) itself, so the two widths cancel and
    \(\mathcal R(\mathbf{y})\leq C=\knob^{-d/2}\) with no assumption on \(q\) beyond positivity. For \(b\neq1\)
    they do not cancel, and a residual power of \(w_{\mathbf{y}}\) survives; the remaining two cases bound it,
    from above when \(b<1\) and from below when \(b>1\).

    \emph{Step 3\('\): \(b<1\).} Here \(\nabla^2\Phi_{\mathbf{y}}\preceq-(1-\kappa)I\), i.e.\ \(\Phi_{\mathbf{y}}\) is
    strongly concave and attains its maximum at a unique \(\mathbf{x}^*_{\mathbf{y}}\). Concavity applied to
    \(\mathbf{x}^*_{\mathbf{y}}+b(\mathbf{x}-\mathbf{x}^*_{\mathbf{y}})\) gives \(b(\Phi_{\mathbf{y}}(\mathbf{x})-m_{\mathbf{y}})\leq\Phi_{\mathbf{y}}\bigl(\mathbf{x}^*_{\mathbf{y}}+b(\mathbf{x}-\mathbf{x}^*_{\mathbf{y}})\bigr)-m_{\mathbf{y}}\);
    exponentiating and substituting \(u=\mathbf{x}^*_{\mathbf{y}}+b(\mathbf{x}-\mathbf{x}^*_{\mathbf{y}})\) gives
    \(\int e^{b(\Phi_{\mathbf{y}}-m_{\mathbf{y}})}\leq b^{-d}w_{\mathbf{y}}\) (Jacobian \(b^{-d}\)). Strong concavity also gives
    \(\Phi_{\mathbf{y}}(\mathbf{x})\leq m_{\mathbf{y}}-\tfrac{1-\kappa}{2}\|x-\mathbf{x}^*_{\mathbf{y}}\|^2\), hence \(w_{\mathbf{y}}\leq(2\pi/(1-\kappa))^{d/2}\).
    Since \(1-b>0\), substituting both into \eqref{eq:DISCRETE-width-identity} gives
    \(\mathcal R\leq C b^{-d}(2\pi/(1-\kappa))^{(1-b)d/2}=\bigl[(\sqrt\knob\,b)^{d}(1-\kappa)^{(1-b)d/2}\bigr]^{-1}\).

    \emph{Step 3\(''\): \(b>1\).} Here \(\Phi_{\mathbf{y}}\leq m_{\mathbf{y}}\) gives \(\int e^{b(\Phi_{\mathbf{y}}-m_{\mathbf{y}})}\leq w_y\)
    directly. Taylor with integral remainder and \(\nabla^2\Phi_{\mathbf{y}}\succeq-(1+L)I\), anchored at an
    arbitrary \(\mathbf{x}_0\), gives
    \(\Phi_{\mathbf{y}}(\mathbf{x})\geq\Phi_{\mathbf{y}}(\mathbf{x}_0)+\langle\nabla\Phi_{\mathbf{y}}(\mathbf{x}_0),\mathbf{x}-\mathbf{x}_0\rangle-\tfrac{1+L}{2}\|\mathbf{x}-\mathbf{x}_0\|^2\), so
    \(\int e^{\Phi_{\mathbf{y}}}\geq e^{\Phi_{\mathbf{y}}(\mathbf{x}_0)}(2\pi/(1+L))^{d/2}\); taking the supremum over \(\mathbf{x}_0\) gives
    \(w_{\mathbf{y}}\geq(2\pi/(1+L))^{d/2}\). (The same inequality applied to \(\log q\) in place of \(\Phi_{\mathbf{y}}\)
    gives \(\sup_{\mathbf{x}}q\leq\bigl(\int q\bigr)((1+L)/2\pi)^{d/2}<\infty\), which is where \(\int q<\infty\)
    is used, and which makes \(m_y\) finite.) Since \(1-b<0\), \eqref{eq:DISCRETE-width-identity} gives
    \(\mathcal R\leq C(2\pi/(1+L))^{(1-b)d/2}=(1+L)^{(b-1)d/2}\knob^{-d/2}\).
\end{proof}

\begin{proof}[Proof of Prop.~\ref{prop:DISCRETE-curvature-implies-above-boundedness}]
    Write \(\alpha_k\doteq\beta_k/\sumbetakshort\), so \(\sum_k\alpha_k=1\), and set
    \(q\doteq\prod_k(\pk)^{\alpha_k}\), so that \(\log q=\sum_k\alpha_k\log\pk\). The Hölder step
    Eq.~\ref{eq:DISCRETE-holder-step}, with \(\alpha_k\) in place of \(\beta_k\) (valid since \(\alpha_k>0\) and \(\sum_k\alpha_k=1\)), gives
    \(\bigl(q*\mathcal N(0,I)\bigr)(\mathbf{y})\leq\prod_k\bigl(\pk*\mathcal N(0,I)\bigr)(\mathbf{y})^{\alpha_k}\) for every \(\mathbf{y}\), and the same inequality with \(f_k=\pk\) gives \(\int q\leq\prod_k\bigl(\int\pk\bigr)^{\alpha_k}=1\). Raising the
    first bound to the power \(\sumbetakshort>0\), and using \(q^{\sumbetakshort}=\prod_k(\pk)^{\beta_k}\)
    exactly,
    \begin{equation}
        \bigl(q*\mathcal N(0,I)\bigr)^{\sumbetakshort}\leq Z_{target}\,p^{target},
        \qquad
        q^{\sumbetakshort}*\mathcal N(0,\knob I)= Z_{proposal}\,p^{proposal}_\knob,
    \end{equation}
    so that \(p^{proposal}_\knob/p^{target}\leq(Z_{target}/Z_{proposal})\,\mathcal R\) with \(\mathcal R\)
    as in \eqref{eq:DISCRETE-single-distribution-ratio} for \(b=\sumbetakshort\). When
    \(\sumbetakshort=1\) the first case of Lemma~\ref{lem:DISCRETE-single-distribution} applies with no
    further work, recovering the proof in Sec.~\ref{sec:discrete-time-result-MAIN}. Otherwise, both curvature constraints are convex in \(\log q\), so \(\log q\) inherits the
    same \(\kappa\) (resp.\ \(-L\)) from RC.~\ref{cond:DISCRETE-one-sided-curvature}, and
    \(\int q\leq1\) supplies the integrability needed in the third case.
\end{proof}

\section{Further experimental details}\label{sec:further-experimental-details}
\paragraph{Calculation of the score divergence terms in the \logweighingfunction{}}
Our \((1+1)\)-parameter family (Eq.~\ref{eq:1-knob-family-in-short-MAIN}) in general requires the calculation of divergence of score functions (Eq.~\ref{eq:score-divergence-term-MAIN}). The naive approach is to do so using auto-differentiation (assuming the score function estimator is differentiable, as is with modern neural network implementations). The more computationally efficient approach is to use the method presented in \cite{he2026rne}, i.e to use the Ito integration identity Eq.~\ref{eq:divergence-calculation-trick-backward-forward-ito-integral}. For ALDP and LJ-13, we use the latter approach. For dual-target molecular generation we use the former approach. For Mixture-of-Gaussians experiments, we use analytical score functions and analytical score divergence evaluations.

\subsection{Alanine Dipeptide (ALDP)}
For domain background on the ALDP task see \cite{skreta_feynman-kac_2025}.

Our \((1+1)\)-parameter family of WSDEs Eq.~\ref{eq:1-knob-family-in-short-MAIN} is a strict subset of the family of computationally tractable WSDEs sampling from the product-of-intermediate-experts (Eq.~\ref{eq:PoIE-trajectory-MAIN}) in \cite{he2026rne}. Hence (by using the translation formula in Eq.~\ref{eq:translation-from-He26-to-our-notation}) we use the open-sourced software implementation and pre-trained neural network model (score function estimator) provided by \cite{he2026rne}. For details on both neural network training and sampling hyperparameters see \cite[Appendix.~I.1]{he2026rne}. 

We also made two modifications to \cite{he2026rne}'s implementation. First, we use systematic SMC resampling instead of multinomial SMC resampling. Multinomial SMC resampling is provably less efficient than other resampling methods such as stratified and residual resampling \cite{douc2005comparison}, and systematic resampling is largely the most commonly used resampling method in practice because of its ease of implementation and because it "outperforms other resampling schemes in most scenarios" \cite[p.~13]{doucet2009tutorial}. Second, we reduce the batch size from 500 to 312 and the repeat count (per hyperparameter setting in the grid) from 50 to 8 to reduce computational workload.

\subsection{Lennard-Jones potential (LJ-13)}
For domain background on the LJ-13 task see \cite{skreta_feynman-kac_2025}.

We use the pre-trained neural network model (score function estimator) open-source released by \cite{skreta_feynman-kac_2025}. See \cite[Appendix.~F.3 \textit{'Training details'}]{skreta_feynman-kac_2025} for details on training and model definition. We also use the reference samples (samples from the true LJ-13 distribution sampled using MCMC) open-source released by \cite[Appendix.~F.3, \textit{'Sampling reference distributions'}]{skreta_feynman-kac_2025}.

We implement a WSDE integrator analogous to that open-source released by \cite{he2026rne} for the ALDP task.

We use systematic resampling, a batch size of 500, and repeat count (per hyperparameter setting in the grid) of 16.

\subsection{Mixture-of-Gaussians experiment}

We implement an analytical score function and score divergence function for all experiments.

\subsection{Dual-target molecular generation}
We use the pre-trained neural network model (score function estimator) open-source released by \cite{skreta_feynman-kac_2025}.

We use the \texttt{functorch} library \cite{functorch2021} to calculate the divergence of the pre-trained neural network using a Hutchinson's estimator. We leave the implementation of a WSDE integrator analogous to that open-source released by \cite{he2026rne} in this task as further work.

To reduce computational requirements of divergence calculation, we calculate the divergence only every 20 (Euler--Maruyama integration) time-steps. Furthermore, we exponentially smooth the divergence estimate across timesteps to reduce detrimental effects of the high variance estimate of divergence on sample efficiency.

All other sampling settings are identical to \cite[Appendix.~F.6]{skreta_feynman-kac_2025}. In particular, we generate 32 molecules for each of five different sizes (atom count) \(\{15, 19, 23, 27, 35\}\). 

\section{Proof of Proposition~\ref{prop:CONJECTURE-Langevin-Infinity-MAIN}}\label{sec:proof-of-Langevin-Infinity}

We prove Prop.~\ref{prop:CONJECTURE-Langevin-Infinity-MAIN} in the form of Thm.~\ref{thm:finite-action} and Cor.~\ref{cor:finite-action-consequences} (App.~\ref{sec:langevin-finite-action}): as the Langevin strength grows, the weights converge to a constant and the unweighted particles converge to the target, under a single finite-action condition. For \(\mathbf{f}_t\equiv0\) and constant \(g\), the weight bound of Thm.~\ref{thm:finite-action} reduces to \cite[Thm.~2]{guo2026complexity}. The result holds for \emph{any} WSDE with zero directional weight --- not just for the PoIE trajectory, not just for diffusion models, and for any drift \(\mathbf{f}_t\) (hence any knob \(\knob\)).

\subsection{Setting}\label{sec:langevin-setting}

Fix a horizon \([\tinit,T]\). Consider the WSDE of Eq.~\ref{eq:wSODE-with-zero-directional-weight-MAIN} (zero directional weight),
\begin{align}\label{eq:proof-wsde}
    \begin{cases}
    d\mathbf{x}_t &= \mathbf{f}_t(\mathbf{x}_t)\,dt + \Langevin_{(\rho_t,g_t)}
           \;=\; \mathbf{f}_t(\mathbf{x}_t)\,dt + \tfrac{g_t^2}{2}\nabla_{\mathbf{x}}\log\rho_t(\mathbf{x}_t)\,dt + g_t\,d\mathbf{W}_t\\
    d\log w_t &= h_t(\mathbf{x}_t)\,dt\\
    \mathbf{x}_{\tinit} &\sim \bar\rho_{\tinit},\qquad w_{\tinit}=1,
    \end{cases}
\end{align}
where, as in the main text, \(\{\rho_t\}\) is the (in general unnormalised) mass trajectory, tied to \(\mathbf{f}_t\) and \(h_t\) by the Fokker--Planck PDE (Prop.~\ref{prop:Fokker--Planck-for-WSDEs} with \(\mathbf{j}_t=0\)),
\begin{equation}\label{eq:proof-fp-unnormalised}
    \partial_t\rho_t = -\nabla_{\mathbf{x}}\cdot(\rho_t \mathbf{f}_t) + h_t\,\rho_t ,
\end{equation}
and
\begin{equation}\label{eq:normaliser-and-normalised-target}
    Z_t\doteq\int\rho_t(\mathbf{x})\,d\mathbf{x},\qquad \bar\rho_t\doteq\rho_t/Z_t
\end{equation}
are its normaliser and the normalised target. Eq.~\ref{eq:proof-fp-unnormalised} is linear in \(\rho_t\), so \(\rho_t\) matters only up to a constant factor: the mass carried by the weighted particles of Eq.~\ref{eq:proof-wsde} starts at \(\bar\rho_{\tinit}\) and equals \(\rho_t/Z_{\tinit}\) at time \(t\). For the PoIE trajectory, \(\rho_t=\prod_k(\pkt)^{\beta_k}\), \(Z_t\) is the normalising factor of Eq.~\ref{eq:PoIE-trajectory-MAIN}, \(\bar\rho_t=\pPoIE_t\), and \(h_t=h^{\knob}_t\) is the \logweighingfunction{} of Eq.~\ref{eq:1-knob-family-logweighingfunction-in-short-MAIN}. Since \(\nabla_{\mathbf{x}}\log\rho_t=\nabla_{\mathbf{x}}\log\bar\rho_t\), normalisation does not affect the Langevin term; we write \(\mathbf{s}_t\doteq\nabla_{\mathbf{x}}\log\rho_t\).

\paragraph{Normalised \logweighingfunction{} and weights.} By Eq.~\ref{eq:rate-of-change-of-normalization-constant-for-general-wsdes}, \(\partial_t\log Z_t=\int h_t\,\bar\rho_t\,d\mathbf{x}\). Subtracting this \(\mathbf{x}\)-independent term gives the \emph{normalised} \logweighingfunction{} and weight
\begin{equation}\label{eq:normalised-knob-logweight}
    \bar h_t \;\doteq\; h_t-\partial_t\log Z_t,
    \qquad
    \bar w_t\;\doteq\;\exp\Big(\int_{\tinit}^t\bar h_s(\mathbf{x}_s)\,ds\Big)\;=\;\frac{Z_{\tinit}}{Z_t}\,w_t .
\end{equation}
Dividing Eq.~\ref{eq:proof-fp-unnormalised} by \(Z_t\), the normalised target solves
\begin{equation}\label{eq:proof-fp}
    \partial_t\bar\rho_t = -\nabla_{\mathbf{x}}\cdot(\bar\rho_t \mathbf{f}_t) + \bar h_t\,\bar\rho_t ,
\end{equation}
and \(\bar h_t\) is \emph{mean-zero under the target},
\begin{equation}\label{eq:proof-mean-zero}
    \int \bar h_t(\mathbf{x})\,\bar\rho_t(\mathbf{x})\,d\mathbf{x} = 0 \qquad\text{for every }t.
\end{equation}
Because \(\bar w_t\) and \(w_t\) differ by the deterministic factor \(Z_{\tinit}/Z_t\), self-normalised importance weights, and hence the ESS, are the same for both. The proofs below are stated for \(\bar h_t\) and \(\bar w_t\); Cor.~\ref{cor:finite-action-consequences}(i) translates the result back to \(w_t\).

\paragraph{Unweighted particles.} Throughout, \(q_t\) denotes the law of the \emph{unweighted} particle \(\mathbf{x}_t\) --- the ``proposal'' marginal, which is \textbf{not} \(\bar\rho_t\). By Prop.~\ref{prop:Fokker--Planck-for-WSDEs} with \(\bar h_t\) in place of \(h_t\) (whose mass trajectory is \(\bar\rho_t\), by Eq.~\ref{eq:proof-fp}), \(\int\varphi\,\bar\rho_t=\mathbb E[\bar w_t\varphi(\mathbf{x}_t)]\) (equivalently \(\int\varphi\,\rho_t=Z_{\tinit}\,\mathbb E[w_t\varphi(\mathbf{x}_t)]\)) for test functions \(\varphi\); taking \(\varphi\equiv1\) gives \(\mathbb E[\bar w_t]=1\).

\subsection{Main result: the weights become constant under a finite-action condition}\label{sec:langevin-finite-action}

We show that, as the Langevin strength grows, the weights converge to a constant, in \(L^1\) and in probability, and the unweighted particles converge to the target, under a single condition: the transport that the drift \(\mathbf{f}_t\) leaves undone must have finite cost. For \(\mathbf{f}_t\equiv0\) and constant \(g\), the weight bound Eq.~\ref{eq:finite-action-weights} reduces exactly to \cite[Thm.~2]{guo2026complexity}, and the proof follows theirs \cite[App.~C.2]{guo2026complexity}. Throughout, \(g_t>0\) on \([\tinit,T]\).

\begin{condition}[Finite residual action]\label{cond:finite-action}
    For every \(t\in[\tinit,T]\) there is a vector field \(\mathbf{u}_t\) with
    \begin{equation}\label{eq:residual-transport}
        \nabla_{\mathbf{x}}\cdot(\bar\rho_t\mathbf{u}_t)\;=\;-\bar h_t\,\bar\rho_t \qquad\text{(weakly)},
    \end{equation}
    such that
    \(U_t\doteq\int_{\tinit}^{t}\lVert\mathbf{u}_s\rVert^2_{L^2(\bar\rho_s)}\,g_s^{-2}\,ds<\infty\) for every \(t\in[\tinit,T]\), where \(\lVert\mathbf{u}\rVert^2_{L^2(\bar\rho)}\doteq\int\bar\rho\,\inlinesqnorm{\mathbf{u}}\).
\end{condition}

The normalised \(\bar h_t\) appears in Eq.~\ref{eq:residual-transport} because the left-hand side integrates to zero, so a solution can only exist if the right-hand side does too (Eq.~\ref{eq:proof-mean-zero}). By Eq.~\ref{eq:proof-fp}, Eq.~\ref{eq:residual-transport} says exactly that \(\mathbf{v}_t\doteq\mathbf{f}_t+\mathbf{u}_t\) solves the continuity equation \(\partial_t\bar\rho_t+\nabla\cdot(\bar\rho_t\mathbf{v}_t)=0\), i.e.\ \(\mathbf{v}_t\) is a vector field that \emph{generates} the curve \(\{\bar\rho_t\}\) in the sense of \cite[App.~A.2]{guo2026complexity}. So \(\mathbf{u}_t\) is the part of the transport that the drift \(\mathbf{f}_t\) does not supply, and that the weights must otherwise make up for; if \(\bar h_t\equiv0\), one can take \(\mathbf{u}_t\equiv0\). For \(\mathbf{f}_t\equiv0\), the smallest possible \(\lVert\mathbf{u}_t\rVert_{L^2(\bar\rho_t)}\) is the \(W_2\) metric derivative \(|\dot{\bar\rho}|_t\) \cite[Lem.~4]{guo2026complexity}, so for this smallest choice \(\int\lVert\mathbf{u}_t\rVert^2_{L^2(\bar\rho_t)}dt\) is the action \(\mathcal A\) of \cite[Assump.~1]{guo2026complexity}; any other choice gives a value at least \(\mathcal A\). (Lem.~4 of \cite{guo2026complexity} requires the curve \(\{\bar\rho_t\}\) to be absolutely continuous in \(W_2\), with finite second moments, as in their Assump.~1.)

\begin{condition}[Technical regularity for path measures]\label{cond:technical-path}
    \(t\mapsto g_t\) is continuous and positive on \([\tinit,T]\); \(\rho_t>0\) and \((t,\mathbf{x})\mapsto\rho_t(\mathbf{x})\) is \(C^{1,2}\); \((t,\mathbf{x})\mapsto\mathbf{f}_t(\mathbf{x})\) is continuous and continuously differentiable in \(\mathbf{x}\), so that \(\nabla\cdot\mathbf{f}_t\) (in Eq.~\ref{eq:proof-fp} and in the proof of Lemma~\ref{lem:path-ratio}) exists and is continuous; the SDEs Eq.~\ref{eq:proof-wsde}, Eq.~\ref{eq:backward-wsde} and Eq.~\ref{eq:reference-sde} below have unique solutions, whose marginals are the unique solutions of the corresponding Fokker--Planck equations; and Girsanov's theorem applies to each pair of them (e.g.\ Novikov's condition holds), for every value of the Langevin strength considered, as assumed throughout \cite{guo2026complexity} (their Rem.~3).
\end{condition}

\paragraph{Three path measures.}
Fix \(t\in(\tinit,T]\). All path measures below are laws of paths \((\mathbf{x}_r)_{r\in[\tinit,t]}\).
\begin{itemize}
    \item \(\mathbb P^{\rightarrow}\) is the law of the (unweighted) particle path of Eq.~\ref{eq:proof-wsde}. Its marginal at time \(r\) is \(q_r\).
    \item \(\mathbb P^{\leftarrow}\) is the law of the backward SDE, in the sense of \cite[Def.~1]{guo2026complexity},
    \begin{equation}\label{eq:backward-wsde}
        d\mathbf{x}_r=\big[\mathbf{f}_r-\tfrac{g_r^2}{2}\mathbf{s}_r\big](\mathbf{x}_r)\,dr+g_r\,d\mathbf{W}^{\leftarrow}_r,\qquad \mathbf{x}_t\sim\bar\rho_t .
    \end{equation}
    Equivalently, the reversed path \(\mathbf{y}_\sigma\doteq\mathbf{x}_{t-\sigma}\) solves the ordinary (forward) SDE \(d\mathbf{y}_\sigma=\big[-\mathbf{f}_{t-\sigma}+\tfrac{g_{t-\sigma}^2}{2}\mathbf{s}_{t-\sigma}\big](\mathbf{y}_\sigma)\,d\sigma+g_{t-\sigma}\,d\mathbf{B}_\sigma\), \(\mathbf{y}_0\sim\bar\rho_t\): it is Eq.~\ref{eq:proof-wsde} run backwards from \(\bar\rho_t\), with the drift \(\mathbf{f}\) reversed and the same Langevin gauge term.
    \item \(\mathbb P\) is the law of the reference SDE
    \begin{equation}\label{eq:reference-sde}
        d\mathbf{x}_r=\big[\mathbf{v}_r+\tfrac{g_r^2}{2}\mathbf{s}_r\big](\mathbf{x}_r)\,dr+g_r\,d\mathbf{W}_r,\qquad \mathbf{x}_{\tinit}\sim\bar\rho_{\tinit},\qquad \mathbf{v}_r\doteq\mathbf{f}_r+\mathbf{u}_r .
    \end{equation}
\end{itemize}
\cite{guo2026complexity} state their Girsanov lemmas (Lems.~1 and~3) for a constant noise level \(\sigma\). For a scalar \(g_r>0\), they hold with \(\sigma^2\) replaced by \(g_r^2\) inside the time integrals: under the time change \(\tau(r)=\int_{\tinit}^rg_{r'}^2\,dr'\), the noise becomes standard and every drift is divided by \(g_r^2\), and changing back gives exactly this replacement. We use them in this form.

\begin{lemma}[Weights as a ratio of path measures]\label{lem:path-ratio}
    (Under RC.~\ref{cond:technical-path}.) For every \(t\in(\tinit,T]\), \(\mathbb P^{\leftarrow}\ll\mathbb P^{\rightarrow}\), and the normalised weight \(\bar w_t\) of Eq.~\ref{eq:normalised-knob-logweight} is
    \begin{equation}\label{eq:path-ratio}
        \bar w_t\;=\;\frac{d\mathbb P^{\leftarrow}}{d\mathbb P^{\rightarrow}}\big((\mathbf{x}_r)_{r\in[\tinit,t]}\big)\qquad\text{almost surely.}
    \end{equation}
    In particular \(\mathbb E[\bar w_t\,\Phi((\mathbf{x}_r)_{r\le t})]=\mathbb E_{\mathbb P^{\leftarrow}}[\Phi]\) for every bounded path functional \(\Phi\); the identity \(\mathbb E[\bar w_t\varphi(\mathbf{x}_t)]=\int\varphi\,\bar\rho_t\) used throughout is the case \(\Phi=\varphi(\mathbf{x}_t)\).
        \LineBetweenStatementAndProof
    \begin{proof}
\textbf{Overview.} Write down the Radon--Nikod\'ym derivative between a forward and a backward SDE given by Girsanov's theorem, and simplify it with It\^o's formula for \(\log\bar\rho_r(\mathbf{x}_r)\) and the Fokker--Planck equation Eq.~\ref{eq:proof-fp}. For \(\mathbf{f}\equiv0\) this is the computation of \cite[App.~C.1]{guo2026complexity}; the drift \(\mathbf{f}\) cancels.

\smallskip\noindent\textbf{Step 1 (Girsanov).} The forward drift of Eq.~\ref{eq:proof-wsde} is \(\mathbf{a}_r\doteq\mathbf{f}_r+\tfrac{g_r^2}{2}\mathbf{s}_r\), and the backward drift of Eq.~\ref{eq:backward-wsde} is \(\mathbf{b}_r\doteq\mathbf{f}_r-\tfrac{g_r^2}{2}\mathbf{s}_r\). By \cite[Lem.~3, part~2]{guo2026complexity}, evaluated along a path \(\mathbf{x}\sim\mathbb P^{\rightarrow}\),
\begin{align*}
    \log\frac{d\mathbb P^{\rightarrow}}{d\mathbb P^{\leftarrow}}(\mathbf{x})
    &=\log\frac{\bar\rho_{\tinit}(\mathbf{x}_{\tinit})}{\bar\rho_t(\mathbf{x}_t)}
    +\int_{\tinit}^{t}\frac{1}{g_r^2}\Big(\langle\mathbf{a}_r(\mathbf{x}_r),d\mathbf{x}_r\rangle-\langle\mathbf{b}_r(\mathbf{x}_r),*d\mathbf{x}_r\rangle\Big)
    \\
    &\quad {}-\int_{\tinit}^{t}\frac{\inlinesqnorm{\mathbf{a}_r}-\inlinesqnorm{\mathbf{b}_r}}{2g_r^2}(\mathbf{x}_r)\,dr ,
\end{align*}
where \(\langle\cdot,*d\mathbf{x}_r\rangle\) is the backward (right-endpoint) stochastic integral \cite[Def.~2]{guo2026complexity}.

\smallskip\noindent\textbf{Step 2 (backward integral to It\^o integral).} The backward integral equals the It\^o integral plus the cross-variation \cite[Eq.~(15)]{guo2026complexity}. Since \(d\mathbf{x}_r\) has noise part \(g_r\,d\mathbf{W}_r\), the cross-variation of \(\mathbf{b}_r(\mathbf{x}_r)\) with \(\mathbf{x}_r\) is \(g_r^2\,\nabla\cdot\mathbf{b}_r(\mathbf{x}_r)\,dr\) (cf.\ \cite[App.~C.1]{guo2026complexity}), so
\(\langle\mathbf{b}_r,*d\mathbf{x}_r\rangle=\langle\mathbf{b}_r,d\mathbf{x}_r\rangle+g_r^2\,\nabla\cdot\mathbf{b}_r\,dr\).

\smallskip\noindent\textbf{Step 3 (insert the drifts).} Since \(\mathbf{a}_r-\mathbf{b}_r=g_r^2\mathbf{s}_r\), \(\mathbf{a}_r+\mathbf{b}_r=2\mathbf{f}_r\) and \(\nabla\cdot\mathbf{s}_r=\Delta\log\bar\rho_r\),
\begin{align*}
    \tfrac{1}{g_r^2}\big(\langle\mathbf{a}_r,d\mathbf{x}_r\rangle-\langle\mathbf{b}_r,*d\mathbf{x}_r\rangle\big)
    &=\langle\mathbf{s}_r,d\mathbf{x}_r\rangle-\nabla\cdot\mathbf{b}_r\,dr
    =\langle\mathbf{s}_r,d\mathbf{x}_r\rangle+\tfrac{g_r^2}{2}\Delta\log\bar\rho_r\,dr-\nabla\cdot\mathbf{f}_r\,dr,\\
    \tfrac{1}{2g_r^2}\big(\inlinesqnorm{\mathbf{a}_r}-\inlinesqnorm{\mathbf{b}_r}\big)
    &=\tfrac{1}{2g_r^2}\langle\mathbf{a}_r-\mathbf{b}_r,\mathbf{a}_r+\mathbf{b}_r\rangle=\mathbf{f}_r\cdot\mathbf{s}_r .
\end{align*}

\smallskip\noindent\textbf{Step 4 (It\^o's formula).} Since \(d\mathbf{x}_r\,d\mathbf{x}_r^\top=g_r^2I\,dr\),
\(d\log\bar\rho_r(\mathbf{x}_r)=\partial_r\log\bar\rho_r\,dr+\langle\mathbf{s}_r,d\mathbf{x}_r\rangle+\tfrac{g_r^2}{2}\Delta\log\bar\rho_r\,dr\). Integrating,
\(\int_{\tinit}^t\big(\langle\mathbf{s}_r,d\mathbf{x}_r\rangle+\tfrac{g_r^2}{2}\Delta\log\bar\rho_r\,dr\big)=\log\bar\rho_t(\mathbf{x}_t)-\log\bar\rho_{\tinit}(\mathbf{x}_{\tinit})-\int_{\tinit}^t\partial_r\log\bar\rho_r\,dr\).

\smallskip\noindent\textbf{Step 5 (collect).} Substituting Steps 3--4 into Step 1, the boundary terms \(\log\bar\rho_{\tinit}(\mathbf{x}_{\tinit})\) and \(\log\bar\rho_t(\mathbf{x}_t)\) cancel, leaving
\begin{equation*}
    \log\frac{d\mathbb P^{\rightarrow}}{d\mathbb P^{\leftarrow}}(\mathbf{x})=-\int_{\tinit}^t\big(\partial_r\log\bar\rho_r+\nabla\cdot\mathbf{f}_r+\mathbf{f}_r\cdot\mathbf{s}_r\big)(\mathbf{x}_r)\,dr=-\int_{\tinit}^t\bar h_r(\mathbf{x}_r)\,dr=-\log\bar w_t ,
\end{equation*}
where the middle equality is Eq.~\ref{eq:proof-fp} divided by \(\bar\rho_r\): \(\partial_r\log\bar\rho_r=-\nabla\cdot\mathbf{f}_r-\mathbf{f}_r\cdot\mathbf{s}_r+\bar h_r\). This is Eq.~\ref{eq:path-ratio}. The last claim follows from \(\mathbb E_{\mathbb P^{\rightarrow}}[\tfrac{d\mathbb P^{\leftarrow}}{d\mathbb P^{\rightarrow}}\Phi]=\mathbb E_{\mathbb P^{\leftarrow}}[\Phi]\), and the marginal of \(\mathbb P^{\leftarrow}\) at time \(t\) is \(\bar\rho_t\).
    \end{proof}
\end{lemma}

\begin{theorem}[Weights and particles converge under finite residual action]\label{thm:finite-action}
    (Under RC.~\ref{cond:finite-action},~\ref{cond:technical-path}.) For every \(t\in[\tinit,T]\),
    \begin{align}
        \mathbb E\,\big|\bar w_t-1\big|\;\le\;2\sqrt{U_t},
        \quad&\text{hence}\quad
        \Pr\big(|\bar w_t-1|\ge\varepsilon\big)\;\le\;\frac{2\sqrt{U_t}}{\varepsilon}\quad\text{for every }\varepsilon>0,\label{eq:finite-action-weights}\\
        \mathrm{KL}(\bar\rho_t\,\Vert\,q_t)&\;\le\;\tfrac12U_t,
        \quad\quad
        \mathrm{TV}(q_t,\bar\rho_t)\;\le\;\tfrac12\sqrt{U_t}.\label{eq:finite-action-particles}
    \end{align}
    If \(g_t=g\gamma_t\) with a fixed profile \(\gamma_t>0\), then \(U_t=O(g^{-2})\). (This is because \(\mathbf{u}_t\) does not depend on \(g\): Eq.~\ref{eq:residual-transport} involves only \(\bar\rho_t\) and \(\bar h_t\), and \(h_t\) is fixed by \((\mathbf{f}_t,\rho_t)\) through Eq.~\ref{eq:proof-fp-unnormalised}, which does not involve \(g_t\); by Lemma~\ref{prop:langevin-gauge-MAIN}, the Langevin term changes neither the mass trajectory nor \(h_t\). So one and the same \(\mathbf{u}_t\) serves for every \(g\), and only the factor \(g_t^{-2}\) in \(U_t\) changes.) Hence the normalised weights converge to \(1\) in \(L^1\) and in probability at rate \(O(g^{-1})\), and the unweighted particles converge to the target in KL at rate \(O(g^{-2})\). The drift \(\mathbf{f}_t\) enters only through \(\mathbf{u}_t\). For \(\mathbf{f}_t\equiv0\), constant \(g\) and the horizon \([0,1]\), the smallest choice of \(\mathbf{u}_t\) gives \(U_1=\mathcal A/g^2\). After the time change \(\tau=g^2t/2\) (their \(T=g^2/2\)), the probability bound in Eq.~\ref{eq:finite-action-weights} is then exactly \cite[Thm.~2]{guo2026complexity}, and the KL bound in Eq.~\ref{eq:finite-action-particles}, \(\tfrac12U_1=\mathcal A/(4T)\), is exactly \cite[Thm.~1]{guo2025provable}.
        \LineBetweenStatementAndProof
    \begin{proof}
\textbf{Overview.} This is the proof of \cite[Thm.~2]{guo2026complexity} (their App.~C.2) with their \(\mathbf{v}_t\) replaced by \(\mathbf{u}_t\). The reference process \(\mathbb P\) has the exact marginals \(\bar\rho_r\); Girsanov's theorem gives its KL divergence to \(\mathbb P^{\rightarrow}\) and to \(\mathbb P^{\leftarrow}\); Pinsker's inequality and Lemma~\ref{lem:path-ratio} turn this into the bounds.

\smallskip\noindent\textbf{Step 1 (the reference process has marginals \(\bar\rho_r\)).} The Fokker--Planck equation of Eq.~\ref{eq:reference-sde} is \(\partial_r\nu_r=-\nabla\cdot\big(\nu_r(\mathbf{v}_r+\tfrac{g_r^2}{2}\mathbf{s}_r)\big)+\tfrac{g_r^2}{2}\Delta\nu_r\). For \(\nu_r=\bar\rho_r\), the two Langevin terms cancel because \(\bar\rho_r\mathbf{s}_r=\nabla\bar\rho_r\), and \(-\nabla\cdot(\bar\rho_r\mathbf{v}_r)=\partial_r\bar\rho_r\) by Eq.~\ref{eq:residual-transport} and Eq.~\ref{eq:proof-fp}. By uniqueness (RC.~\ref{cond:technical-path}), the marginal of \(\mathbb P\) at time \(r\) is \(\bar\rho_r\). By Nelson's relation \cite[Lem.~2]{guo2026complexity}, \(\mathbb P\) is also the law of the backward SDE \(d\mathbf{x}_r=\big[\mathbf{v}_r-\tfrac{g_r^2}{2}\mathbf{s}_r\big](\mathbf{x}_r)\,dr+g_r\,d\mathbf{W}^{\leftarrow}_r\), \(\mathbf{x}_t\sim\bar\rho_t\), since \(\mathbf{v}_r+\tfrac{g_r^2}{2}\mathbf{s}_r-g_r^2\mathbf{s}_r=\mathbf{v}_r-\tfrac{g_r^2}{2}\mathbf{s}_r\).

\smallskip\noindent\textbf{Step 2 (Girsanov).} Forwards, \(\mathbb P\) and \(\mathbb P^{\rightarrow}\) start from the same law \(\bar\rho_{\tinit}\), and their drifts differ by \(\big(\mathbf{v}_r+\tfrac{g_r^2}{2}\mathbf{s}_r\big)-\big(\mathbf{f}_r+\tfrac{g_r^2}{2}\mathbf{s}_r\big)=\mathbf{u}_r\). By \cite[Lem.~1]{guo2026complexity}, and since \(\mathbf{x}_r\sim\bar\rho_r\) under \(\mathbb P\),
\(\mathrm{KL}(\mathbb P\Vert\mathbb P^{\rightarrow})=\int_{\tinit}^t\frac{1}{2g_r^2}\mathbb E_{\mathbb P}\inlinesqnorm{\mathbf{u}_r(\mathbf{x}_r)}\,dr=\tfrac12U_t\).
Backwards, \(\mathbb P\) and \(\mathbb P^{\leftarrow}\) end at the same law \(\bar\rho_t\), and their backward drifts differ by \(\big(\mathbf{v}_r-\tfrac{g_r^2}{2}\mathbf{s}_r\big)-\big(\mathbf{f}_r-\tfrac{g_r^2}{2}\mathbf{s}_r\big)=\mathbf{u}_r\), so \cite[Lem.~3, part~1]{guo2026complexity} gives \(\mathrm{KL}(\mathbb P\Vert\mathbb P^{\leftarrow})=\tfrac12U_t\) as well.

\smallskip\noindent\textbf{Step 3 (weights).} By Lemma~\ref{lem:path-ratio}, \(\mathbb E|\bar w_t-1|=\mathbb E_{\mathbb P^{\rightarrow}}\big|\tfrac{d\mathbb P^{\leftarrow}}{d\mathbb P^{\rightarrow}}-1\big|=2\,\mathrm{TV}(\mathbb P^{\leftarrow},\mathbb P^{\rightarrow})\). By the triangle inequality and Pinsker's inequality \(\mathrm{TV}\le\sqrt{\mathrm{KL}/2}\),
\begin{equation*}
    2\,\mathrm{TV}(\mathbb P^{\leftarrow},\mathbb P^{\rightarrow})\le2\,\mathrm{TV}(\mathbb P,\mathbb P^{\leftarrow})+2\,\mathrm{TV}(\mathbb P,\mathbb P^{\rightarrow})
    \le2\sqrt{\tfrac14U_t}+2\sqrt{\tfrac14U_t}=2\sqrt{U_t}.
\end{equation*}
The probability bound is Markov's inequality.

\smallskip\noindent\textbf{Step 4 (particles).} The marginals at time \(t\) of \(\mathbb P\) and \(\mathbb P^{\rightarrow}\) are \(\bar\rho_t\) and \(q_t\). Taking a marginal does not increase KL (data-processing inequality), so \(\mathrm{KL}(\bar\rho_t\Vert q_t)\le\mathrm{KL}(\mathbb P\Vert\mathbb P^{\rightarrow})=\tfrac12U_t\), and Pinsker gives \(\mathrm{TV}(q_t,\bar\rho_t)\le\sqrt{U_t/4}\).

\smallskip\noindent\textbf{Step 5 (rate and the case \(\mathbf{f}\equiv0\)).} If \(g_r=g\gamma_r\), then \(U_t=g^{-2}\int\lVert\mathbf{u}_r\rVert^2_{L^2(\bar\rho_r)}\gamma_r^{-2}\,dr\), and the integral does not depend on \(g\) because \(\mathbf{u}_r\) does not (see the statement). For \(\mathbf{f}\equiv0\), constant \(g\) and the smallest \(\mathbf{u}_r\), \(\lVert\mathbf{u}_r\rVert_{L^2(\bar\rho_r)}=|\dot{\bar\rho}|_r\) \cite[Lem.~4]{guo2026complexity}, so \(U_1=\mathcal A/g^2\). With \(T=g^2/2\), the right-hand side of \cite[Eq.~(18)]{guo2026complexity} is \(\tfrac{\sqrt2}{\varepsilon}\cdot2\sqrt{\mathcal A/(4T)}=\tfrac{2}{\varepsilon}\sqrt{\mathcal A/g^2}\), which is the probability bound in Eq.~\ref{eq:finite-action-weights}.
    \end{proof}
\end{theorem}

\begin{corollary}[Unnormalised weights and effective sample size]\label{cor:finite-action-consequences}
    (Under the hypotheses of Thm.~\ref{thm:finite-action}.)
    \begin{enumerate}
        \item[(i)] \emph{Unnormalised weights.} The weights \(w_t\) of Eq.~\ref{eq:proof-wsde}, i.e.\ those computed with the unnormalised \logweighingfunction{} \(h_t\) (for the PoIE trajectory, \(h_t=h^{\knob}_t\), the \logweighingfunction{} of the main text), are \(w_t=(Z_t/Z_{\tinit})\,\bar w_t\) (Eq.~\ref{eq:normalised-knob-logweight}), so
        \(\mathbb E\big|w_t-Z_t/Z_{\tinit}\big|\le2\,(Z_t/Z_{\tinit})\sqrt{U_t}\): \(w_t\to Z_t/Z_{\tinit}\) in \(L^1\) and in probability.
        \item[(ii)] \emph{Effective sample size.} For \(N\) independent particles with weights \(w^{(1)}_t,\dots,w^{(N)}_t\), the effective sample size \(\mathrm{ESS}=1/\sum_n(W^{(n)}_t)^2\) of Sec.~\ref{sec:preliminaries-WSDEs-MAIN}, with normalised weights \(W^{(n)}_t=w^{(n)}_t/\sum_mw^{(m)}_t=\bar w^{(n)}_t/\sum_m\bar w^{(m)}_t\), satisfies
        \begin{equation*}
            \Pr\Big(\mathrm{ESS}<N\,\tfrac{(1-\varepsilon)^2}{(1+\varepsilon)^2}\Big)\;\le\;\frac{2N\sqrt{U_t}}{\varepsilon}\qquad\text{for every }\varepsilon\in(0,1).
        \end{equation*}
        Hence, for every fixed \(N\), \(\mathrm{ESS}/N\to1\) in probability as \(U_t\to0\). The normalised weights, and so the ESS, are the same for \(w_t\) and \(\bar w_t\).
    \end{enumerate}
        \LineBetweenStatementAndProof
    \begin{proof}
\textbf{(i)} Since \(h_t-\bar h_t=\partial_t\log Z_t\) does not depend on \(\mathbf{x}\), \(\log w_t-\log\bar w_t=\int_{\tinit}^t\partial_s\log Z_s\,ds=\log(Z_t/Z_{\tinit})\) on every path. Multiply Eq.~\ref{eq:finite-action-weights} by \(Z_t/Z_{\tinit}\).

\smallskip\noindent\textbf{(ii)} On the event that \(|\bar w^{(n)}_t-1|<\varepsilon\) for every \(n\), each normalised weight \(\bar w^{(n)}_t\) lies in \((1-\varepsilon,1+\varepsilon)\), so \(\mathrm{ESS}=\big(\sum_n\bar w^{(n)}_t\big)^2/\sum_n(\bar w^{(n)}_t)^2\ge N^2(1-\varepsilon)^2/\big(N(1+\varepsilon)^2\big)\). By a union bound over the \(N\) particles and Eq.~\ref{eq:finite-action-weights}, the complement of this event has probability at most \(N\cdot2\sqrt{U_t}/\varepsilon\). The deterministic factor \(Z_t/Z_{\tinit}\) cancels in \(W^{(n)}_t\).
    \end{proof}
\end{corollary}

\begin{remark}[Weight variance under an exponential moment]\label{rem:finite-action-variance}
    The proof of \cite[Prop.~3]{guo2026complexity} (their App.~G.4) carries over in the same way, with \(\mathbf{v}_t/\sqrt2\) replaced by \(\mathbf{u}_t/g_t\), and gives
    \begin{equation*}
        \Var[\bar w_t]\;\le\;\Big(\mathbb E_{\mathbb P}\exp\Big(28\int_{\tinit}^t\frac{\inlinesqnorm{\mathbf{u}_r(\mathbf{x}_r)}}{g_r^2}\,dr\Big)\Big)^{1/2}-1 ,
    \end{equation*}
    with \(\mathbb P\) the reference process Eq.~\ref{eq:reference-sde}. For example, if \(\lVert\mathbf{u}_r(\mathbf{x})\rVert_2\le\bar u_r\) for all \(\mathbf{x}\), then \(\Var[\bar w_t]\le\exp\big(14\int_{\tinit}^t\bar u_r^2/g_r^2\,dr\big)-1=O(g^{-2})\). This needs an exponential moment of \(\mathbf{u}_t\) along the paths of \(\mathbb P\), which is stronger than RC.~\ref{cond:finite-action}. (For the unnormalised weights, \(\Var[w_t]=(Z_t/Z_{\tinit})^2\,\Var[\bar w_t]\).)
\end{remark}

\begin{remark}[Caveat: Perfect sampling only at the cost of increasingly fine (Euler--Maruyama integration) step size]\label{rem:caveat-require-fine-step-size}
    The result is a statement about the continuous-time WSDE. Integrating Eq.~\ref{eq:proof-wsde} with Euler--Maruyama requires \(\Delta t \lesssim 1/(g_t^2L_t)\) for stability and accuracy, where \(L_t\) is the Lipschitz constant of \(\nabla_{\mathbf{x}}\log\rho_t\), so the variance reduction is bought with proportionally more steps --- as it must be, since at \(g\to\infty\) the scheme degenerates to running exact MCMC at every time slice. The theorem says the trade-off is available, not that it is free; this is the sense in which Prop.~\ref{prop:CONJECTURE-Langevin-Infinity-MAIN} is ``trivial''.
\end{remark}

\end{document}